\documentclass{article}
\usepackage{icml2026}

\usepackage{amsmath,amssymb}
\usepackage{bbm}
\usepackage{graphicx}
\usepackage{booktabs}
\usepackage{multirow}
\usepackage{algorithm}
\usepackage{algorithmic}
\usepackage{xcolor}
\usepackage{hyperref}
\usepackage{placeins}
\usepackage{xcolor}
\definecolor{darkgreen}{rgb}{0.0, 0.5, 0.0}
\definecolor{mplblue}{rgb}{0.1216,0.4667,0.7059}
\newcommand{\norm}[1]{\left\lVert#1\right\rVert}

\title{Drift Estimation for Stochastic Differential Equations with Denoising Diffusion Models}
\icmltitlerunning{Drift Estimation for SDEs using Denoising Diffusion Models}

\author{
Marcos Tapia Costa$^{1,2}$ \and
Nikolas Kantas$^{1}$ \and
George Deligiannidis$^{2}$
}
\icmlcorrespondingauthor{Marcos Tapia Costa}{}
\begin{document}
\maketitle

\footnotetext[1]{Department of Mathematics, Imperial College London, UK}
\footnotetext[2]{Department of Statistics, University of Oxford, UK}

\setlength{\abovedisplayskip}{6pt}
\setlength{\belowdisplayskip}{6pt}
\setlength{\abovedisplayshortskip}{4pt}
\setlength{\belowdisplayshortskip}{4pt}
\begin{abstract}
We study the estimation of time-homogeneous drift functions in multivariate stochastic differential equations with known diffusion coefficient, from multiple trajectories observed at high frequency over a fixed time horizon. We formulate drift estimation as a denoising problem conditional on previous observations, and propose an estimator of the drift function which is a by-product of training a conditional diffusion model capable of simulating new trajectories dynamically. Across different drift classes, the proposed estimator was found to match classical methods in low dimensions and remained consistently competitive in higher dimensions, with gains that cannot be attributed to architectural design choices alone.
\end{abstract}
\section{Problem Statement}
Suppose one is interested in modelling a time series as a continuous-time stochastic differential equation (SDE),
\begin{equation}
 dY_{t} = \mu(Y_{t})dt + \sigma dB_{t}, t \in [0, T], \label{eq:BSDE}
\end{equation}
where \((B_{t})_{t\in[0, T]}\) is a \(D\)-dimensional Brownian motion and \(\mu: \mathbbm{R}^{D}\rightarrow \mathbbm{R}^{D}\) is a time-homogeneous drift function. We assume \(\sigma\) is known or well-estimated, i.e., using the quadratic variation of a single path, see \cite{Jacod1994, BarndorffNielsenShephard2002, JacodProtter2012}. We also assume \(\mu\) is locally Lipschitz to ensure a strong solution to \eqref{eq:BSDE} up to terminal time \(T\), see \cite{KaratzasShreve2000}.

We will investigate the case we are 
given a dataset \(\mathcal{D}\) consisting of \(I\gg 1\) i.i.d. trajectories, each with \(J\) observations of dimension \(D\) observed at times \(t_{j} = j\Delta \ \forall \ j \in \{1, \cdots, J\}\), such that \(\mathcal{D}=\{Y_{t_{1}}^{(i)}, \cdots, Y_{t_{J}}\}_{i\in[I]}, [I]:=\{1,\cdots, I\}\) and \(\Delta:=t_{j+1}-t_{j}\) is small. In our setting, the number of trajectories \(I\) and the number of discrete observations \(J\) are fixed with a constant sampling frequency \(\Delta\). 

Without loss of generality, we assume \(T=1\) and \(Y_{0}^{(i)} \sim Q_{0}\), where \(Q_{0}\) admits a density with respect to the Lebesgue measure \(dy_{0}\); \(Q_{0}(dy_{0}) = q_{0}(y_{0})dy_{0}\). We will study the problem of estimating the drift function \(\mu\) from the dataset \(\mathcal{D}\).

\subsection{Drift estimation as a regression}\label{section:DriftEstimationProblemStatement}
Letting \(Z_{t_{j}} := Y_{t_{j}}-Y_{t_{j-1}}\), the solution over \(s\in[t_{j-1},t_{j}]\) to \eqref{eq:BSDE} takes the form, 
\begin{equation}
   Z_{t_{j}} = \int_{t_{j-1}}^{t_{j}}\mu(Y_{s})ds +\sqrt{\Delta}\ \omega, \ \omega \sim \mathcal{N}(0, I). \label{eq:BSDEIncrement}
\end{equation}
For small \(\Delta\),  an Euler-Maruyama (EM) approximation to \eqref{eq:BSDEIncrement} above yields,
\begin{equation}
    \widehat{Z}_{t_{j}} = \mu(Y_{t_{j-1}})\Delta + \sqrt{\Delta}\omega, \omega\sim\mathcal{N}(0, I)\label{eq:EMApproximation}
\end{equation}
and the drift enters \(\mu\) through the conditional mean:
\begin{equation}
    \mathbbm{E}\left[\widehat{Z}_{t_{j}}\mid Y_{t_{j-1}}\right] = \mu(Y_{t_{j-1}})\Delta. \label{eq:EMApproxCondMean}
\end{equation}
In our multiple trajectory setting, with \(I \gg 1\), one could use \eqref{eq:EMApproxCondMean} to learn \(\mu\) by regressing states against increments through a nonparametric function \(D_{\theta}\) \cite{ZhaoLiuHoffmann2025}, 
\begin{equation}
    D_{\theta^{*}} = \arg\min_{\theta} \frac{1}{2}\sum_{i,j}\left[\norm{D_{\theta}(Y^{(i)}_{t_{j-1}})- Z^{(i)}_{t_{j}}}_{2}^{2}\right].\label{eq:RegressionOfDrift}
\end{equation}
It is well known that in our fixed \(T, J, \Delta\) setting, the drift \(\mu\) cannot be estimated consistently from a single trajectory \cite{Sorensen2015}. However, in the multiple trajectory setting, \cite{ZhaoLiuHoffmann2025} provide asymptotic convergence guarantees for separable drift functions under conditions on the class of functions \(D_{\theta}\). Yet, the drift estimator \(D_{\theta^{*}}\) can suffer from high variance, as the drift contribution scales as \(\mathcal{O}(\Delta)\), and the diffusion noise scales as \(\mathcal{O}(\sqrt{\Delta})\) \cite{Gobet2002, Kutoyants2004}, and the Fisher information per observation vanishes as \(\Delta \rightarrow 0\). In high dimensions, this difficulty is further compounded by deteriorating nonparametric regression convergence rates with dimension, requiring significantly more data to achieve the same accuracy as in lower dimensions \cite{Tsybakov2009}.

A general approach to mitigate variance in regression problems is to inject additive Gaussian noise into the inputs, which acts as a form of Tikhonov regularisation on \(D_{\theta}\) \cite{Bishop1995}. More generally, under the loss in \eqref{eq:RegressionOfDrift}, regression with noisy inputs corresponds to learning the conditional expectation of the target given the corrupted input, i.e., learning a \textit{denoiser}. Moreover, \cite{StarnesWebster2025} show learning the denoiser from Gaussian perturbations can smooth low variance, high curvature directions during optimisation and lead to a better conditioned stochastic gradient descent problem. \cite{Vincent2011} further showed that such denoising objectives are equivalent to score matching under Gaussian perturbations. Modern denoising diffusion models extend this principle by introducing a controlled, multi-scale noising process and learning denoisers across noise levels, with great success in image and video applications \cite{SohlDicksteinEtAl2015, HoEtAl2020, SongEtAl2020}.

In this paper, we extend the methodology of denoising diffusion models to introduce a controlled noising process and derive an estimator for the drift \(\mu\). We will propose a novel construction for the estimator, and present systematic empirical results showing that the denoising-based estimator improves out-of-sample performance in high-dimensional settings with strong interactions or chaotic dynamics, while remaining competitive when standard regression already generalises well. Theoretical guarantees as in \cite{ZhaoLiuHoffmann2025} are left for future work.

\section{Related Literature}
\subsection{Drift estimation for stochastic differential equations}
The problem of estimating the drift function of an SDE has been extensively studied. In the single trajectory setting, \cite{Kessler1997} derived an efficient estimator for the drift function of an ergodic scalar diffusion observed at discrete times when transition densities are unknown. \cite{ShojiOzaki1998} developed a pseudo-likelihood estimator from a locally linear approximation of the SDE, while \cite{AitShalia1998} used polynomial Hermite expansions of the transition density to obtain an approximate maximum likelihood estimator.

Nonparametric approaches were introduced through the kernel regression estimator of \cite{Watson1964}, and \cite{Nadaraya1964} was first applied to Brownian SDEs by \cite{FlorensZmirou1993}, who estimate both drift and diffusion coefficients in scalar diffusions, and \cite{BandiPhillips2003, BandiPhillips2007} provided uniform convergence rates for its application to high frequency financial data. Furthermore, \cite{Dalalyan2006} derived a nonparametric estimator for homogeneous drift functions that achieves minimax optimal rates, while \cite{ComteGenonCatalot2009, ComteGenonCatalot2010} proposed penalised least squares projection methods to jointly estimate the drift and diffusion coefficients nonparametrically. More recently, \cite{OgaKoike2023} develop theoretical guarantees for multidimensional drift estimation for a class of deep neural networks, given a single path with many observations.

In the multiple trajectory setting, which is the observation regime of this paper, \cite{BonalliRudi2025} used a Reproducing Kernel Hilbert Space approximation of the Fokker-Plank equation to estimate multidimensional non-linear drift functions.  \cite{MohammadiEtAl2024} use Functional Data Analysis to develop a drift estimator given sparse observations. \cite{MarieEtAl2021} extend kernel smoothing estimators of the drift function to the multiple i.i.d. trajectory setting, while \cite{ComteGenonCatalot2020, DenisEtAl2021} extend penalised least squares estimators to the same setting. More recently, \cite{ZhaoLiuHoffmann2025} provide asymptotic convergence guarantees for separable drift estimation in a compact domain using feed-forward neural networks. Their empirical evaluation, however, is restricted to a single test drift function and assesses recovery only along one coordinate of a high-dimensional system, leaving performance on fully coupled dynamics unexamined.

\subsection{Score-based generative models}
Score-based generative models, or \emph{Denoising Diffusion Models (DDMs)}, were introduced in \cite{HoEtAl2020} and \cite{SongEtAl2020}, drawing ideas from score-matching \cite{Hyvarinen2005,Vincent2011} and building on earlier work \cite{SohlDicksteinEtAl2015}. By using neural networks to estimate the log density of a smoothed data distribution, i.e. the \textit{score}, they have achieved state-of-the-art results in image generation \cite{HoEtAl2020, DhariwalNichol2021, SahariaEtAl2022, BrooksEtAl2023}, and have found numerous applications in video generation \cite{HoEtAl2022}, molecule and drug design, \cite{GeomXuEtAl2023,CorsoEtAl2023}, strategy testing \cite{KoshiyamaEtAl2021}, and data sharing \cite{Assefa2020}. 

Their state-of-the-art performance across several domains, and the fact that they are less susceptible to training instabilities or mode-collapse, have established DDMs as the gold-standard in generative modelling \cite{DhariwalNichol2021}. In particular, \cite{SongErmon2019} extended \cite{SohlDicksteinEtAl2015} to a practical setting by formalising Stochastic Gradient Langevin Dynamics (SGLD) as the latent variable process, while \cite{HoEtAl2020} introduced Denoising Diffusion Probabilistic Models (DDPM). These approaches were unified under an SDE-based framework in \cite{SongEtAl2021B}, who show that the forward processes in DDPM and SGLD are discretisations of continuous-time Itô-SDEs. Further generalisations have been achieved in recent years, see \cite{SongEtAl2021A, FranzeseEtAl2023A, ZhangEtAl2023, BentonEtAl2023}. Other contributions have proposed improvements to accelerate sampling or to improve sample quality, such as \cite{JolicoeurMartineauEtAl2021, BaoEtAl2022, WatsonEtAl2022, XuEtAl2022, RombachEtAl2022}. Finally, using ideas from optimal transport, \cite{DeBortoliEtAl2021, ShiEtAl2022, ShiEtAl2023} developed a parallel line of work, connecting diffusion models to Schrödinger bridge problems.

\subsubsection{Learning the score for time series data}
Beyond image and spatial data, score-based diffusion models have also been extensively applied to time-series settings, see \cite{DiffusionTimeSurvey2023} for a comprehensive survey. While the methods reviewed below are not used directly in this work, they illustrate how diffusion models have been adapted to temporal and conditional structures, and therefore provide relevant context for our setting.

In the field of time series forecasting, \cite{LimEtAl2023} draw inspiration from the TimeGrad model in \cite{RasulEtAl2021, KongEtAl2021} and the ScoreGrad model in \cite{YanEtAl2021} to propose the TSGM model, a conditional diffusion model that operates on a low-dimensional representation of the data. \cite{ShenKwok2023} remove the sequence-based architecture (LSTM, Transformer) to achieve non-autoregressive forecasting. For time-series with missing values, \cite{TashiroEtAl2021, LopezAlcarazStrodthoff2023, KolloviehEtAl2023} proposed conditional diffusion frameworks that impute unobserved entries. For long-horizon multivariate sequence generation, \cite{YangEtAl2024} introduce SigDiffusion, which leverages path signatures. In the domain of filtering, \cite{BaoEtAl2022, BaoEtAl2023} applied score-based conditional diffusion models to non-linear filtering problems.

\section{Background}
\subsection{Score-based diffusion models}
Score-based diffusion models are generative models that learn a reversible transformation between two distributions, \(p_{0}\) and \(p_{1}\). \cite{SongEtAl2020} used a forward SDE to transform samples distributed according to \(p_{0}\) to samples distributed according to \(p_{1}\). Defining \(\gamma(\tau) = \gamma_{0}+\tau(\gamma_{1}-\gamma_{0})\), the forward SDE takes samples \(X_{0} \sim p_{0}\) and diffuses them towards \(X_{1} \sim p_{1}\) via a Variance-Preserving SDE (VPSDE),
\begin{equation}
 dX_{\tau} = -\frac{1}{2}\gamma(\tau)X_{\tau}d\tau + \sqrt{\gamma(\tau)}dW_{\tau}, \ \tau \in [0,1],\label{eq:VPSDE}
\end{equation}
where \((W_{\tau})_{\tau \in [0,1]}\) is a \(D\)-dimensional Brownian motion, see Section 3.4 in \cite{SongEtAl2020}. We emphasise \eqref{eq:VPSDE} is a different SDE to the data-generating process assumed in \eqref{eq:BSDE}, and which defines the distribution \(p_{0}\). From \cite{Anderson1982}, the forward SDE in \eqref{eq:VPSDE} admits a time-reversal which diffuses \(\widetilde{X}_{0}\sim p_{1}\) back to \(\widetilde{X}_{1} \sim p_{0}\) via,
\begin{equation}
\begin{aligned}
    d\widetilde{X}_{\tau}
    = \left[\frac{1}{2}\widetilde{X}_{\tau}
    + \nabla_{\widetilde{X}_{\tau}}\ln p_{\tau}(\widetilde{X}_{\tau})\right]
    &\gamma(\tau)\, d\tau \\
    \quad &+ \sqrt{\gamma(\tau)}\, d\widetilde{W}_{\tau}, \label{eq:UnconditionalReverseSDE}
\end{aligned}
\end{equation}
such that \(\widetilde{X}_{\tau} \stackrel{d}{=} X_{1-\tau}, \widetilde{W}_{\tau} \stackrel{d}{=}W_{1-\tau}  \ \forall \tau \in [0,1]\), where \( \stackrel{d}{=}\) means equal in distribution.

The reverse-time SDE in \eqref{eq:UnconditionalReverseSDE} requires access to the score \(\nabla_{X_{\tau}}\ln p_{\tau}(X_{\tau})\) which is not available in closed form in general. \cite{Vincent2011, SongEtAl2020} show that the oracle score-matching objective is,
\begin{equation*}
	l(\theta)= \int_{0}^{1}\mathop{\mathbbm{E}}\Big[\lambda(\tau)\left\|s_{\theta}(\tau, X_{\tau}) - \nabla_{X_{\tau}}\ln p_{\tau}(X_{\tau})\right\|_{2}^{2}\Big]d\tau, 
\end{equation*}
where \(X_{0}\sim p_{0}(x_{0}), X_{\tau} \sim q_{\tau}(x_{\tau}\mid X_{0})\) and \(s_{\theta}: [0, 1] \times \mathbbm{R}^{D} \rightarrow \mathbbm{R}^{D}\) is a neural network parameterised by \(\theta\). The function \(\lambda : [0, 1] \rightarrow \mathbbm{R}^{+}\) is a positive weighting function, and \(q_{\tau}(x_{\tau}\mid X_{0})\) is the transition density of \eqref{eq:VPSDE},
\begin{equation}
 q_{\tau}(x_{\tau}\mid X_{0}) = \mathcal{N}(x_{\tau};\beta_{\tau}X_{0}, \sigma_{\tau}^{2}I), \label{eq:VPSDETransition}
\end{equation}
where we have defined \(\sigma_{\tau}^{2} = 1 - \beta^{2}_{\tau}\) and \(\beta_{\tau} = \exp\left[-0.25\tau^{2}(\gamma_{1}-\gamma_{0}) - 0.5\tau\gamma_{0}\right]\).

However, the objective \(l(\theta)\) cannot be evaluated in practice since we do not have access to the score. Under sufficient expressivity of the neural network \(s_{\theta}\), \cite{Vincent2011} show that the denoising score-matching objective has the same global minimum as \(l(\theta)\),
\begin{equation*}
\begin{aligned}
\widetilde{l}(\theta) = \int_{0}^{1}\mathbbm{E}\Big[\lambda(\tau)\big\|
s_{\theta}(\tau, X_{\tau})-\nabla_{X_{\tau}}\ln q_{\tau}(X_{\tau} \mid X_{0})\big\|_{2}^{2}\Big]\, d\tau.
\end{aligned}
\end{equation*}
\cite{SongErmon2019}, \cite{SongEtAl2020} proposed the weighting function \(\lambda(\tau) = \sigma_{\tau}^{2}\) to reduce the variance of the target \(\nabla_{X_{\tau}}\ln q_{\tau}(X_{\tau} \mid X_{0})\ = -\sigma_{\tau}^{-2}(X_{\tau}-\beta_{\tau}X_{0})\).
In practice, the objective \(\widetilde{l}(\theta)\) is approximated via Monte Carlo sampling. In particular, one samples \((\tau, X_{0}) \sim \mathcal{U}[\epsilon, 1] \times p_{0}\) and then \(X_{\tau} \sim q_{\tau}(x_{\tau}\mid X_{0})\) for each sample. The constant \(\epsilon >0\) is chosen to avoid the explosion of the target since \(\lim_{\tau\rightarrow 0}\sigma_{\tau}=0\).

\subsection{Conditional diffusion models}\label{section:ConditionalDiffusionModels}
Letting \((X_{0}, Y)\sim p_{0}(x_{0}, y), Y \in \mathbbm{R}^{D}\),  \cite{SongEtAl2020}, \cite{TashiroEtAl2021} apply score-based diffusion models to generate samples from \(p_{0}(x_{0}\mid Y)\). For each fixed \(Y\), applying the forward diffusion \eqref{eq:VPSDE} to \(X_{0}\) induces a push-forward distribution \(p_{\tau}(x_{\tau}\mid Y) \,  \forall \,  \tau \in [0,1]\). Similar to \eqref{eq:UnconditionalReverseSDE}, the reverse-time SDE for conditional generation is:
\begin{equation}
\begin{aligned}
 d\widetilde{X}_{\tau} = \Big[\frac{1}{2}\widetilde{X}_{\tau} + \nabla_{\widetilde{X}_{\tau}}\ln p_{\tau}(\widetilde{X}_{\tau}\mid Y)&\Big]\gamma(\tau)d\tau  \\
\quad &+  \sqrt{\gamma(\tau)}d\widetilde{W}_{\tau}.
 \end{aligned}\label{eq:ConditionalReverseVPSDE}
\end{equation}
(Theorem 1, Appendices B.1-B.2) \cite{Batzolis2021} show one can learn the conditional score in \eqref{eq:ConditionalReverseVPSDE} consistently by minimising the following objective,
\begin{equation}
\begin{aligned}
\tilde{L}(\theta) = \mathbbm{E}\Bigl[ \lambda(\tau) \bigl\|s_{\theta}(\tau,& \, X_{\tau}, Y)\\
&- \nabla_{X_{\tau}}\ln q_{\tau}(X_{\tau} \mid X_{0}) \bigr\|_{2}^{2} \Bigr],
\end{aligned}
\label{eq:FeasibleConditionalObjective}
\end{equation}

The conditional score can be written in terms of the denoiser,
\begin{equation}
    \nabla_{X_{\tau}}\ln p_{\tau}(X_{\tau} \mid Y) = -\frac{X_{\tau}}{\sigma_{\tau}^{2}} + \frac{\beta_{\tau}}{\sigma_{\tau}^{2}}\mathbbm{E}\left[X_{0}\mid X_{\tau}, Y\right],\label{eq:ScorePosteriorExpectation}
\end{equation}
where the denoiser is given by, 
\begin{equation}
    \mathbbm{E}\left[X_{0}\mid X_{\tau}, Y\right] = \int_{\mathbbm{R}^{D}} x_{0}p(x_{0}\mid X_{\tau}, Y)dx_{0}.\label{eq:PosteriorExpectationSolution}
\end{equation}
To learn the denoiser directly, one can re-parameterise \(s_{\theta}\) using a network \(D_{\theta}\),
\begin{equation}
    s_{\theta}(\tau, X_{\tau}, Y) = -\frac{X_{\tau}}{\sigma_{\tau}^{2}} + \frac{\beta_{\tau}}{\sigma_{\tau}^{2}}D_{\theta}(\tau, X_{\tau}, Y)\label{eq:NetworkReparameterisation}
\end{equation}
which can be plugged into the learning objective \eqref{eq:FeasibleConditionalObjective} for \(s_{\theta}\) to yield a learning objective for \(D_{\theta}\),
\begin{equation}
	\tilde{L}(\theta) = \mathop{\mathbbm{E}}\left[\left\|D_{\theta}(\tau, X_{\tau}, Y) - X_{0}\right\|_{2}^{2}\right] \label{eq:FeasiblePosteriorConditionalObjective}.
\end{equation}

\section{Estimation of the drift function}
In this section, we study how to recover the drift \(\mu(Y)\) from a denoising model trained to approximate the denoiser \(\mathbb{E}[X_0 \mid X_\tau, Y]\). We exploit the Gaussian structure of the forward process to construct a Monte Carlo estimator for \(\mu\).

\subsection{Data specification for conditional diffusion models}\label{section:DriftEstimationDiffusionModelMethodology}
Given trajectories \(Y^{(i)}\) observed at times \(t_j = j\Delta\), recall we define \(\forall j\in\{0,\dots,J-1\}\) the increment,
\begin{equation*}
    Z^{(i)}_{t_j} := Y^{(i)}_{t_{j+1}} - Y^{(i)}_{t_j}.
\end{equation*}
From \eqref{eq:BSDEIncrement}, \(Z^{(i)}_{t_{j}}\) depends on the past trajectory \((Y^{(i)}_{t_s})_{s=0}^j\) only through the current observation \(Y^{(i)}_{t_j}\).
Consequently, under the time-homogeneous dynamics in \eqref{eq:BSDE}, the conditional law of \(Z_{t_{j}}\) given \(Y_{t_j}\) is independent of \(j\), and we work with generic pairs \((Y, Z)\), suppressing the indices \(i\) and \(j\). 

Henceforth, \(X_{0}\) will denote an increment from the data generating distribution \(p_{0}(x_{0}\mid Y)\), and \(X_{\tau}\) is the diffused version of \(X_{0}\) after applying the forward process \(q_{\tau}\) in \eqref{eq:VPSDETransition}. We aim to learn an approximation \(D_{\theta}(\tau, X_{\tau}, Y)\) to the denoising target \(\mathbb{E}[X_0 \mid X_\tau, Y]\) by minimising \eqref{eq:FeasibleConditionalObjective}.

\subsection{Learning the drift by denoising}\label{section:DriftEstimationPostMean}
We suppose henceforth that \(\theta^{*}\) is the minimiser of \eqref{eq:FeasiblePosteriorConditionalObjective} such that \(D_{\theta^{*}}(\tau, X_{\tau}, Y)\) is the trained approximation to \( \mathbbm{E}\left[X_{0}\mid X_{\tau}, Y\right]\). We also use the Euler-Maruyama approximation to the increment distribution in \eqref{eq:EMApproxCondMean}, \(p_{0}(x_{0}\mid Y) \approx  \widehat{p}_{0}(x_{0}\mid Y) = \mathcal{N}(x_{0};\mu(Y)\Delta, \Delta)\) to derive an estimator for \(\mu\).

Under the VPSDE, the transition kernel \(q_{\tau}(x_{\tau}\mid X_{0})\) in \eqref{eq:VPSDETransition} is a Gaussian linear in \(X_{0}\). Since the conditional distribution \(\widehat{p}_{0}(x_{0}\mid Y)\) is also Gaussian, a closed-form solution for the denoiser is obtained under the EM approximation,
\begin{equation*}
    \mathbbm{E}\left[X_{0}\mid X_{\tau}, Y\right] \approx \frac{\sigma_{\tau}^{2}\Delta}{\sigma_{\tau}^{2}+\beta_{\tau}^{2}\Delta}\left[\mu(Y) + \frac{\beta_{\tau}}{\sigma_{\tau}^{2}}X_{\tau}\right],
\end{equation*}
and the drift is obtained through, 
\begin{equation}
    \mu(Y)  =  a(\tau)X_{\tau} + b_{\Delta}(\tau){\mathbbm{E}}\left[X_{0}\mid X_{\tau}, Y\right], \label{eq:IdealDriftEstimator} 
\end{equation}
where,
\begin{align}
    a(\tau) := -\frac{\beta_{\tau}}{\sigma_{\tau}^{2}},  \, \, \, \, b_{\Delta}(\tau) :=  \frac{\sigma_{\tau}^{2}+\beta_{\tau}^{2}\Delta}{\sigma_{\tau}^{2}\Delta}. \label{eq:EstimatorConstants}
\end{align}

For a given \(Y=y\), let \(x_{\tau}^{(k)} \sim p_{\tau}(x_{\tau}\mid y) \, \forall k \in [\mathcal{K}]\). We can construct a single-sample estimator, 
\begin{equation}
    \widehat{\mu}(\tau,   x^{(k)}_{\tau}, y)=a(\tau)x_{\tau}^{(k)}+b_{\Delta}(\tau)D_{\theta^{*}}(\tau, x^{(k)}_{\tau}, y), \label{eq:SingleSampleEstimator}
\end{equation}
where \(x_{\tau}^{(k)}\) enters as a linear correction to the learned denoiser, constructed in a way that preserves unbiasedness in the ideal case of \eqref{eq:IdealDriftEstimator}. By averaging over the \(\mathcal{K}\) samples \(x_{\tau}^{(k)}\), we can estimate the drift with,
\begin{equation}
    \bar{\mu}(\tau, y) = \frac{1}{\mathcal{K}}\sum_{k\in[\mathcal{K}]}\widehat{\mu}(\tau, x^{(k)}_{\tau}, y).\label{eq:GeneralISDriftEstimation}
\end{equation}
For sufficiently large \(\mathcal{K}\), the estimator \(\bar{\mu}(\tau,y)\) removes the linear dependence of the rescaled denoiser on \(X_{\tau}\) in \eqref{eq:IdealDriftEstimator}, and can be cast as a lower-variance alternative to \(\hat{\mu}(\tau, x_{\tau}^{(k)}, y)\).

We emphasise that the Euler-Maruyama approximation used to derive \eqref{eq:GeneralISDriftEstimation} is not imposed during training, but only at the stage of drift estimation.

\subsection{Choice of diffusion time for drift estimator}\label{section:ChoiceDiffusionTime}

To implement the estimator in \eqref{eq:GeneralISDriftEstimation}, a value of \(\tau \in [0, 1]\) must be chosen, and \(\mathcal{K}\) samples \(x^{(k)}_{\tau} \sim p_{\tau}(x_{\tau} \mid Y=y)\) must be drawn. One expects there to exist an intermediate noise level \(0 < \tau < 1\) for which the denoising problem is well conditioned and the drift approximation error in \(\bar{\mu}\) is low. 

We propose setting \(\tau=1\) as a convenient and empirically robust choice, as motivated in the remainder of this section. At \(\tau=1\), one should first sample \(\mathcal{K}\) standard normal random variables \(x^{(k)}_{1}\sim \mathcal{N}(0, I)\). The drift in \eqref{eq:SingleSampleEstimator} is then,
\begin{equation*}
    \widehat{\mu}(1, x^{(k)}_{1},y) = 
    a(1)x^{(k)}_{1}+b_{\Delta}(1)D_{\theta^{*}}(1, x^{(k)}_{1}, y),
\end{equation*}
Averaging over all \(\mathcal{K}\) samples yields \(\bar{\mu}(y):=\bar{\mu}(1, y)\),
\begin{equation}
    \bar{\mu}(y) = \frac{1}{\mathcal{K}}\sum_{k\in[\mathcal{K}]}\widehat{\mu}(1, x^{(k)}_{1}, y). \label{eq:AvgDriftEstimator}
\end{equation}

To justify why \(\tau=1\) is appropriate, we empirically investigate the dependence of the estimator \(\bar{\mu}(\tau, y)\) on \(\tau\). In principle, samples \(x^{(k)}_{\tau}\sim p_{\tau}(x_{\tau}\mid y)\) can be drawn by sampling \((x_{0}, y) \sim p_{0}(x_{0},  y)\) from the dataset \(\mathcal{D}\) and applying the forward process \(q_{\tau}\) in \eqref{eq:VPSDETransition}. However, this approach is limited to values of \(y\) observed in the dataset and scales poorly with dimension. Therefore,  sampling \(x^{(k)}_{\tau}\) from \(p_{\tau}(x_{\tau}\mid y)\) for different values of \(\tau\) is implemented by simulating the reverse diffusion \eqref{eq:ConditionalReverseVPSDE}, see \cite{SongEtAl2020, KarrasEtAl2022}, for details on different numerical implementations. The VPSDE forward process in \eqref{eq:VPSDETransition} ensures \(X_{\tau}\simeq \mathcal{N}(0, I)\) for large \(\tau\). We can therefore sample \(x^{(k)}_{1}\sim\mathcal{N}(0, I)\) and run the reverse diffusion in \eqref{eq:ConditionalReverseVPSDE} to obtain approximate samples from \(p_{\tau}(x_{\tau} \mid y)\) for intermediate diffusion times.

During validation, we set \(\mathcal{K} = 10\) and allow oracle access to the true drift to select the model which minimises the \(\tau\)-averaged drift mean-squared error (MSE),
\begin{equation*}
    \widehat{\theta} = \arg\min_{\theta}\mathbbm{E}_{\tau}\mathbbm{E}\left[\norm{\bar{\mu}(\tau, Y)-\mu(Y)}_{2}^{2}\right],
\end{equation*}

We choose to test this approach using two representative drift functions,
\begin{align}
    \mu^{(1)}(y) &= -y + \sin(25y), \, y \in \mathbbm{R}\\
    \mu^{(2)}_{d}(y) &= -4a_{d} y_{d}^{3} -2b_{d}y_{d}, \, y \in \mathbbm{R}^{D}, \, d\in[D]\label{eq:DDimBiPotDrift}
\end{align}
where  \(a_{d}>0, b_{d}<0 \ \forall d \in [D]\). Given a set of \(N\) points uniformly spaced \(y_{n} \in \mathcal{R}\subset\mathbbm{R}^{D}\), we approximate the drift MSE as, 
\begin{equation}
    e^{2}(\tau) \approx \frac{1}{N}\sum_{n\in[N]}\left\|\mu(y_{n})-\overline{\mu}(\tau,y_{n})\right\|_{2}^{2}. \label{eq:DriftMSEFuncTau}
\end{equation}

Figure \ref{fig:TauQuadSinHF} shows the error \(e^{2}(\tau)\) when estimating \(\mu^{(1)}\) for \(\mathcal{K}\in\{1, 10, 100\}\). In all cases, the drift MSE diverges as \(\tau \rightarrow 0\), as the approximation error in the network \(D_{\theta^{*}}\) is expected to increase for smaller \(\tau\) \cite{SongEtAl2020}. The error is minimised for \(\tau \approx 0.2\), but as \(\mathcal{K}\) increases, the error curve becomes flat and less sensitive to \(\tau\), and larger values yield similar performance. Indeed, the bottom two panels in Figure \ref{fig:TauQuadSinHF} demonstrate accurate recovery of \(\mu^{(1)}\) for \(\tau \in \{0.2, 1\}\) when \(\mathcal{K}=100\).

Similarly, Figure \ref{fig:TauDDimsSep} illustrates the drift error \(e^{2}(\tau)\) when estimating \(\mu^{(2)}\) for \(D=8\) (top row) and \(D=12\) (bottom row), exhibiting behaviour similar to Figure \ref{fig:TauQuadSinHF} even in higher dimensions.

\begingroup
\centering
\setlength{\abovecaptionskip}{0pt}
\setlength{\belowcaptionskip}{0pt}
\setlength{\parskip}{0pt}
\begin{tabular}{cc}
\includegraphics[width=0.46\columnwidth]{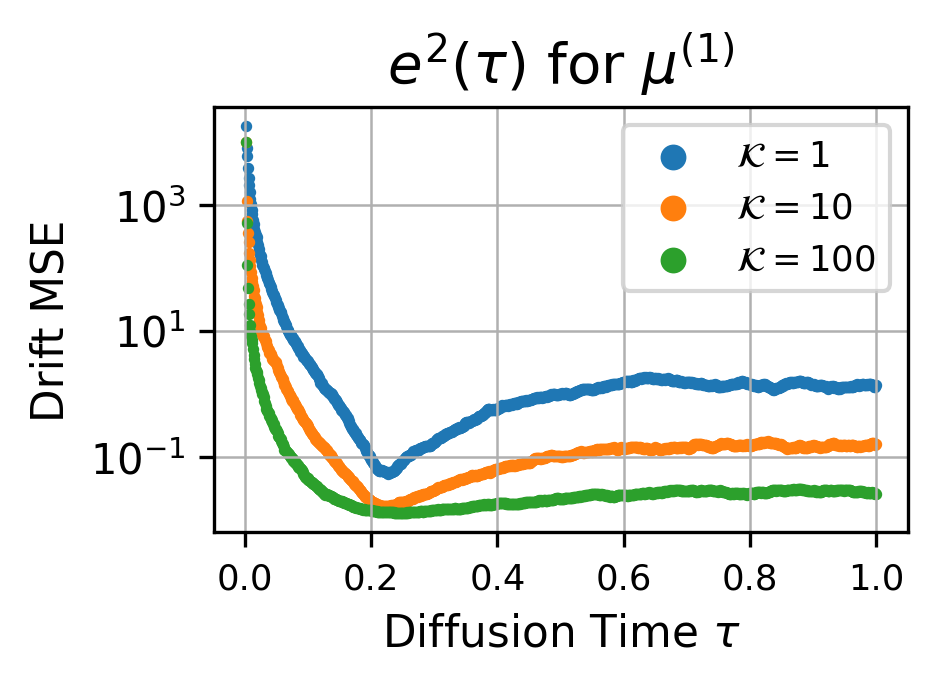} &
\includegraphics[width=0.46\columnwidth]{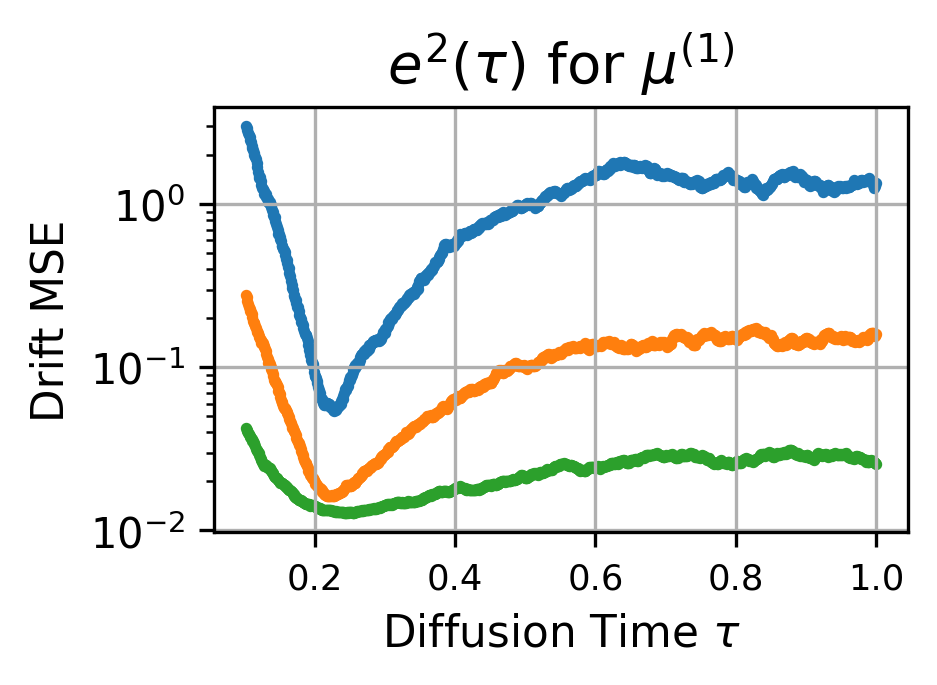} \\
\includegraphics[width=0.46\columnwidth]{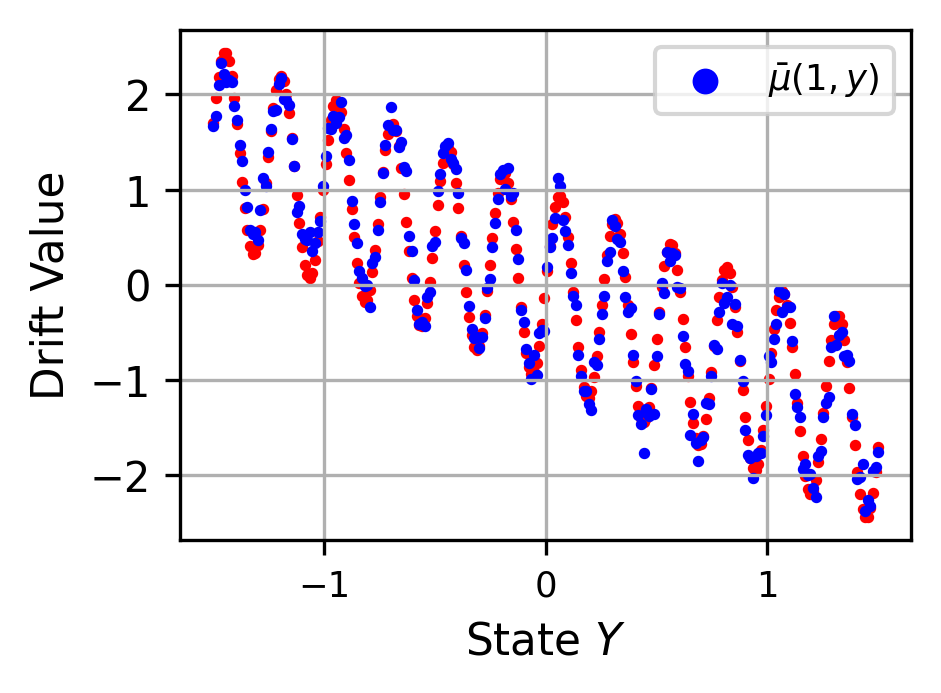} &
\includegraphics[width=0.46\columnwidth]{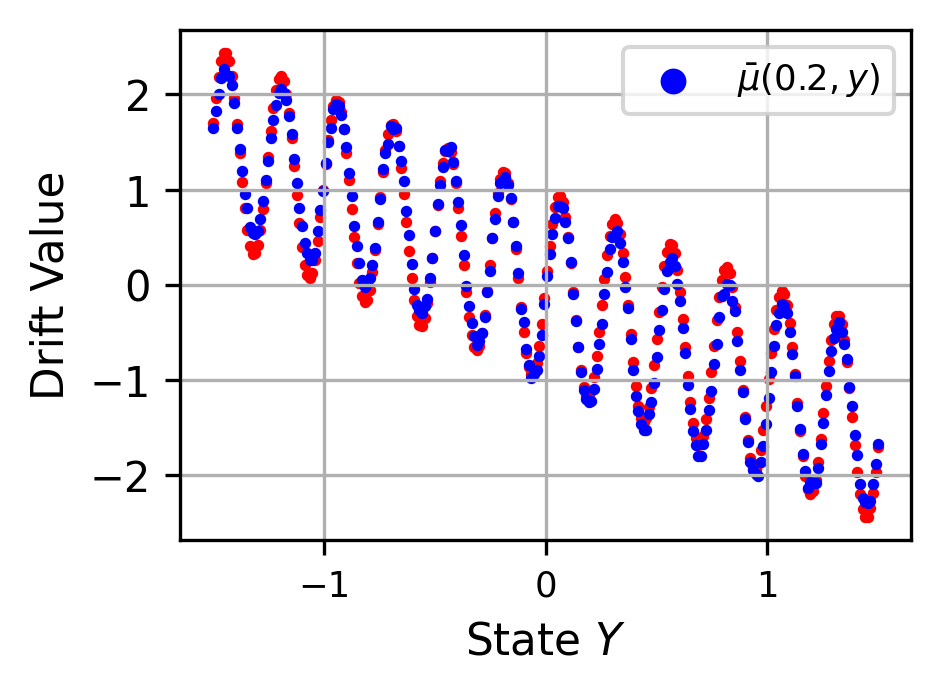} \\
\end{tabular}
\captionof{figure}{The top-left panel shows the drift error \(e^{2}(\tau)\) when estimating \(\mu^{(1)}\); the top-right panel restricts the range of \(\tau\) to \(\tau \in[0.15,1]\). The bottom row compares the \textcolor{red}{true} drift with \(\mathcolor{blue}{\bar{\mu}}(\tau, y)\) for \(\tau=1\) (left) and optimal \(\tau\approx0.2\) (right) for \(\mathcal{K}=100\). Positions \(Y\) are uniformly spaced between \([-1.5, 1.5]\), covering \(99\%\) of the training distribution.}
\label{fig:TauQuadSinHF}
\endgroup

\begingroup
\centering
\setlength{\abovecaptionskip}{0pt}
\setlength{\belowcaptionskip}{0pt}
\setlength{\parskip}{0pt}
\begin{tabular}{cc}
\includegraphics[width=0.46\columnwidth]{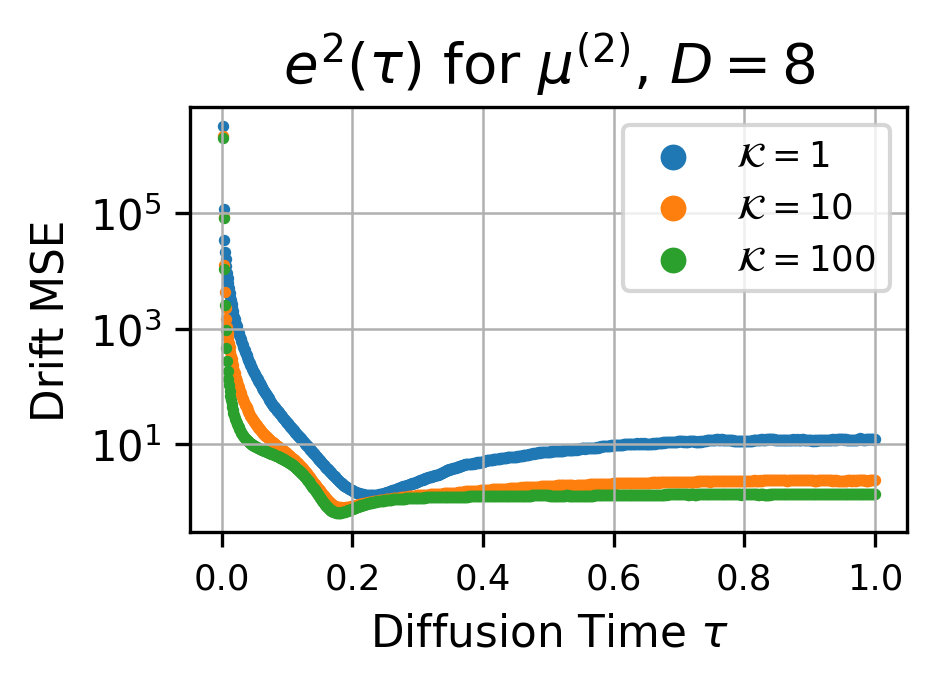} &
\includegraphics[width=0.46\columnwidth]{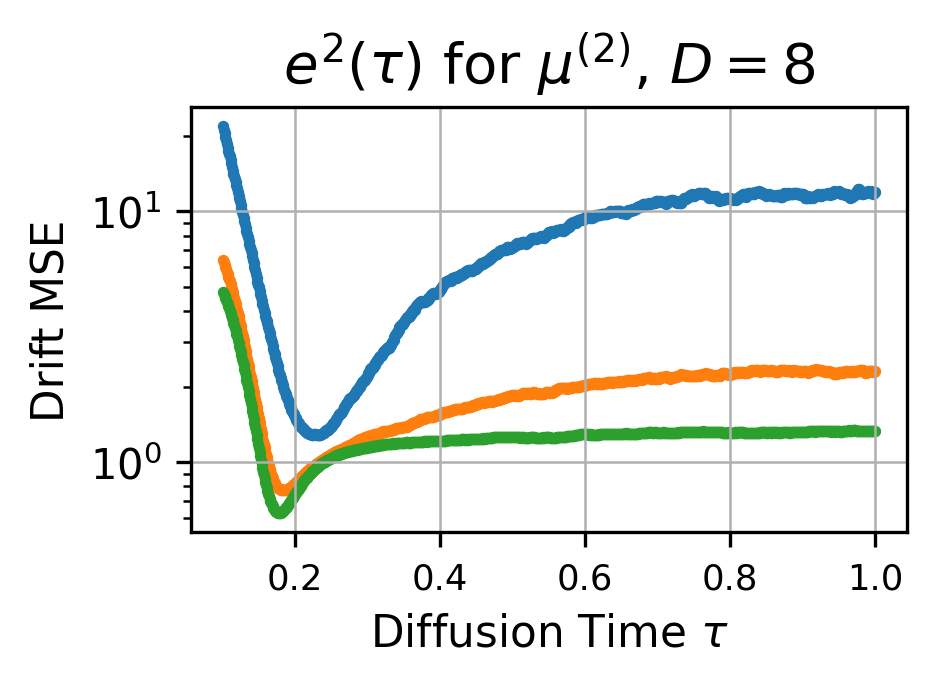} \\
\includegraphics[width=0.46\columnwidth]{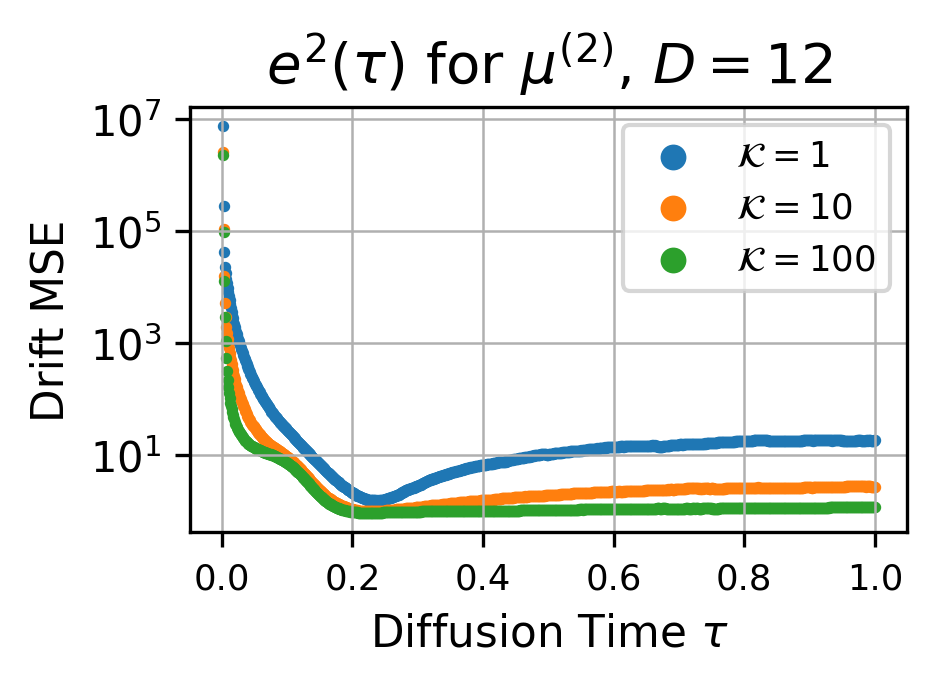} &
\includegraphics[width=0.46\columnwidth]{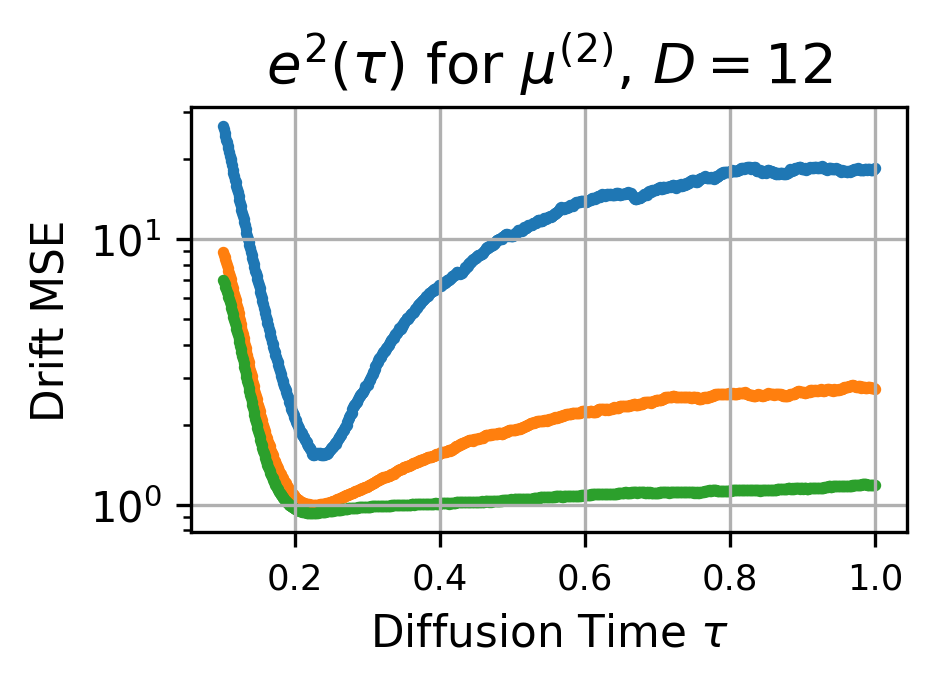} \\
\end{tabular}
\captionof{figure}{The left panels show the drift error \(e^{2}(\tau) \forall \tau \in[0,1]\); the right panels restrict the range of \(\tau\) to \(\tau \in[0.15,1]\). The top row shows the drift error \(e^{2}(\tau)\) \(\mu^{(2)}\) with \(D=8\); the bottom row shows the same for \(D=12\). }
\label{fig:TauDDimsSep}
\endgroup

We emphasise here that there is no closed form solution to selecting the optimal \(\tau\) since the approximation error in \(D_{\theta^{*}}\) is architecture and data dependent. Further, the choice of forward noise schedule will impact the dependence of the estimator on \(\tau\), see Appendix \ref{appendix:VESDEAblation} for a comparison with the Variance Exploding SDE. Therefore, there is no guarantee that a specific \(\tau\) is optimal across architectures and noise schedules. Moreover, the computational cost of simulating the reverse diffusion process is high. In contrast, sampling from the approximate terminal distribution \(X_{1}\sim\mathcal{N}(0, I)\) is cheap, and for large \(\mathcal{K}\), the error at \(\tau=1\) is similar to the optimal error.  As such, we suggest using \(\tau=1\) to implement our estimator \eqref{eq:AvgDriftEstimator}, and the results in Section \ref{section:FullDataResults} confirm that even with this choice of \(\tau\), we achieve competitive performance.

\section{Experiments}
\subsection{Evaluation Criterion}
We evaluate the performance of a drift estimator \(\widetilde{\mu}\) using the time-integrated mean-squared error,
\begin{equation}
    E^{(\mu)}_{t} = \frac{1}{t}\int_{0}^{t}\mathbbm{E}\norm{\widetilde{\mu}(Y_{s}) - {\mu}(Y_{s})}_{2}^{2}ds\label{eq:ProbWeightedMSE}
\end{equation}
where the expectation is over the law of the process \((Y_{s})_{s}\).

In practice, we approximate \eqref{eq:ProbWeightedMSE} in the following way. Let \(\{Y^{(i)}\}_{i}, i \in [\mathcal{I}]\) be a set of i.i.d trajectories from \eqref{eq:BSDE} with \(\mathcal{J}\) discrete observations. We approximate \eqref{eq:ProbWeightedMSE} using \( \forall j \in [\mathcal{J}]\),
\begin{equation*}
    E^{(\mu)}_{j\Delta} := \frac{1}{j\mathcal{I}}\sum_{i\in[\mathcal{I}]}\sum_{k\in[j]}\norm{\widetilde{\mu}(Y^{(i)}_{t_{k}}) - {\mu}(Y^{(i)}_{t_{k}})}_{2}^{2},
\end{equation*}
where \(\mathcal{I}, \mathcal{J}\) denote the number of trajectories and observations, respectively, used for error evaluation. In contrast, \(I, J\) denote the number of training trajectories and observations, respectively.

\subsection{Implementation}\label{section:Implementation}

In all experiments, training datasets consist of \(I = 10^{3}\) sample paths generated from an Euler discretisation of  \eqref{eq:BSDE} with \(\sigma=1\) for \(t\in [0, 1]\) at a frequency of \(\Delta = 256^{-1}\). All drift estimators are fitted with fixed random seeds for reproducibility. In-sample errors are computed by simulating \(\mathcal{I}=10^3\) new trajectories over the in-sample horizon \(t\in[0,1]\) with \(\Delta=256^{-1}\), while out-of-sample errors are computed over \(t\in[0, 20]\). We will denote by \(\mathrm{DN}\) the denoising estimator \(\bar{\mu}\) in \eqref{eq:AvgDriftEstimator} which uses the trained network \(D_{\theta^{*}}\), see Appendix \ref{appendix:ModelArchitecture} for architectural details.

For one dimensional drift functions, we compare \(\mathrm{DN}\) against three nonparametric estimators used in drift estimation under i.i.d trajectories. Specifically, we implement the IID Nadaraya–Watson estimator \cite{MarieEtAl2021}, a kernel-based estimator obtained by averaging increments across independent trajectories; the IID Ridge \cite{DenisEtAl2021}, a spline-based least-squares estimator using B-spline bases with ridge regularization, and the IID Hermite projection \cite{ComteGenonCatalot2020}, which estimates the drift via least-squares projection onto a finite-dimensional Hermite basis, see Appendix \ref{appendix:Baselines} for more details. 

In higher dimensions, the computational and data requirements of the Ridge and Hermite estimators of \cite{DenisEtAl2021} and \cite{ComteGenonCatalot2020} grow exponentially with \(D\) \cite{ZhaoLiuHoffmann2025}, and kernel-based estimators such as the Nadaraya Watson deteriorate rapidly, as seen in Appendix \ref{appendix:KernelEstimators}. We therefore omit them from the main text when estimating high dimensional drift functions. 

To benchmark \(\mathrm{DN}\) in high dimensions, we implement the proposed estimator in \cite{ZhaoLiuHoffmann2025}, and refer to it as \(\mathrm{FC}\). The estimator \(\mathrm{FC}\) is the output of a feed-forward ReLU network trained to regress the scaled increments \(\Delta^{-1}X_{0}\) from the states \(Y\) using \eqref{eq:RegressionOfDrift}, corresponding to a standard regression objective, without any denoising mechanism. We follow \cite{ZhaoLiuHoffmann2025} in imposing sparsity and weight-clamping constraints on the network parameters during training. These constraints restrict the hypothesis class to separable drifts and are part of the estimator definition in \cite{ZhaoLiuHoffmann2025}. For consistency, all neural-network based estimators are trained to target \(\Delta^{-1}X_{0}\).

Our denoising network \(D_{\theta^{*}}\) includes a learnable embedding applied to states \(Y\), referred to as \emph{MLPStateMapper}, which combines polynomial and Fourier features and feeds into the remaining network. It also includes a parallel one-dimensional convolutional module to capture local spatial structure in \((Y, X_{\tau})\), see Appendix \ref{appendix:ModelArchitecture} for more details. As the performance of \(\mathrm{DN}\) depends on both denoising-based training and architectural inductive biases in \(D_{\theta^{*}}\), we introduce three control baselines to attempt to disentangle their contributions.

We denote by \(\mathrm{FC}^{+}\) the \(\mathrm{FC}\) estimator with its parameter count matched to \(D_{\theta^{*}}\), and remove the sparsity and weight-clamping restrictions in \cite{ZhaoLiuHoffmann2025}. We further introduce \(\mathrm{FC}^{+}\)-Conv, which augments the fully connected layers with a similar parallel convolutional module as \(D_{\theta^{*}}\). Both \(\mathrm{FC}^{+}, \mathrm{FC}^{+}\)-Conv are trained on the same objective as \(\mathrm{FC}\), i.e., the standard regression objective. Therefore, estimators \(\mathrm{FC}^{+}, \mathrm{FC}^{+}\)-Conv help to isolate the effect of denoising by controlling for differences in model capacity and inductive bias. Finally, we define \(\mathrm{DN}\)-Lin as a variant of \(\mathrm{DN}\) in which the \textit{MLPStateMapper} in \(D_{\theta^{*}}\) is replaced by a fully connected mapping, helping to isolate the effect of learned state embeddings in drift estimation. 

All estimators are selected by minimising the oracle error \(E^{(\mu)}_{j\Delta}\) at the in-sample terminal time \(j\Delta = 1\). For the one-dimensional benchmarks, hyperparameters are chosen via grid search, see Appendix \ref{appendix:FullDataAppendix}. For neural network based estimators, (\(\mathrm{FC}\), \(\mathrm{FC}^{+}\), \(\mathrm{FC}^{+}\)-Conv, \(\mathrm{DN}\)-Lin and \(\mathrm{DN}\)), we evaluate the error \(E^{(\mu)}_{1}\) on \(\mathcal{I}=50\) held-out trajectories at every epoch and terminate training when \(E^{(\mu)}_{1}\) has not improved over 100 epochs, see Appendix \ref{appendix:FullDataAppendix} for more details. Sensitivity to reduced training dataset size and to sampling frequency \(\Delta\) are explored in Appendix \ref{appendix:LessDataAppendix} and \ref{appendix:SamplingFrequency}, respectively, and Appendix \ref{appendix:FeasibleValidation} provides a performance comparison when we use a feasible model selection objective for \(\mathrm{DN}, \mathrm{DN}\)-Lin.

\subsection{Test Drift Functions}\label{section:DriftFunctions}

For \(D=1\), we consider the following one-dimensional drift functions,
\begin{align*}   
    \mu_{1}(y) &= -\sin(2y)\log(1+5|y|), \\
    \mu_{2}(y) &= -y + \sin(25y), \\
    \mu_{3}(y) &= -y^{3} + y.
\end{align*}
To consider high-dimensional drifts derived from scalar potentials, we use the following parametric family,
\begin{align*}
    \mu_{4, d}(y) = &-4a_{d} y_{d}^{3} -2b_{d}y_{d} \\
    &-\frac{c}{2}\,\psi_{d}'(y_{d})\big[\psi_{d-1}(y_{d-1})+\psi_{d+1}(y_{d+1})\big],
\nonumber
\end{align*}
where the coefficients satisfy \(a_{d}>0, b_{d}<0\), and \(c\) is a coupling constant. For \(c=0\), \(\mu_{4}\) collapses to a separable bi-stable potential with wells at \(y_{*,d}=\pm \sqrt{-\tfrac{b_{d}}{2a_{d}}}\). For \(c>0\), coordinate interaction is introduced via  \(\psi_{d}(y_{d})=\exp\!\big(-\tfrac{(y_{d}^{2}-y_{*, d}^{2})^{2}}{2s_{d}^{2}}\big)\), and \(s_{d}=\sqrt{c}y_{*, d}\). 

To consider non-potential and chaotic dynamics, we use the following drift, \(\forall d \in \{0, ..., D-1\}\) and \(F>0\),
\begin{equation*}
    \mu_{5, d}(y) = (y_{d+1} - y_{d-2})y_{d-1} - y_{d}+F,
\end{equation*}
where indices are taken modulo \(D\). For \(F<1\), \(\mu_{5}\) defines the Lorenz-96 system which exhibits stable behaviour, whereas chaotic behaviour emerges for standard choices such as \(F=8\) \cite{Lorenz1996}.

\subsection{Results}\label{section:FullDataResults}
\subsubsection{In Sample Results}

Table~\ref{tab:FinalTimeFullDataTrueLawOneDimension} reports final-time in-sample errors \(E^{(\mu)}_{1}\) for the one-dimensional benchmarks. The denoising-based estimator attains errors of the same order of magnitude as classical nonparametric methods, and Figure \ref{fig:OneDimensionalDenoisingDrift} illustrates that it accurately recovers the structure of the drift.

\begingroup
\refstepcounter{table} {\itshape Table \thetable.}{ In-sample drift errors \(E^{(\mu)}_{1}\), for \(\mu_1, \mu_{2}, \mu_3\). Best estimator is starred, second-best is underlined.}
\par\vspace{0.3em}
\centering
\renewcommand{\arraystretch}{0.95}
\setlength{\tabcolsep}{3pt}
\centerline{
\begin{tabular}{ c c c c c }
\toprule
\(E^{(\mu)}_{1}\)  & \(\mathrm{DN}\) & NW & Hermite & Ridge  \\
\midrule
\(\mu_1\) & \underline{0.02127} &0.04564 & \({0.007621}^{*}\) & 0.02382 \\
\(\mu_2\) & \({0.01870}^{*}\) &0.1320 & 0.5049&\underline{0.0574} \\
\(\mu_3\) & \underline{0.00754} &0.01464 & \({0.005082}^{*}\) & 0.1739 \\
\bottomrule
\end{tabular}
}
\label{tab:FinalTimeFullDataTrueLawOneDimension}
\endgroup

\begin{minipage}{\columnwidth}
\centering
\setlength{\abovecaptionskip}{0pt}
\setlength{\belowcaptionskip}{0pt}
\setlength{\parskip}{0pt}

\makebox[\columnwidth][l]{%
  \hspace*{-0.07\columnwidth}%
  \begin{tabular}{ccc}
    \includegraphics[width=0.32\columnwidth]{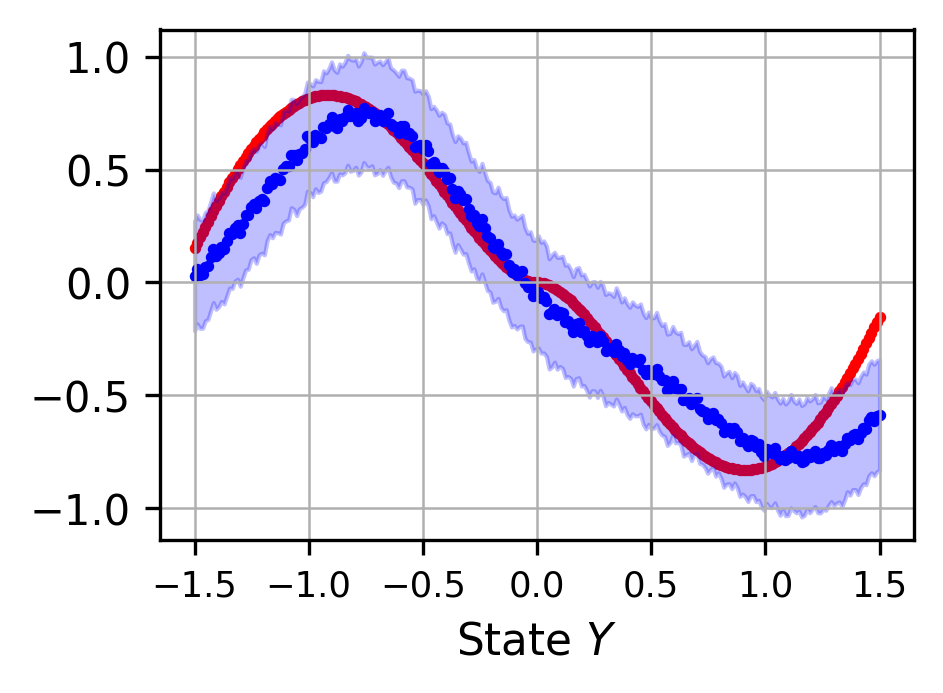} &
    \includegraphics[width=0.32\columnwidth]{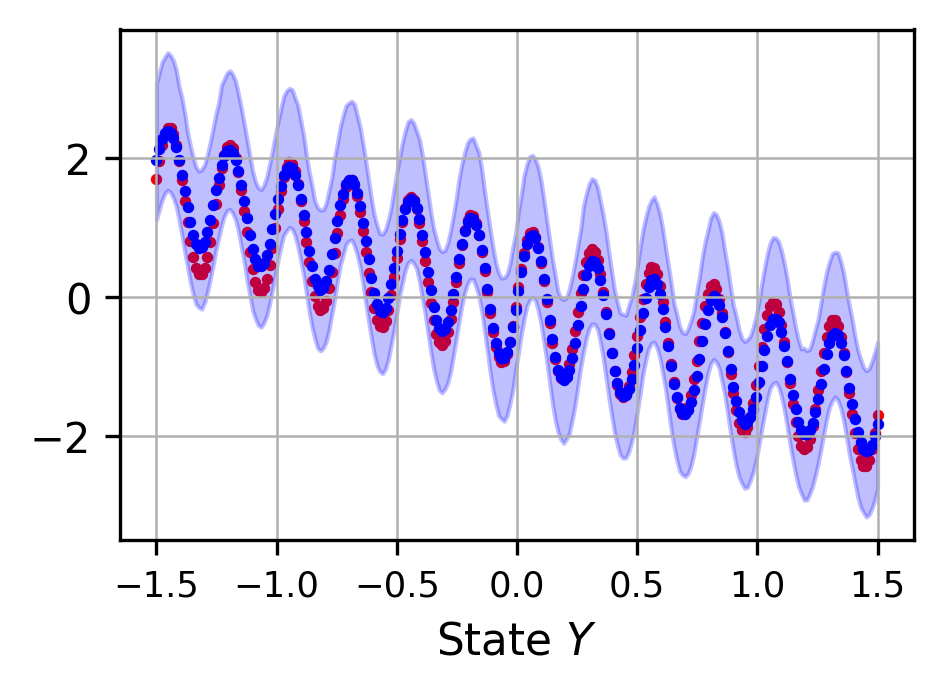} &
    \includegraphics[width=0.32\columnwidth]{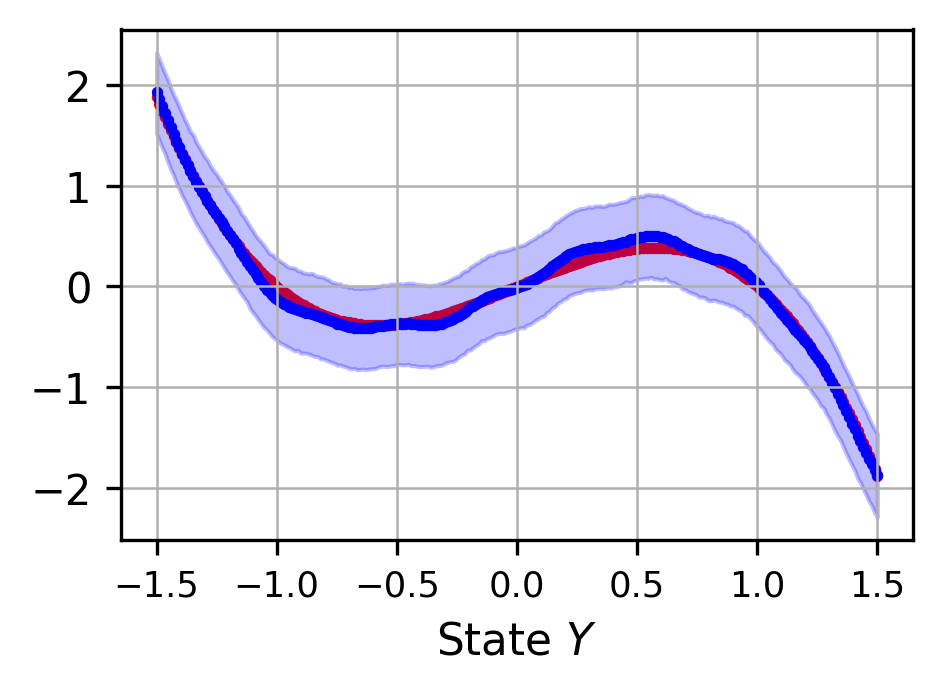}
  \end{tabular}
}
\captionof{figure}{Panels (left to right) correspond to \(\mu_{1}, \mu_{2}, \mu_{3}\). \textcolor{red}{True} drift against \textcolor{blue}{\(\mathrm{DN}\)} drift. Shaded region shows \(10-90\%\) quantile envelope over \(\mathcal{K}\). State values cover \(99\%\) of the training distribution.}
\label{fig:OneDimensionalDenoisingDrift}
\end{minipage}

In higher dimensions, Table~\ref{tab:FinalTimeFullDataTrueLawBiPotDDims} shows that the denoising estimator \(\mathrm{DN}\) achieves the lowest in-sample error when estimating the bistable drift \(\mu_{4}\) across both separable (\(c=0\)) and strongly coupled (\(c=20\)) regimes. In the separable case, (\(\mathrm{DN}\)-Lin) performs poorly in relation to \(\mathrm{DN}\), suggesting that denoising-based drift estimation depends on an inductive bias aligned with the drift structure in order to accurately learn the drift over the in-sample distribution.

\begingroup
\refstepcounter{table}{\itshape Table \thetable.}{ In-sample drift errors \(E^{(\mu)}_{1}\), for \(\mu_4\). Best estimator is starred, second-best is underlined.}
\par\vspace{0.3em}
\centering
\setlength{\tabcolsep}{3pt}
\renewcommand{\arraystretch}{0.95}
\centerline{
\begin{tabular}{c c c c c c c}
\toprule
$c$ & $D$ & \(\mathrm{DN}\) & \(\mathrm{DN}\)-Lin & \(\mathrm{FC}^{+}\)-Conv & \(\mathrm{FC}^{+}\) & \(\mathrm{FC}\) \\
\midrule
\multirow{2}{*}{0}
 & 8  & 1.047\(^{*}\)&6.941&\underline{3.898}&8.865&7.687 \\
 & 12 & 1.681\(^{*}\)&11.294&\underline{6.316}&14.276&14.157 \\
 \midrule
\multirow{2}{*}{20}
 & 8  & 8.490\(^{*}\)&10.335&\underline{9.048}&16.287&35.365\\
 & 12 & 10.660\(^{*}\)&\underline{13.648}&14.076&29.528&46.166 \\
\bottomrule
\end{tabular}
}
\label{tab:FinalTimeFullDataTrueLawBiPotDDims}
\endgroup

Figure~\ref{fig:DDimsSelectedDims} overlays the true and estimated drift \(\mathrm{DN}\) for selected dimensions of \(\mu_{4}\), illustrating accurate recovery in intermediate dimensions and expected degradation in the highest-index dimension.

\begingroup
\centering
\setlength{\abovecaptionskip}{0pt}
\setlength{\belowcaptionskip}{0pt}
\setlength{\parskip}{0pt}
\includegraphics[width=0.46\columnwidth]{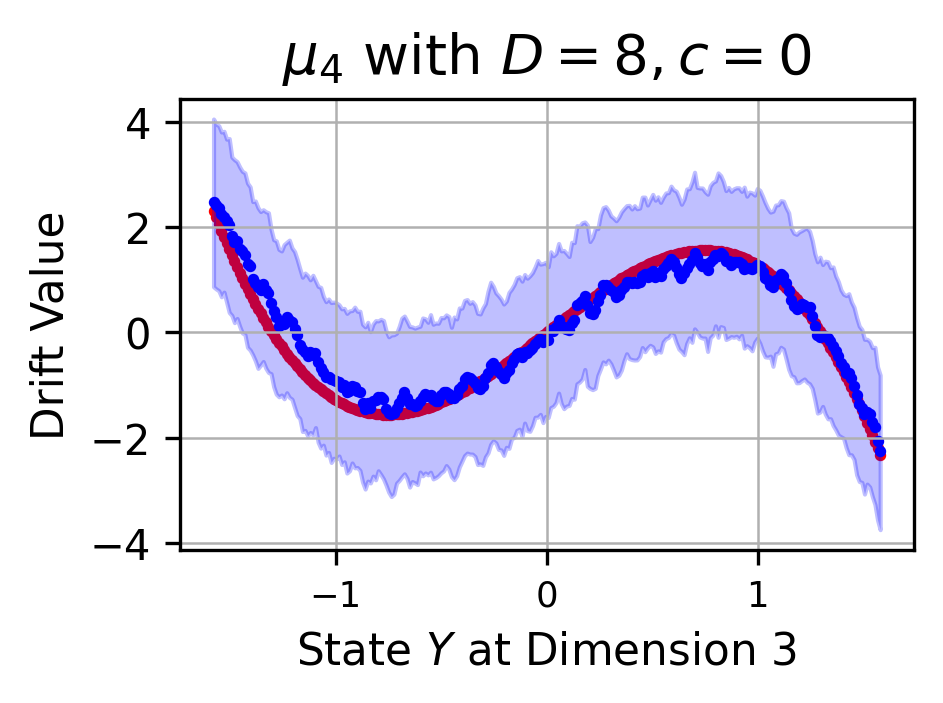} \hspace{0.02\columnwidth}%
\includegraphics[width=0.46\columnwidth]{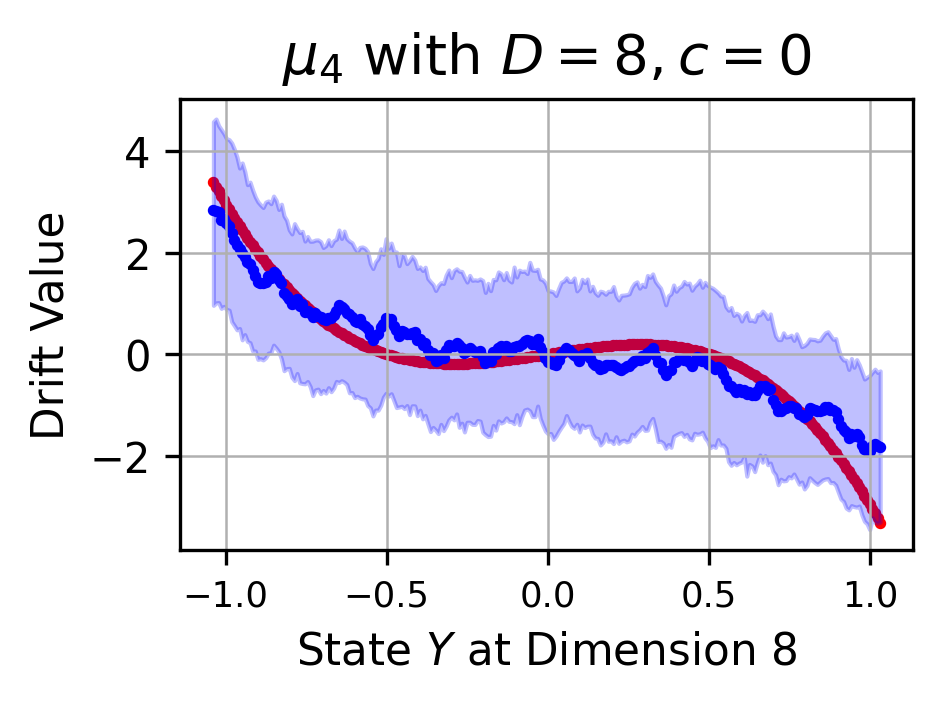}  \\
\includegraphics[width=0.46\columnwidth]{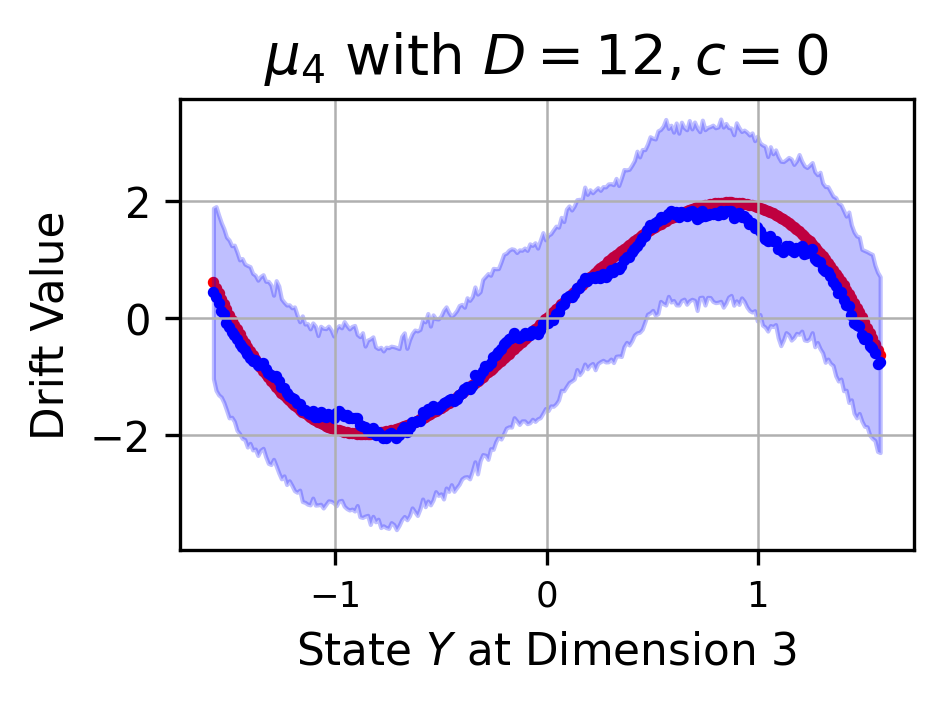}  \hspace{0.02\columnwidth}%
\includegraphics[width=0.46\columnwidth]{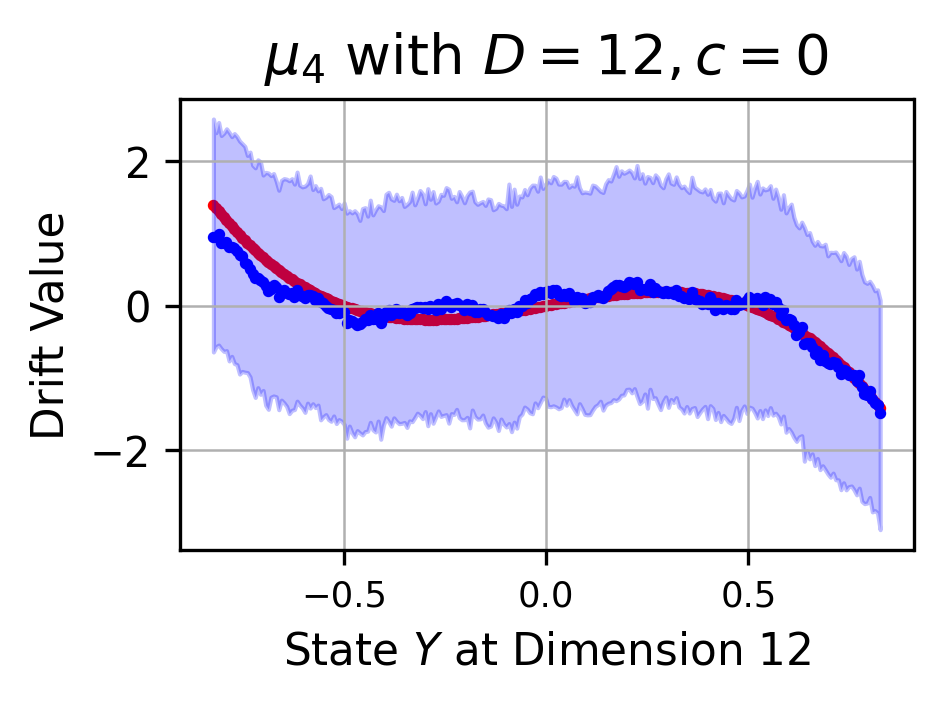}  \\
\captionof{figure}{\textcolor{red}{True} drift against the \textcolor{blue}{\(\mathrm{DN}\)} drift for \(\mu_{4}, c=0\). State values cover \(99\%\) of the training distribution in each dimension. Top panels show dimensions \(3\) (left) and \(8\) (right) for \(D=8\). Bottom panels show dimensions \(3\) (left) and \(12\) (right) for \(D=12\). Shaded region denotes \(10-90\%\) quantile envelope over \(\mathcal{K}\).}
\label{fig:DDimsSelectedDims}
\endgroup

Table~\ref{tab:FinalTimeFullDataTrueLawDLnz} reports in-sample results for the Lorenz 96 drift \(\mu_{5}\). Performance is dominated by architectural inductive bias, as estimators with convolutional architectures (\(\mathrm{DN}\), \(\mathrm{DN}\)-Lin, \(\mathrm{FC}^{+}\)-Conv) achieve similar accuracy while fully connected baselines perform poorly.

\begingroup
\refstepcounter{table} {\itshape Table \thetable.}{ In-sample drift errors \(E^{(\mu)}_{1}\), for \(\mu_5\).}
\par\vspace{0.3em}
\centering
\renewcommand{\arraystretch}{0.95}
\setlength{\tabcolsep}{3pt}
\centerline{
\begin{tabular}{c  c  c  c  c  c  c}
\toprule
\(F\) & \(D\) & \(\mathrm{DN}\) & \(\mathrm{DN}\)-Lin & \(\mathrm{FC}^{+}\)-Conv & \(\mathrm{FC}^{+}\) & \(\mathrm{FC}\) \\
\midrule
\multirow{2}{*}{\(\frac{1}{2}\)}
   & 20 & 4.870&3.127\(^{*}\)&\underline{3.733}&18.554&49.862\\
   & 40 & \underline{9.596}&7.766\(^{*}\)&11.940&44.764&62.988\\
\midrule
\multirow{2}{*}{\(8\)}
   & 20 & \underline{9.542}&5.725\(^{*}\)&11.486&127.103&926.707 \\
   & 40 & 17.898&12.203\(^{*}\)&\underline{16.306}&1263.290&2583.505\\
\bottomrule
\end{tabular}
}
\label{tab:FinalTimeFullDataTrueLawDLnz}
\endgroup
\newpage

Overall, the results indicate that denoising-based estimators yield the strongest in-sample performance when paired with a network whose inductive bias is aligned with the underlying dynamics. However, since all networks are selected by oracle in-sample error during training, these results reflect best-case fitting and may not be indicative of the relative generalisation behaviour of denoising and standard regression estimators.

\subsubsection{Out of Sample Results}\label{section:OOSFullDataResults}

We evaluate out-of-sample (OOS) performance by simulating trajectories beyond the training horizon, where the system in \eqref{eq:BSDE} evolves into regions of state space not observed during training.

Table~\ref{tab:OOSFinalTimeFullDataTrueLawBiPotDDims} reports final-time OOS errors for the bistable drift \(\mu_{4}\). In the non-separable regime (\(c=20\)), the denoising estimators (\(\mathrm{DN}, \mathrm{DN}\)-Lin) achieve the lowest OOS error compared to all standard regression baselines across dimensions. Figure~\ref{fig:DDimsFullDataOOSTrueLaw} shows that the OOS error of \(\mathrm{DN}\) grows more slowly over time, leading to increasing separation from \(\mathrm{FC}^{+}\)-Conv. This advantage persists even when the \(\mathrm{FC}^{+}\)-Conv baseline is augmented with the \textit{MLPStateMapper} state embedding (see Appendix~\ref{appendix:AdditionalDDims}), suggesting that the performance gap of \(\mathrm{DN}\) is not explained by architecture alone. 

\begingroup
\refstepcounter{table}{\itshape Table \thetable.}{ Final-time OOS drift errors \(E^{(\mu)}_{20}\), for \(\mu_4\).}
\par\vspace{0.3em}
\centering
\setlength{\tabcolsep}{3pt}
\renewcommand{\arraystretch}{0.95}
\centerline{
\begin{tabular}{c c c c c c c}
\toprule
$c$ & $D$ & \(\mathrm{DN}\) & \(\mathrm{DN}\)-Lin & \(\mathrm{FC}^{+}\)-Conv & \(\mathrm{FC}^{+}\) & \(\mathrm{FC}\) \\
\midrule
\multirow{2}{*}{0}
 & 8  &1.546\(^{*}\)&11.800&\underline{9.744}&14.129&14.365\\
 & 12 & 2.314\(^{*}\)&16.616&\underline{8.344}&21.523&23.658\\
 \midrule
\multirow{2}{*}{20}
 & 8  &50.919\(^{*}\)&\underline{67.297}&76.061&70.134&67.299\\
 & 12 & 81.721\(^{*}\)&\underline{115.750}&121.379&182.499&105.854\\
\bottomrule
\end{tabular}
}
\label{tab:OOSFinalTimeFullDataTrueLawBiPotDDims}
\endgroup

\begingroup
\centering
\setlength{\abovecaptionskip}{0pt}
\setlength{\belowcaptionskip}{0pt}
\setlength{\parskip}{0pt}
\includegraphics[width=0.48\columnwidth]{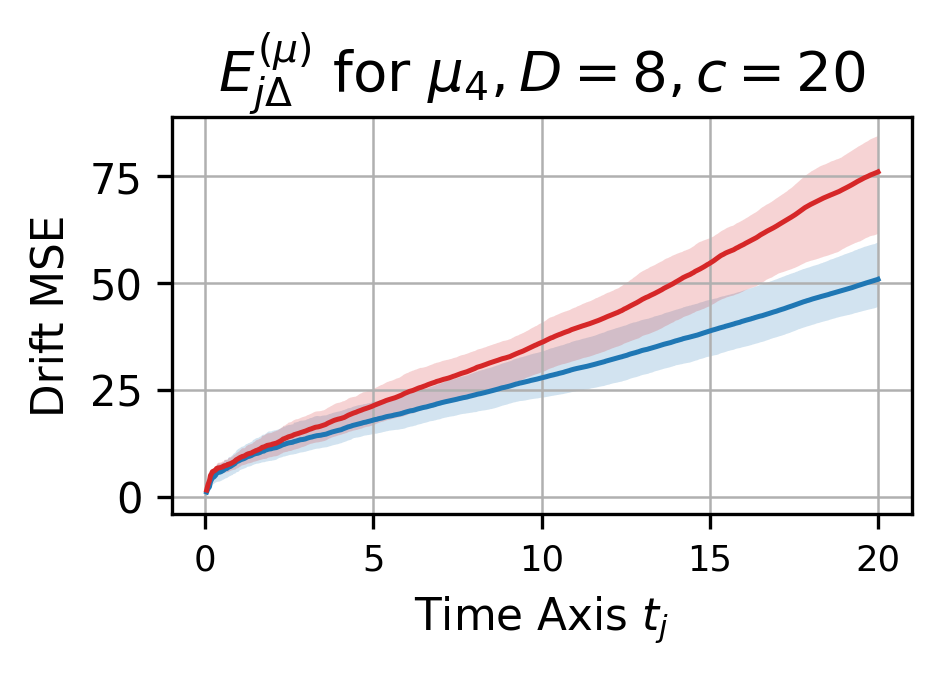}\hspace{0.5em}
\includegraphics[width=0.48\columnwidth]{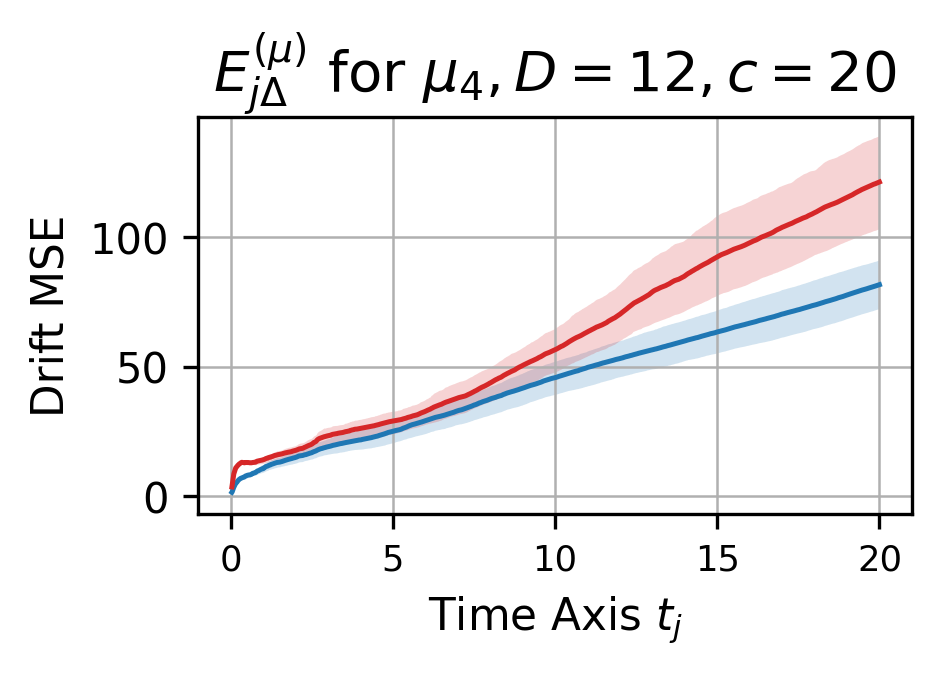}
\captionof{figure}{OOS \(E^{(\mu)}_{j\Delta}\) for \(\mu_{4}, c=20\), for \textcolor{mplblue}{\(\mathrm{DN}\)}, and \textcolor{red}{\(\mathrm{FC}^{+}\)-Conv}. Shaded region shows  \(10-90\%\) quantile envelope over test trajectories \(\mathcal{I}\).}
\label{fig:DDimsFullDataOOSTrueLaw}
\endgroup

Table~\ref{tab:OOSFinalTimeFullDataTrueLawDLnz} reports OOS results for the Lorenz 96 drift \(\mu_{5}\). In the stable regime (\(F=0.5\)), the estimators with convolutional networks achieve comparable OOS performance. In the chaotic regime (\(F=8\)), the denoising-based estimator \(\mathrm{DN}\)-Lin consistently attains lower OOS error than \(\mathrm{FC}^{+}\)-Conv across dimensions. Since both estimators share the same inductive bias, these results suggest a generalisation benefit attributable to the denoising training objective rather than architectural differences. The denoising variant \(\mathrm{DN}\) underperforms in this setting, reflecting misalignment of the \textit{MLPStateMapper} state embedding with the Lorenz 96 dynamics. Figure~\ref{fig:DLnzFullDataOOSTrueLaw} shows that while OOS errors stabilise over time, \(\mathrm{DN}\)-Lin plateaus at a lower error level than standard regression.

\begingroup
\refstepcounter{table} {\itshape Table \thetable.}{ Final-time OOS drift error \(E^{(\mu)}_{20}\), for \(\mu_5\).}
\par\vspace{0.3em}
\centering
\renewcommand{\arraystretch}{0.95}
\setlength{\tabcolsep}{3pt}
\centerline{
\begin{tabular}{c  c  c  c  c  c  c}
\toprule
\(F\) & \(D\) & \(\mathrm{DN}\) & \(\mathrm{DN}\)-Lin & \(\mathrm{FC}^{+}\)-Conv & \(\mathrm{FC}^{+}\) & \(\mathrm{FC}\) \\
\midrule
\multirow{2}{*}{\(\frac{1}{2}\)}
   & 20 & \underline{8.718}&12.024&7.920\(^{*}\)&44.376&56.041\\
   & 40 & \underline{12.198}&11.134\(^{*}\)&12.802&99.322&167.826\\
\midrule
\multirow{2}{*}{\(8\)}
   & 20 & 3973&172.786\(^{*}\)&\underline{226.361}&7457&8425\\
   & 40 & 1243&332.905\(^{*}\)&\underline{527.726}&16280&17656\\
\bottomrule
\end{tabular}
}
\label{tab:OOSFinalTimeFullDataTrueLawDLnz}
\endgroup

\begingroup
\centering
\setlength{\abovecaptionskip}{0pt}
\setlength{\belowcaptionskip}{0pt}
\setlength{\parskip}{0pt}
\includegraphics[width=0.48\columnwidth]{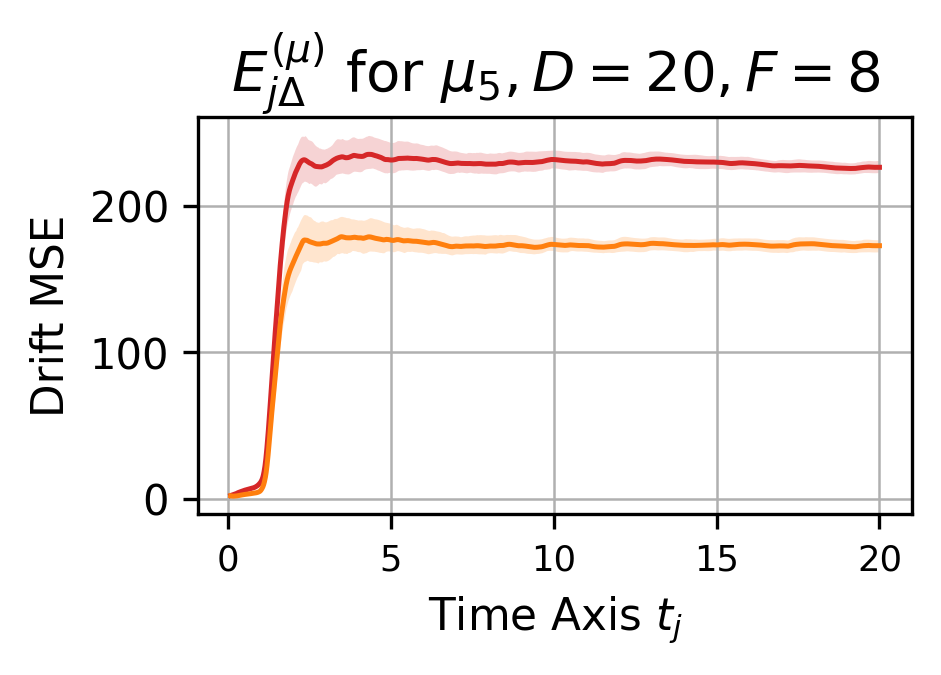}\hspace{0.5em}
\includegraphics[width=0.48\columnwidth]{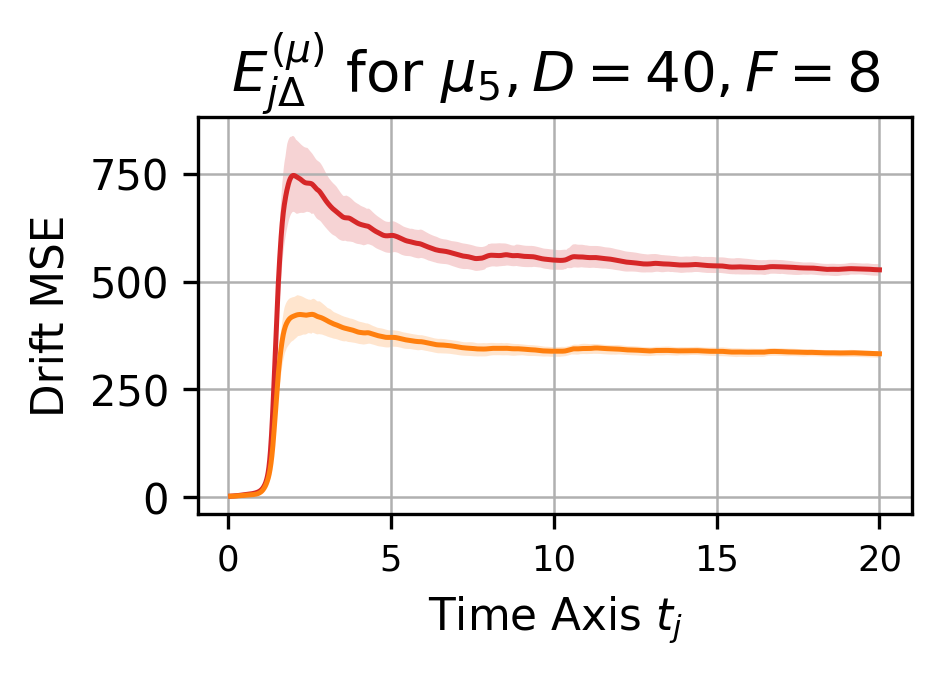}
\captionof{figure}{OOS \(E^{(\mu)}_{j\Delta}\)  for \(\mu_{5}, F=8\) comparing the \textcolor{orange}{\(\mathrm{DN}\)-Lin}, \textcolor{red}{\(\mathrm{FC}^{+}\)-Conv} estimators. Shaded region shows  \(10-90\%\) quantile envelope across test trajectories \(\mathcal{I}\). }
\label{fig:DLnzFullDataOOSTrueLaw}
\endgroup

Taken together, the OOS results for \(\mu_{4}\) and \(\mu_{5}\) suggest standard regression, when paired with an architecture aligned the underlying drift, can achieve strong in-sample performance, yet may generalise poorly out of sample in strongly coupled or chaotic dynamics. In these settings, the denoising estimator with an appropriate architecture provides improved generalisation, while remaining competitive when networks trained via standard regression already generalise well.

\section{Conclusion}
This paper adapts score-based conditional diffusion models to the problem of high-dimensional drift estimation from discretely observed trajectories. By introducing controlled Gaussian corruption to the observed increments, we train a neural network to learn the denoiser of each increment conditional on the previous state. Using the trained network, we derived an estimator for the drift through a first-order approximation of the data-generating process. Across a range of drift structures and dimensions, our results show that denoising can complement architectural inductive bias to improve out-of-sample robustness in settings with strong interactions or chaotic dynamics. In contrast, when standard regression already generalises stably out of sample, denoising can remain competitive. Several important questions remain open. These include developing a theoretical understanding for how different noise schedules affect the bias-variance properties of the estimator under specific architectures, as well as extending the estimator construction to incorporate higher order approximations of the denoising target.

\section{Impact Statement}
This paper presents work whose goal is to advance the field of Machine Learning. There are many potential societal consequences of our work, none which we feel must be specifically highlighted here.
\bibliographystyle{icml2026}
\bibliography{biblio.bib}
\newpage
\appendix
\section{Sensitivity to forward noising process}\label{appendix:VESDEAblation}
In this section, we motivate why the forward noising process \(q_{\tau}(X_{\tau}|X_{0})\) needs to be carefully chosen to ensure our denoising estimator \eqref{eq:AvgDriftEstimator} accurately recovers the drift function. 

Recall the transition kernel for the VPSDE noising process, 
\begin{equation*}
    q_{\tau}(x_{\tau}\mid X_{0}) = \mathcal{N}(x_{\tau}; \beta_{\tau}X_{0}, \sigma_{\tau}^{2}I),
\end{equation*}
where \(\beta_{\tau} = \exp(-0.25\tau^{2}(\gamma_{1}-\gamma_{0})-0.5\tau\gamma_{0})\) and \(\sigma_{\tau}^{2} = 1 - \beta_{\tau}^{2}\). \cite{SongEtAl2020} define an alternative noising process (the Variance Exploding SDE), which takes samples \(X_{0}\sim p_{0}\) and diffuses them with the forward transition density,
\begin{equation}
    q^{\text{VE}}_{\tau}(x_{\tau}\mid X_{0}) = \mathcal{N}(x_{\tau}; X_{0}, \phi_{\tau}^{2}I), \label{eq:VESDETransition}
\end{equation}
where \(\phi_{\tau} = \phi_{0}(\phi_{1}/\phi_{0})^{\tau}, \forall \tau \in [\epsilon, 1]\). As in the VPSDE process, \(\epsilon > 0\) is chosen to avoid vanishing variance as \(\tau\rightarrow 0\). However, in contrast to the VPSDE process, where noise variance \(\sigma_{\tau}^{2}\) increases smoothly and saturates at \(1\), the noise variance \(\phi_{\tau}^{2}\) in VESDE increases exponentially. Under uniform sampling in \(\tau\), this concentrates most training mass at low marginal noise levels. Consequently, the denoising estimator is trained predominantly in high signal-to-noise regimes, where learning the denoising target is hardest. By comparison, the VPSDE distributes training more evenly across signal-to-noise ratios. As such, given the same maximum noise levels \(\phi_{1} = \sigma_{1}\), the VESDE process spends more time in lower-noise regimes, where denoising is hardest.

We extend the experiments in Section \ref{section:ChoiceDiffusionTime} by diffusing the observed increments \(X_{0}\) with the VESDE noising kernel in \eqref{eq:VESDETransition}. We consider two VESDE variants. In VESDE-Matched, we set \(\phi_{1}=1\) to match the maximum marginal variance in the VPSDE. In VESDE-Larger, we increase the maximum marginal noise to \(\phi_{1}=15\) to compensate for the concentration of training mass at low noise levels under VESDE-Matched. In both cases, \(\phi_{0}=0.03\), and we set \(\epsilon = 6.5\times10^{-2}\) in VESDE-Matched and \(\epsilon = 5\times10^{-2}\) in VESDE-Larger in both training and evaluation to ensure the minimum marginal variance for both processes is identical to that of the VPSDE process in Section \ref{section:ChoiceDiffusionTime}.

\begingroup
\centering
\setlength{\abovecaptionskip}{0pt}
\setlength{\belowcaptionskip}{0pt}
\setlength{\parskip}{0pt}
\begin{tabular}{cc}
\includegraphics[width=0.46\columnwidth]{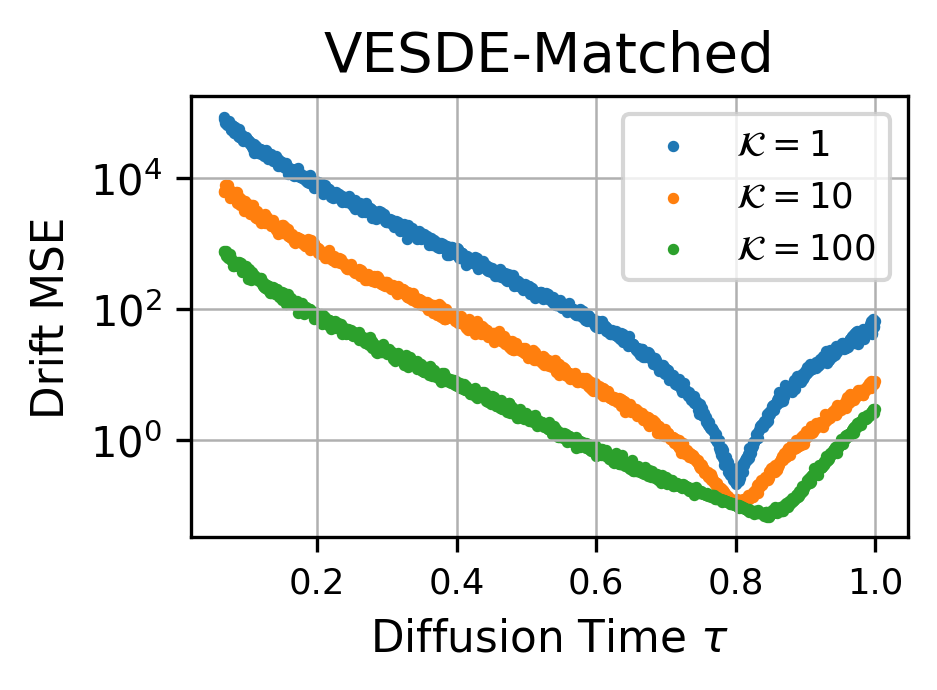} &
\includegraphics[width=0.46\columnwidth]{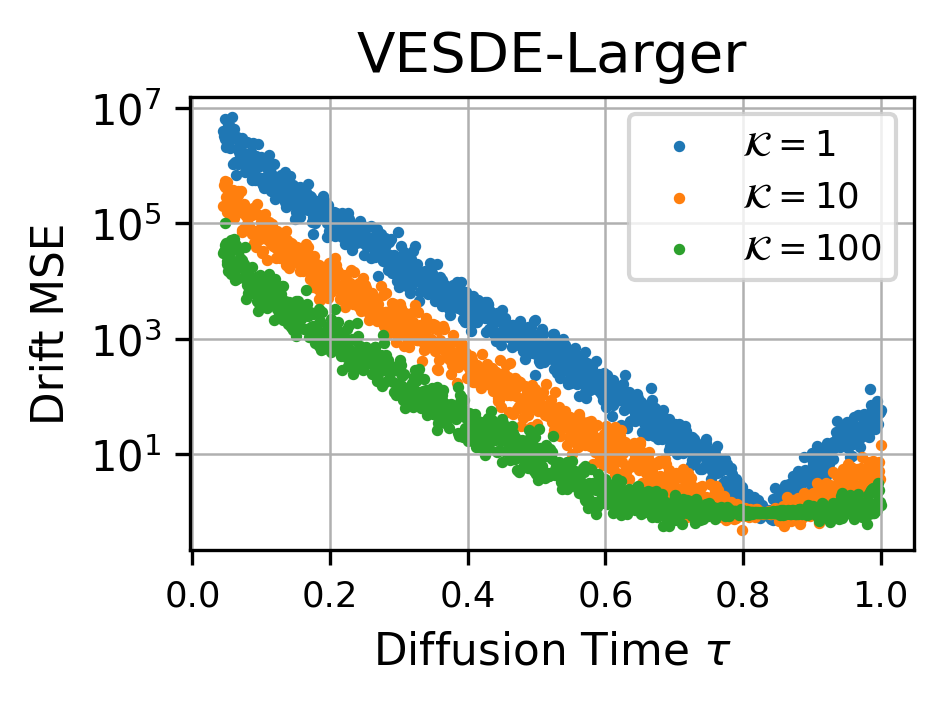} \\
\includegraphics[width=0.46\columnwidth]{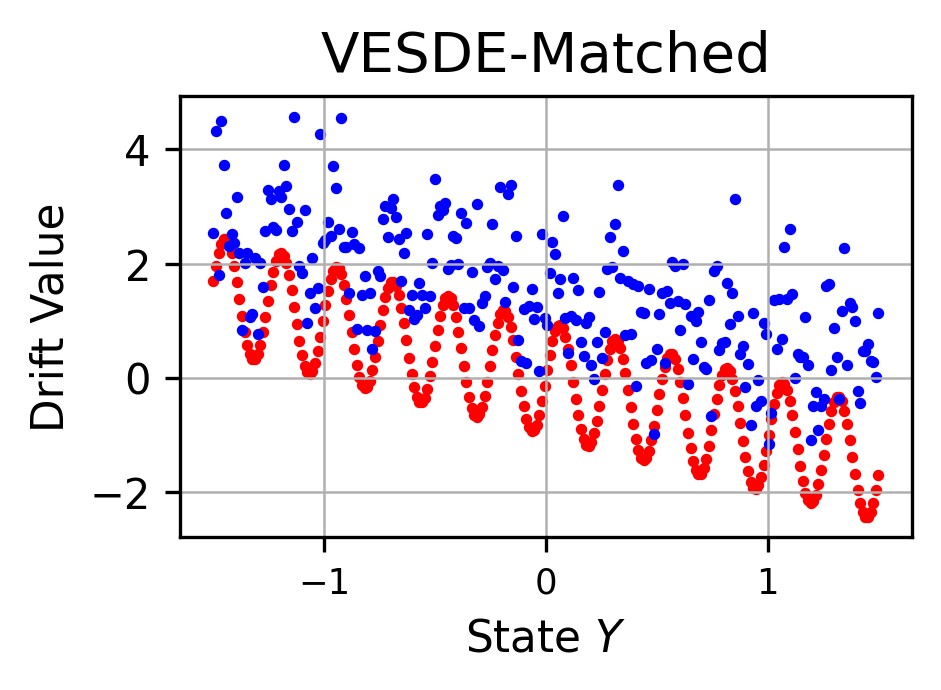} &
\includegraphics[width=0.46\columnwidth]{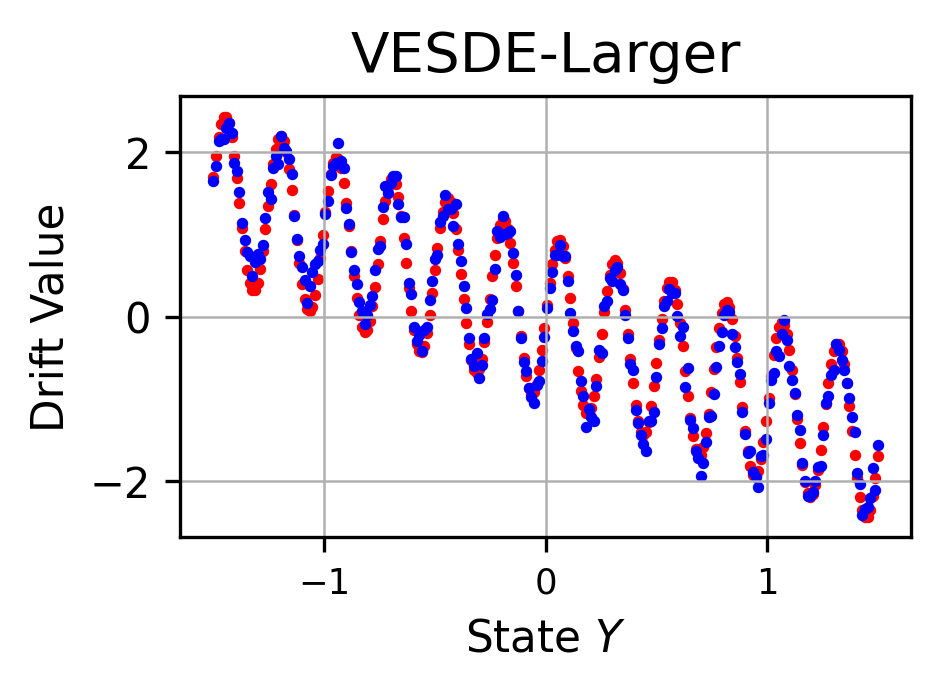} \\
\end{tabular}
\captionof{figure}{The top panels shows the drift error\(e^{2}(\tau)\) when estimating \(\mu^{(1)}\); the bottom panels compares the \textcolor{red}{true} drift to the \textcolor{blue}{estimated} drift at \(\tau=1\). Positions \(Y\) are uniformly spaced between \([-1.5, 1.5]\), covering \(99\%\) of the training distribution.}
\label{fig:VESDETauQuadSinHF}
\endgroup

\begingroup
\centering
\setlength{\abovecaptionskip}{0pt}
\setlength{\belowcaptionskip}{0pt}
\setlength{\parskip}{0pt}
\begin{tabular}{cc}
\includegraphics[width=0.46\columnwidth]{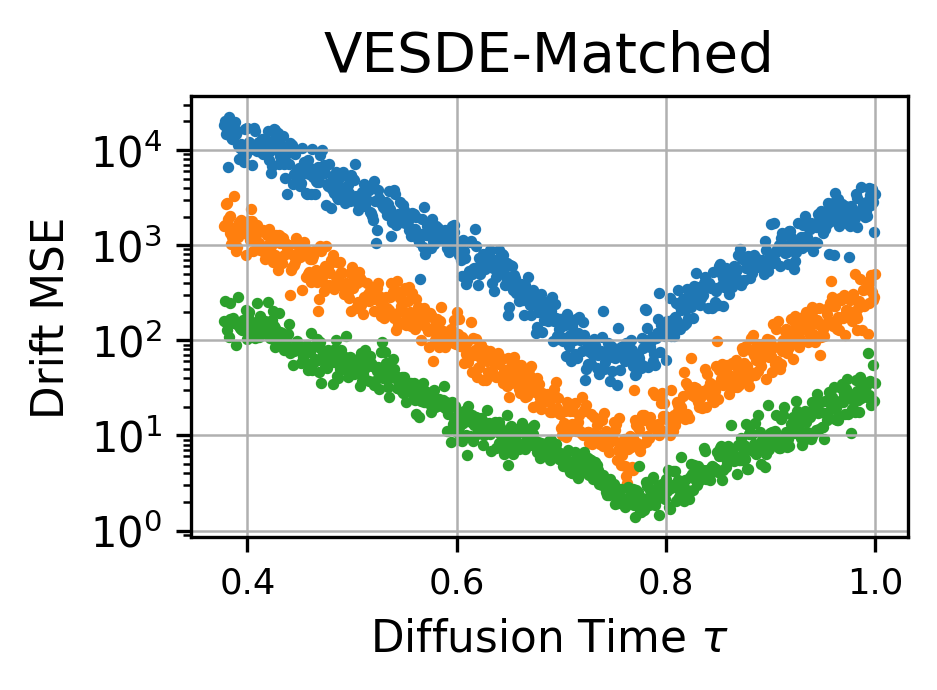} &
\includegraphics[width=0.46\columnwidth]{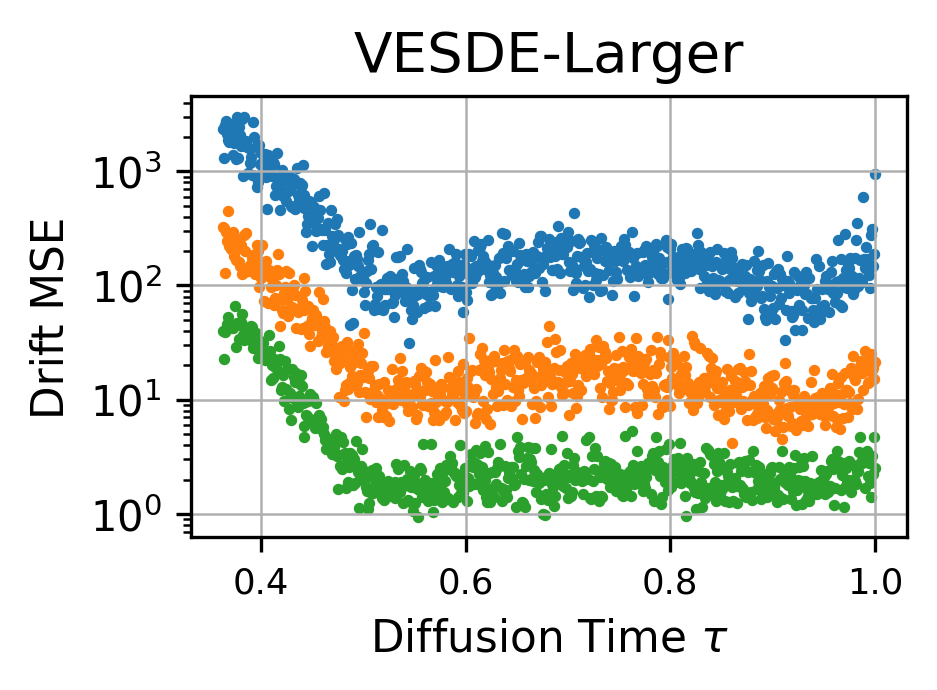} \\
\end{tabular}
\captionof{figure}{Drift error \(e^{2}(\tau)\) for \(\mu^{(2)}\) with \(D=8\) for VESDE-Matched (left) and VESDE-Larger (right).}
\label{fig:VESDETau8DDimsSep}
\endgroup

\begingroup
\centering
\setlength{\abovecaptionskip}{0pt}
\setlength{\belowcaptionskip}{0pt}
\setlength{\parskip}{0pt}
\begin{tabular}{cc}
\includegraphics[width=0.46\columnwidth]{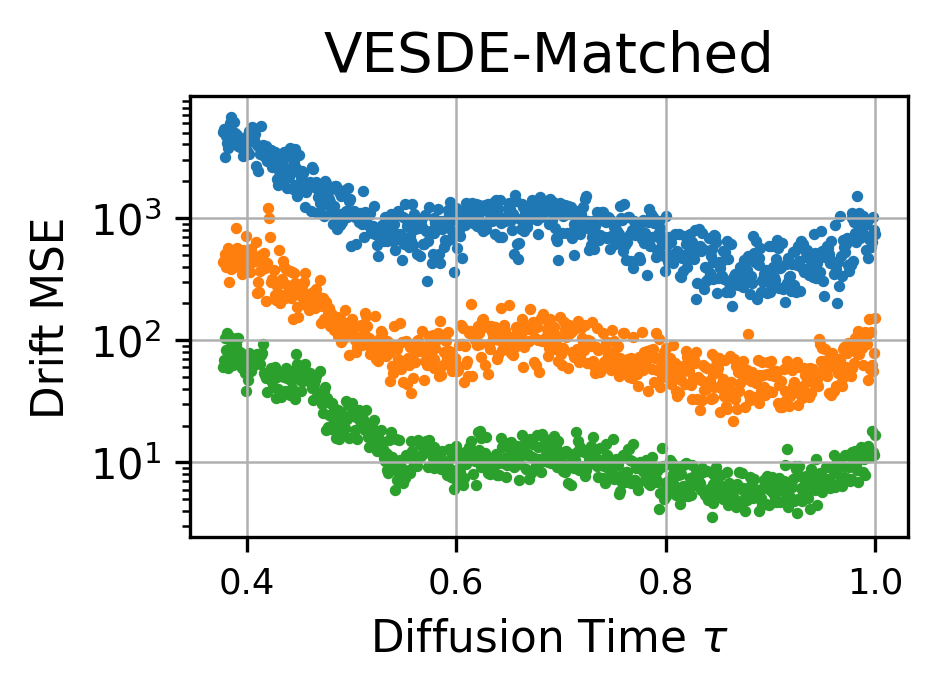} &
\includegraphics[width=0.46\columnwidth]{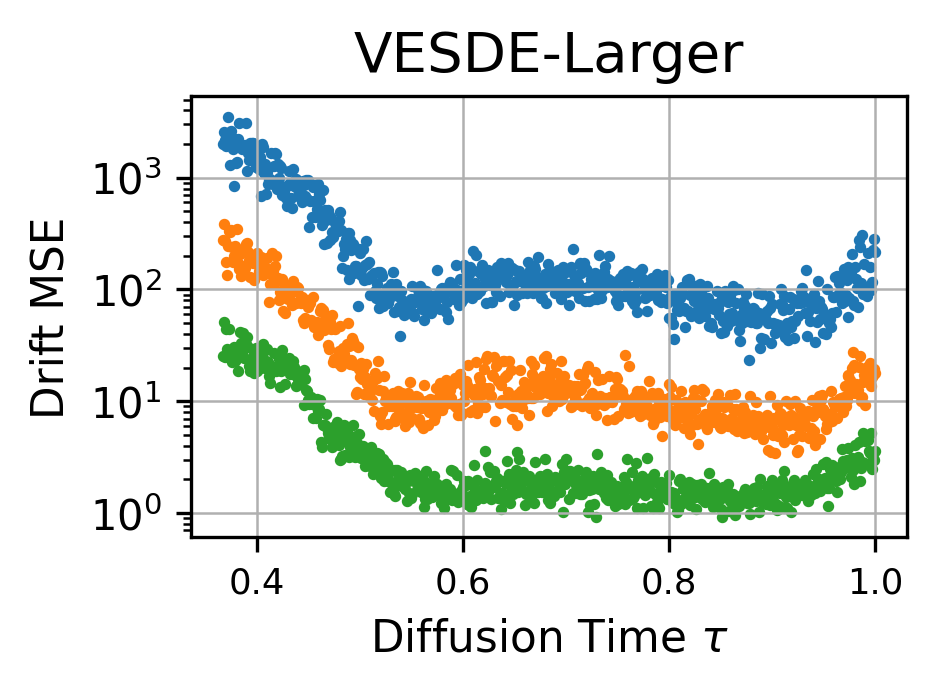} \\
\end{tabular}
\captionof{figure}{Drift error \(e^{2}(\tau)\) for \(\mu^{(2)}\) with \(D=12\) for VESDE-Matched (left) and VESDE-Larger (right).}
\label{fig:VESDETau12DDimsSep}
\endgroup

As in the VPSDE case, Figures \ref{fig:VESDETauQuadSinHF}-\ref{fig:VESDETau12DDimsSep} show increasing \(\mathcal{K}\) reduces the drift MSE uniformly across \(\tau\) for both VESDE variants. However, for fixed \(\mathcal{K}\), VESDE-Matched exhibits uniformly larger errors than VESDE-Larger. This gap is most pronounced for large diffusion times, where the forward process in VESDE-Larger has higher variance. Taken together, these results show that matching maximum marginal variance is insufficient for accurate drift estimation, and that effective denoising requires adequate coverage of high noise regimes during training.

\section{Denoising Network Architecture}\label{appendix:ModelArchitecture}

The architecture described in this section corresponds to the denoising network used for \(\mathrm{DN}\). The variant \(\mathrm{DN}\)-Lin is obtained by removing the \textit{MLPStateMapper} component (see Figure \ref{fig:MLPStateMapper} and replacing it with a fully connected mapping of matching output dimension, and no other module is changed.

The network for \(\mathrm{DN}\) is constructed such that the output is given by, 
\begin{equation*}
D_{\theta}(\tau, X_{\tau}, Y) = f_{\theta}(Y)+g_{\theta}(X_{\tau}, Y) + m_{\theta}(g_{\theta}, \tau).
\end{equation*}
Figure \ref{fig:NeuralNetwork} summarises the main components of the denoiser network architecture. The path followed by modules highlighted in blue represents the implementation of \(g_{\theta}\) as a sequence of \(1D\) convolutional neural networks and a single \(1D\) convolutional residual layer. The function \(m_{\theta}\), highlighted in orange, conditions the network on the diffusion time through a fully connected residual layer. The architecture for \(m_{\theta}\) is motivated by the vanishing dependence of traditional diffusion models on the diffusion time \cite{KimEtAl2024}.

\begingroup
\centering
\setlength{\abovecaptionskip}{0pt}
\setlength{\belowcaptionskip}{0pt}
\setlength{\parskip}{0pt}
\includegraphics[height=0.55\columnwidth, keepaspectratio]{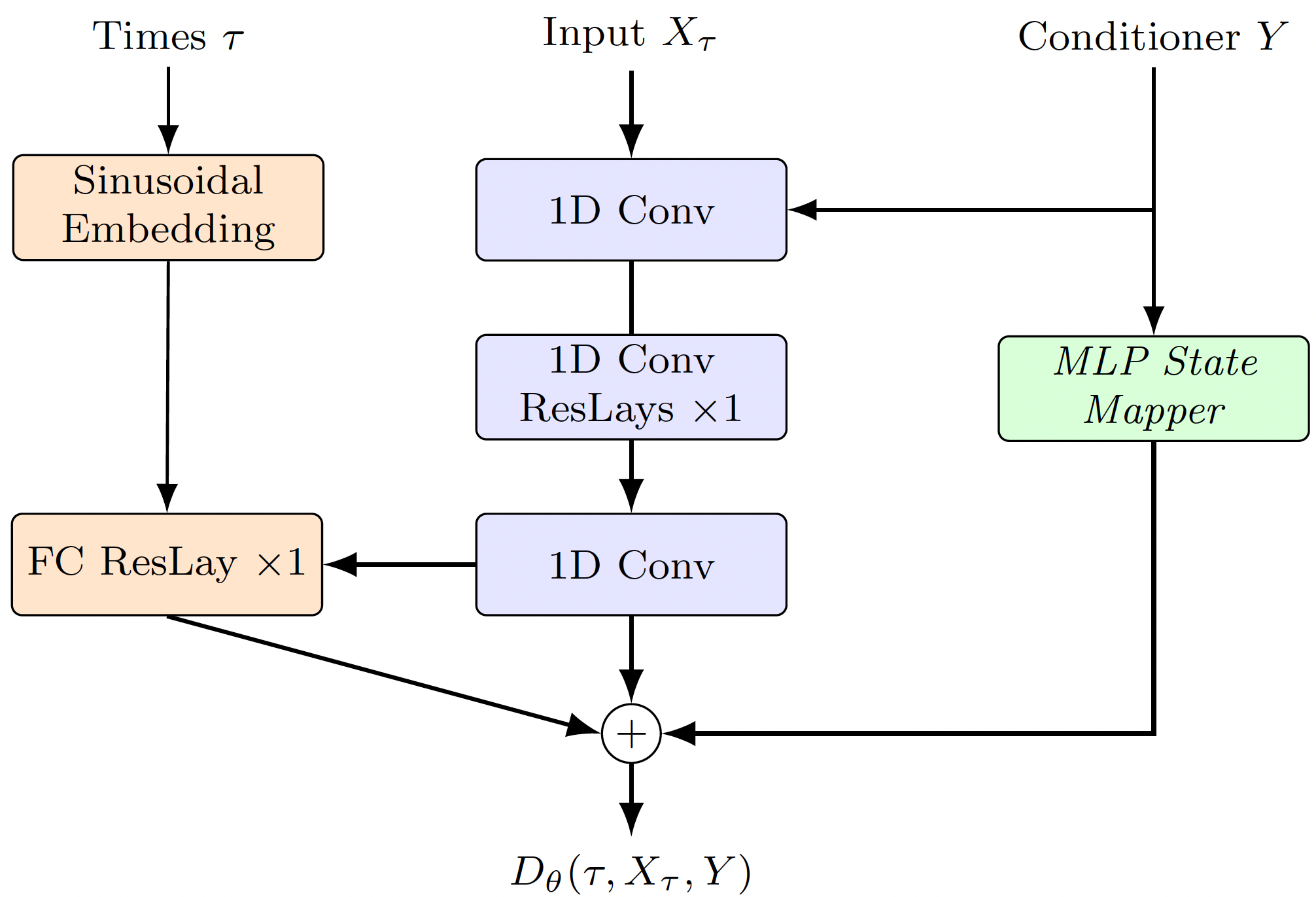}
\par\vspace{0.3em}\par\small
\captionof{figure}{Score network \(D_{\theta}\) used for \(\mathrm{DN}\). }
\label{fig:NeuralNetwork}
\endgroup

In prior diffusion model architectures, the conditioning mapping \(f_{\theta}\) is often implemented using sequence-based architectures such as Long Short-Term Memory (LSTM) networks or Transformers \cite{YanEtAl2021, CorsoEtAl2023, LimEtAl2023, SahariaEtAl2022}. These architectures maintain an internal state that evolves with the history of observed inputs. In this paper, \(f_{\theta}\) is implemented using the \textit{MLPStateMapper}, see Figure \ref{fig:MLPStateMapper}, which defines a non-linear feature mapping of the observed state \(Y\) without maintaining a recurrent or attention-based state. This design reflects the Markovian dependence structure induced by the data-generating SDE in \eqref{eq:BSDE}. The mapping is constructed by combining multiple feature pathways: element-wise polynomial features \((Y, Y^{2}, Y^{3})\), a learned global non-linear transformation implemented as a linear--ELU block, and Fourier features. The Fourier features are modulated by a learnable scalar gate, allowing their contribution to be adaptively scaled during training. All feature components are concatenated and passed through a two-layer fully connected network to produce the final output in \(\mathbb{R}^{D}\).

The number of trainable parameters in \(D_{\theta}\) scales as \(T(D) = 2D^{2}+\mathcal{O}(D) + \mathcal{O}(1)\). The quadratic dependence on the time-series dimensionality arises from the first \(1D\) convolution in \(g_{\theta}\), see Figure \ref{fig:NeuralNetwork}, although the constant term in \(T(D)\) is much larger than the linear or quadratic terms.

\begingroup
\centering
\setlength{\abovecaptionskip}{0pt}
\setlength{\belowcaptionskip}{0pt}
\setlength{\parskip}{0pt}
\includegraphics[height=0.55\columnwidth, keepaspectratio]{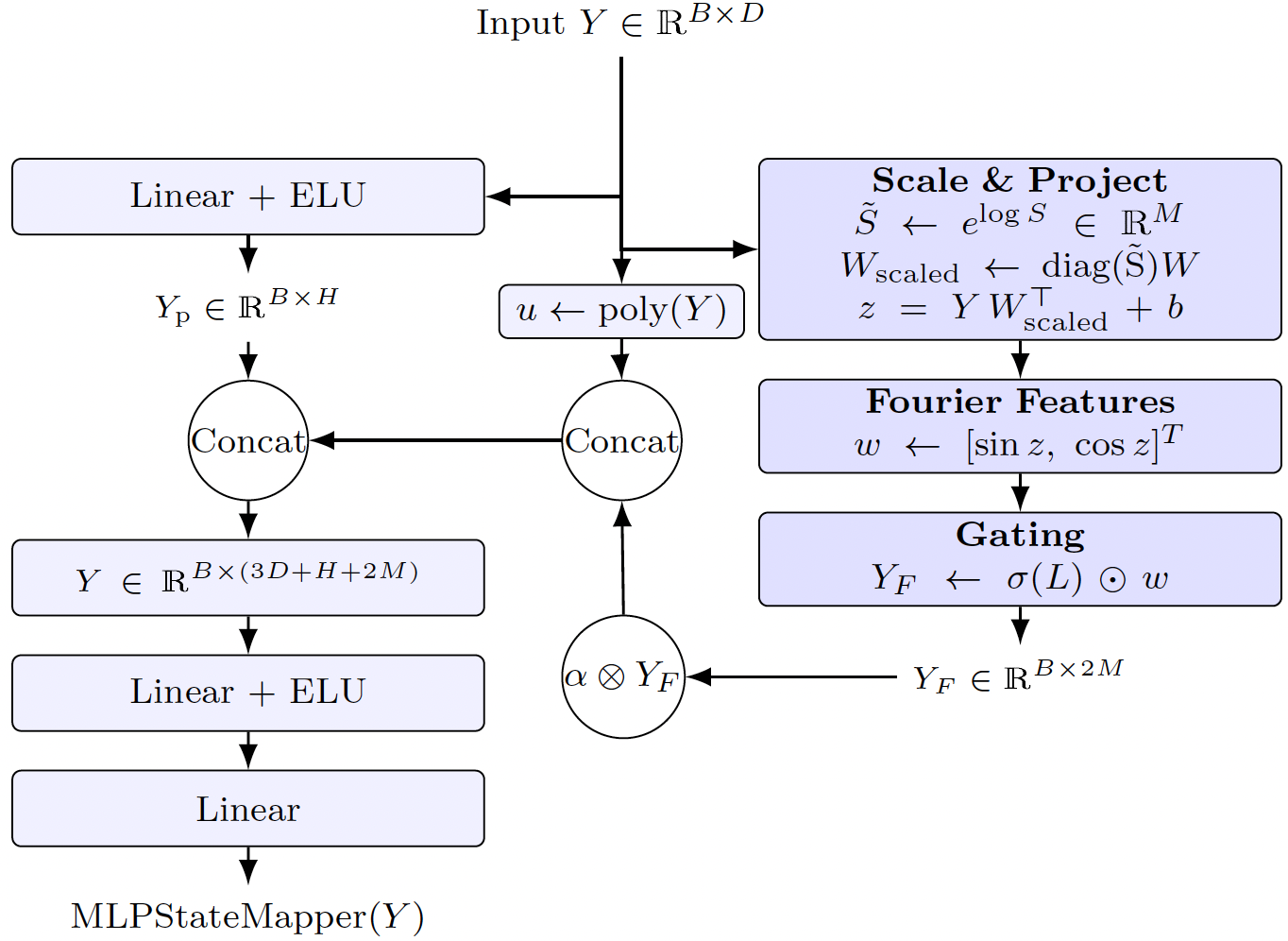}
\par\vspace{0.3em}\par\small

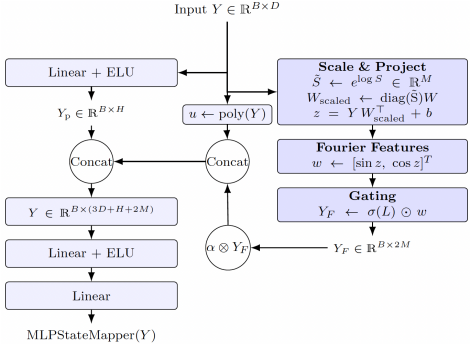
\captionof{figure}{ \textit{MLPStateMapper} module. It provides an inductive bias towards polynomial and Fourier features, which can be learned from the data.}
\label{fig:MLPStateMapper}
\endgroup

\section{Baselines}\label{appendix:Baselines}
\subsection{IID Nadaraya-Watson Estimator}
In \cite{MarieEtAl2021}, the authors improve the standard Nadaraya-Watson estimator for the drift function by averaging over \(\mathcal{I}\) IID sample paths observed discretely \(\mathcal{J}\) times. Assuming each path is defined on the same uniform time grid with \(\Delta = t_{j+1}-t_{j}\), the discrete-time estimator for the drift function is given by,
\begin{align}
    \widehat{b}_{\mathcal{J},\mathcal{I},h}(y) = \frac{\widehat{bf}_{\mathcal{J},\mathcal{I},h}(y)}{\widehat{f}_{\mathcal{J},\mathcal{I},h}(y)},\label{eq:IIDBenchmarkEstimator}
\end{align}
where,
\begin{align}
    \widehat{f}_{\mathcal{J},\mathcal{I},h}(y) = \frac{1}{\mathcal{J}\mathcal{I}}\sum_{i=1}^{\mathcal{I}}\sum_{j=0}^{\mathcal{J}-1}K_{h}(Y_{t_{j}}^{(i)}-y),
\end{align}
and,
\begin{align}
    \widehat{bf}_{\mathcal{J},\mathcal{I},h}(y) = \frac{1}{\mathcal{I}}\sum_{i,j}K_{h}(Y_{t_{j}}^{(i)}-y)Z_{t_{j}}^{(i)},
\end{align}
and \(Y_{t_{j}}^{(i)}\), \(Z_{t_{j}}^{(i)}\) are defined in Section \ref{section:DriftEstimationDiffusionModelMethodology}. In our experiments, we choose \(K_{h}(y)=(2\pi h^{2D})^{-0.5}\exp\left[-0.5h^{-2}\left\|y\right\|_{2}^{2}\right]\).

To choose the optimal bandwidth \(h\), one finds the bandwidth which minimises the leave-one-out-cross validation criterion,
\begin{equation*}
\begin{aligned}
    CV(h) = \sum_{i=1}^{\mathcal{I}}&\sum_{j=0}^{\mathcal{J}-1}\Bigg[\widehat{b}^{-(i)}_{\mathcal{J},\mathcal{I},h}(Y_{t_{j}}^{(i)})^{2}\Delta  \\
    &- 2\widehat{b}^{-(i)}_{\mathcal{J},\mathcal{I},h}(Y_{t_{j}}^{(i)})\big(Y_{t_{j+1}}^{(i)}-Y_{t_{j}}^{(i)}\big)\Bigg]
\end{aligned}
\end{equation*}
where \(\widehat{b}^{-(i)}_{\mathcal{J},\mathcal{I},h}\) is the leave-one-out drift estimator.

Under the regularity assumptions of \cite{MarieEtAl2021}, parameterised by a spatial smoothness index \(\beta \geq 1\) of the marginal density \(q_t(\cdot)\) of \eqref{eq:BSDE}, and a discretization parameter \(\epsilon \in [0,1)\), risk control in expected squared \(L_2\) error is achieved via the truncated estimator,
\begin{align*}
\widetilde{b}_{\mathcal{J},\mathcal{I},h}
=
\widehat{b}_{\mathcal{J},\mathcal{I},h}\,
\mathbbm{1}\!\left(\widehat{f}_{\mathcal{J},\mathcal{I},h} > m/2\right),
\end{align*}
where \(m>0\) is a lower bound on the time-averaged marginal density, \(f\). Under these conditions, the estimator over a compact domain \([A, B]\) satisfies the risk bound,
\begin{equation*}
\begin{aligned}
\mathbbm{E}\bigl\|\widetilde{b}_{\mathcal{J},\mathcal{I},h} - b\bigr\|_{f, A, B}
&=
\mathcal{O}\!\Big((T-t_{0})\mathcal{J}^{-1} + h^{2\beta} \\
&+ \mathcal{I}^{-1}h^{-1} + \mathcal{I}\mathcal{J}h^{3+\epsilon}\Big).
\end{aligned}
\end{equation*}
As noted in \cite{MarieEtAl2021}, when \(T>1\), choosing \(t_{0}\in[1,T-1]\) removes the time-dependent factors in the risk rate. In the worst-case regime, \(h<1\), the rate is maximised at \(\beta=1\), and if the true drift function is bounded, one may take \(\epsilon=0\).
\subsection{IID Ridge Estimator}
The IID Ridge estimator in \cite{DenisEtAl2021} is a nonparametric estimator of a one dimensional drift function using \(B\)-splines.
Letting \(L_{\mathcal{I}}, K_{\mathcal{I}}, M > 0\), and end points \(A_{\mathcal{I}} = - B_{\mathcal{I}} < 0\), \cite{DenisEtAl2021} define a sequence of knots, 
\begin{align*}
u_s =
\begin{cases}
A_{\mathcal I},
& s = -M,\dots,0, \\[6pt]
A_{\mathcal I} + \dfrac{s}{K_{\mathcal I}}\bigl(B_{\mathcal I}-A_{\mathcal I}\bigr),
& s = 1,\dots,K_{\mathcal{I}}, \\[6pt]
B_{\mathcal I},
& s = K_{\mathcal{I}}+1,\dots,K_{\mathcal I}+M.
\end{cases}
\end{align*}

Denoting the spline basis by \(B_{s, M, u}\), and the projection subspace as \(S_{K_{\mathcal{I}}, M, {u}} = \text{span}\{(B_{s, M, {u}}) \ \forall s \in \{-M, \cdots, K_{\mathcal{I}}-1\}\}\), such that \(\text{dim}(S_{K_{\mathcal{I}}, M, {u}}) = K_{\mathcal{I}}+M\), then for any \(y\in [A_{\mathcal{I}}, B_{\mathcal{I}}]\), the drift estimator is given by,
\begin{align}
    \widehat{b}_{\mathcal{I}, \mathcal{J}}(y) = \sum_{s=-M}^{K_{\mathcal{I}}-1}\widehat{a}_{s}B_{s, M, {u}}(y) \label{eq:IIDRidgeEstimator}
\end{align}
Defining the entries in the matrix \({B}\) as \(B_{r,s}=B_{s,M,u}\bigl(Y^{(i)}_{t_j}\bigr)\) with row index \(r=\mathcal J(i-1)+j\), and \({Z}\in\mathbbm{R}^{\mathcal{I}\mathcal{J}}\) where \(Z_{r} = \Delta^{-1}(Y_{t_{j+1}}^{(i)}-Y_{t_{j}}^{(i)})\), the solution to \eqref{eq:IIDRidgeEstimator} satisfies \(\widehat{{a}} = ({B}^{T}{B})^{-1}{B}^{T}{Z}\) if \({B}^{T}{B}\) is full column rank  and \(\|\widehat{{a}}\|_{2}^{2}\leq (K_{\mathcal{I}}+M)L_{\mathcal{I}}\). Alternatively, letting \(\widehat{\lambda}\) be the unique solution to \(\|({B}^{T}{B}+{\lambda}{I})^{-1}{B}^{T}{Z}\|_{2}^{2} = (K_{\mathcal{I}}+M)L_{\mathcal{I}}\), the solution is given by \(\widehat{{a}} = ({B}^{T}{B}+\widehat{\lambda}{I})^{-1}{B}^{T}{Z}\).

Assuming only the necessary conditions for the existence and uniqueness of a strong solution to \eqref{eq:BSDE}, \cite{DenisEtAl2021} show that the risk bound is controlled by a bias term and a variance term which scales as \(\left(\text{dim}(S_{K_{\mathcal{I}}, M, {u}})\mathcal{I}^{-1}\right)^{-0.5}+\Delta\). Under further regularity conditions for the drift function, \cite{DenisEtAl2021} obtain a risk bound of order \(\left(\ln^{2}\mathcal{I}/\mathcal{I}\right)^{\frac{2\beta}{2\beta+1}}\).
\subsection{IID Hermite Projection Estimator}
Comte and Genon-Catalot \cite{ComteGenonCatalot2020} derive a nonparametric estimator for a one dimensional drift function using a least-squares projection onto a subspace spanned by Hermite polynomials. Given the first \(m\) Hermite polynomial basis functions \((h_{k})_{k=0}^{m-1}\), and \(\mathcal{I}\) sample paths observed continuously on \(t\in [0, T]\), the drift estimator is given by
\begin{align*}
    \widehat{b}_{m}(y) = \sum_{k=0}^{m-1}\widehat{\theta}_{k}h_{k}(y),
\end{align*}
where \((\widehat{\theta}_{k})_{k=0}^{m-1}\) is the vector of parameters obtained via the least-squares projection onto the space spanned by the Hermite polynomials, \(h_{k}\),
\begin{align}
    \widehat{{\theta}}_{k} = e_{k}^{T}\widehat{{\Phi}}_{m}^{-1}\widehat{{Z}}_{m}\label{eq:HermiteSolution}.
\end{align}
where \(e_{k}\) is the \(k\)-th canonical basis vector. In \eqref{eq:HermiteSolution}, \(\widehat{\Phi}_{m}\in \mathbbm{R}^{m\times m}, \widehat{Z}_{m}\in\mathbbm{R}^{m}\) are defined as,
\begin{align}
    \widehat{\Phi}_{m, (u,v)} &= \frac{1}{\mathcal{I}T}\sum_{i=1}^{\mathcal{I}}\int_{0}^{T}h_{v}(Y^{(i)}_{s})h_{u}(Y^{(i)}_{s}) ds\nonumber\\
    \widehat{Z}_{m, u} &= \frac{1}{\mathcal{I}T}\sum_{i=1}^{\mathcal{I}}\int_{0}^{T}h_{u}(Y_{s}^{(i)})dY_{s}^{(i)}\nonumber,
\end{align}
where \(u, v \in \{0, \cdots m-1\}\).  

To find the optimal truncation level \(m\), \cite{ComteGenonCatalot2020} propose to optimise
\begin{align*}
    \widehat{m} = \arg_{m\in \mathcal{M}}\min\left[ -\left\|\widehat{b}_{m}\right\|^{2}_{\mathcal{I}} + \kappa\left\|\widehat{{\Phi}}^{-1}_{m}\widehat{{\Phi}}_{m, \sigma^{2}}\right\|_{op}\frac{m}{\mathcal{I}T}\right],
\end{align*}
where \(\mathcal{M} = \left\{m: m\leq 10, m\left\|\widehat{{\Phi}}^{-1}_{m}\right\|^{1/4}_{op}\leq \mathcal{I}T\right\}\), see \cite{ComteGenonCatalot2020} for notation.

In the case of continuous observations of the sample paths and bounded diffusion coefficient, Theorem 1 in \cite{ComteGenonCatalot2020} proves the risk bound on \(\widehat{b}_{\widehat{m}}\) is order \(\mathcal{O}(1/\mathcal{I}T)\).

\subsection{Neural-Network Based Benchmark Estimators}
The benchmark \(\mathrm{FC}\) is constructed as a fully-connected feedforward network with ReLU activation functions, as in \cite{ZhaoLiuHoffmann2025}. The number of layers and the width of each layer are chosen based on the empirical results in the code repository by \cite{ZhaoLiuHoffmann2025}.  

However, for a fixed dataset, the number of parameters in \(\mathrm{FC}\) can be several orders of magnitude smaller than our denoising network \(D_{\theta}\). The benchmark \(\mathrm{FC}^{+}\) increases the depth of \(\mathrm{FC}\) until the number of parameters exceeds the parameter count in \(D_{\theta}\). Architectural details achieving this constraint are fixed deterministically and reported in the code repository.

We implement \(\mathrm{FC^{+}}\)-Conv as the sum of a fully-connected backbone and a convolutional module,
\begin{equation*}
(\mathrm{FC}^{+}\text{-Conv})_{\theta}(Y) \equiv \mathrm{FC}_{\theta}(Y) + g_{\theta}(Y),
\end{equation*}
The convolutional module \(g_{\theta}\) follows the same design pattern as the convolutional path in \(\mathrm{DN}\), but with the first channel dimension adjusted to account for the absence of noisy inputs \(X_{\tau}\). The fully-connected backbone of \(\mathrm{FC^{+}}\)\text{-Conv} is resized relative to  \(\mathrm{FC^{+}}\) so that the \emph{total} number of trainable parameters of \(\mathrm{FC^{+}}\)\text{-Conv} matches that of \(D_{\theta}\) for the same dataset.

\section{Estimator hyperparameter selection}\label{appendix:FullDataAppendix}
\subsection{IID Hermite Estimator}
Figure \ref{fig:HermiteValidation} shows how the number of basis functions for the Hermite estimator in \cite{ComteGenonCatalot2020} is selected by optimising \(E^{(\mu)}_{1}\). As suggested in \cite{ComteGenonCatalot2020}, increasing the number of basis functions beyond \(m=10\) yields small marginal improvements in estimator accuracy; nevertheless, we conduct a grid search for \(m>10\) for completeness.

\begingroup
\centering
\setlength{\abovecaptionskip}{0pt}
\setlength{\belowcaptionskip}{0pt}
\setlength{\parskip}{0pt}
\includegraphics[width=0.48\columnwidth]{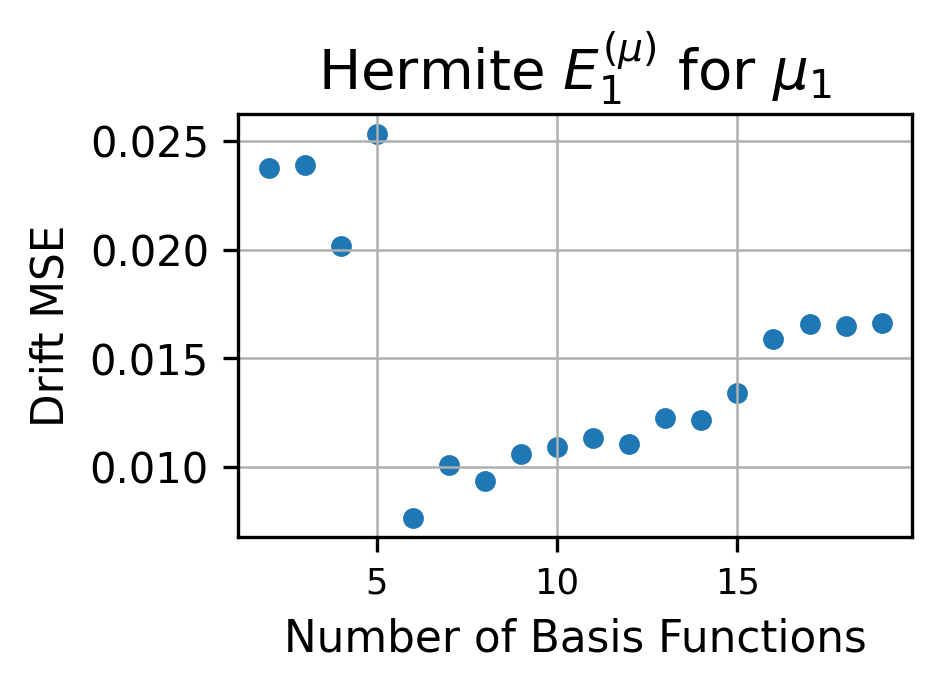}\hspace{0.01\columnwidth}%
\includegraphics[width=0.48\columnwidth]{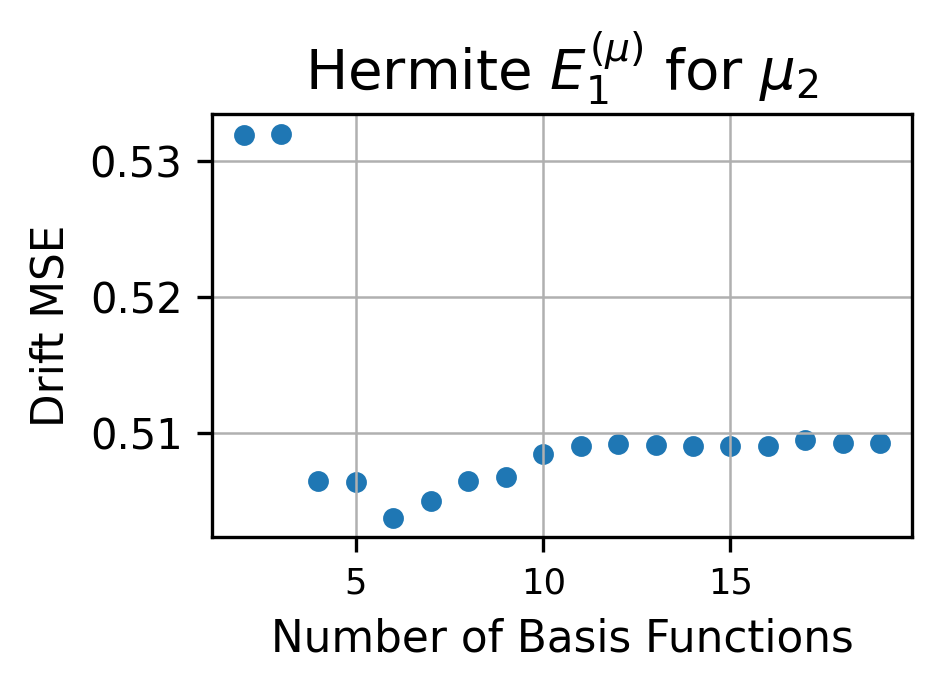}
\vspace{0.25em}
\includegraphics[width=0.48\columnwidth]{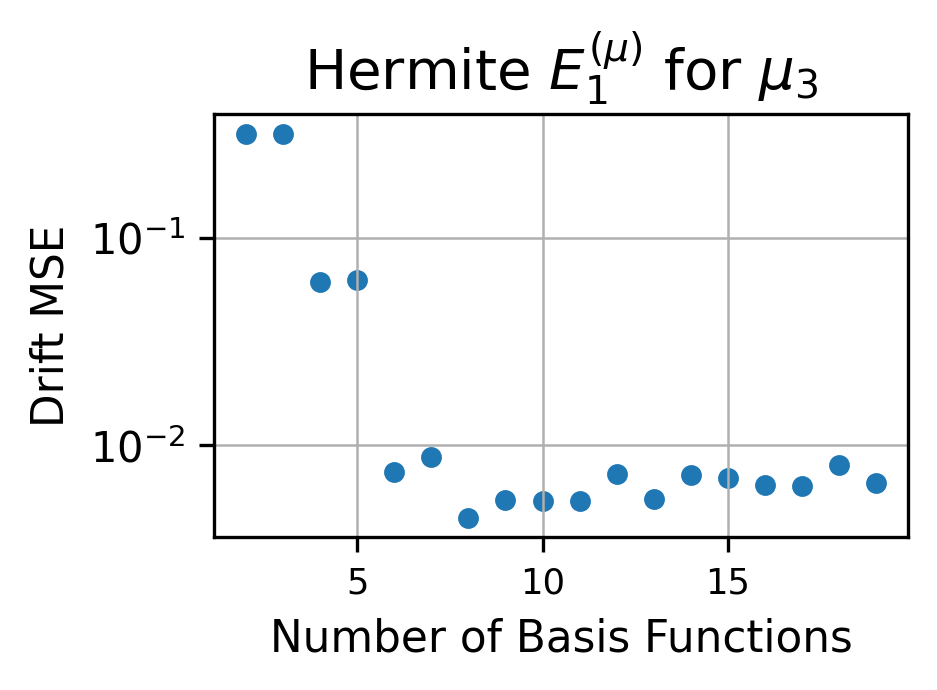}
\par\vspace{0.25em}\par
\small
\captionof{figure}{Hermite drift error \(E^{(\mu)}_{1}\), against number of basis functions. From left to right, top to bottom, the estimation for \(\mu_{1}\), \(\mu_{2}\), \(\mu_{3}\).}
\label{fig:HermiteValidation}
\endgroup

\subsection{IID Ridge Estimator}
Figure \ref{fig:RidgeValidation} shows how the number of subspace dimensions for the Ridge estimator from \cite{DenisEtAl2021} is chosen by minimising \(E^{(\mu)}_{1}\). The Ridge estimator requires a higher \(B\)-spline dimension to accurately estimate \(\mu_{2}\), plausibly due to the presence of a high-frequency sinusoidal component in the drift function. In contrast, for \(\mu_{1}, \mu_{3}\), a relatively small number of \(B\)-splines are sufficient for accurate drift recovery.
 
\begingroup
\centering
\setlength{\abovecaptionskip}{0pt}
\setlength{\belowcaptionskip}{0pt}
\setlength{\parskip}{0pt}
\includegraphics[width=0.48\columnwidth]{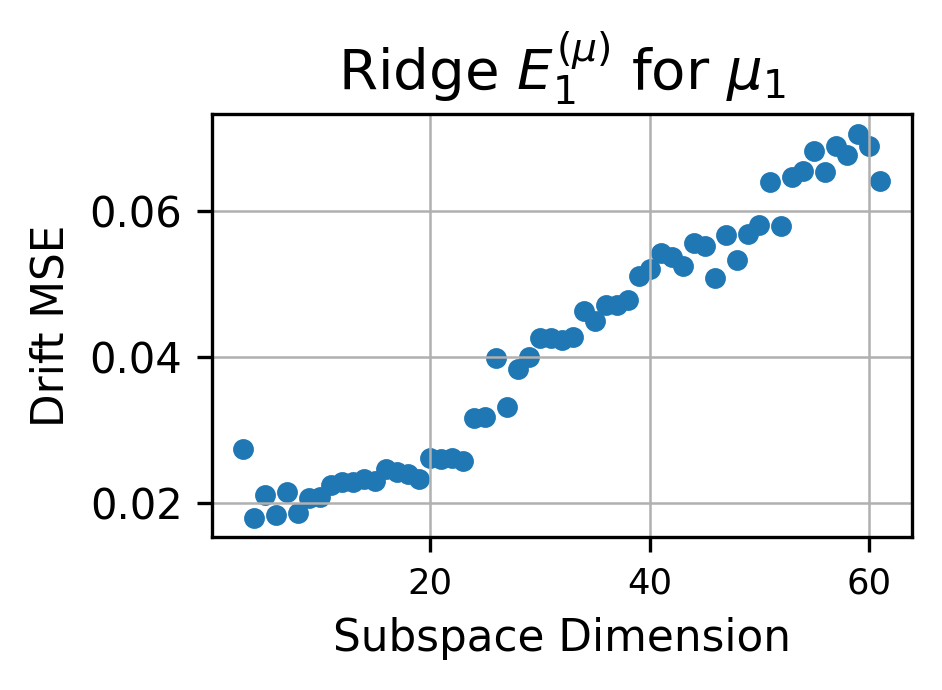}\hspace{0.01\columnwidth}%
\includegraphics[width=0.48\columnwidth]{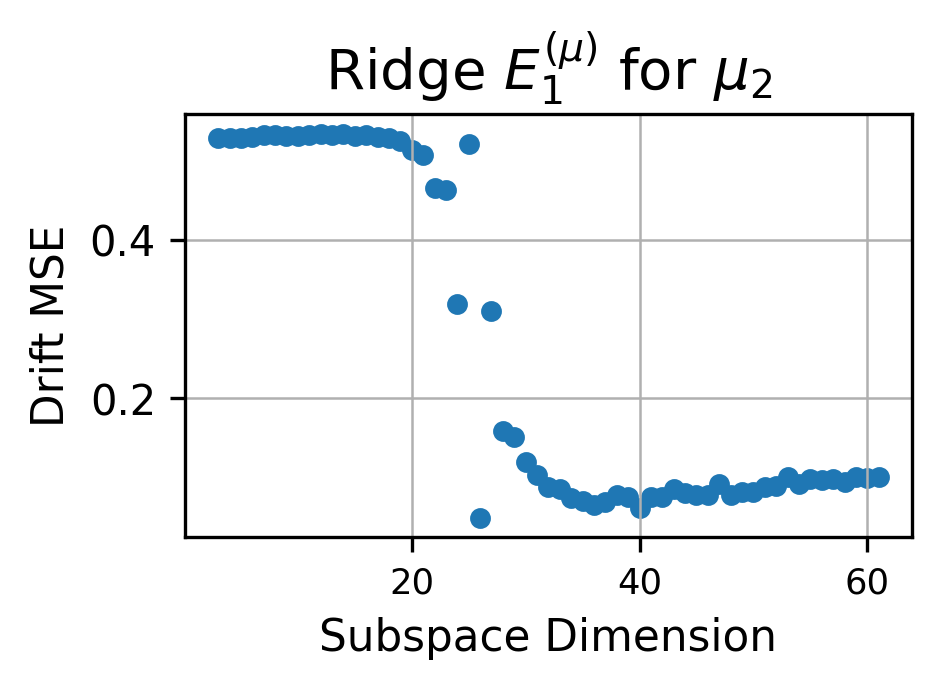}
\vspace{0.25em}
\includegraphics[width=0.48\columnwidth]{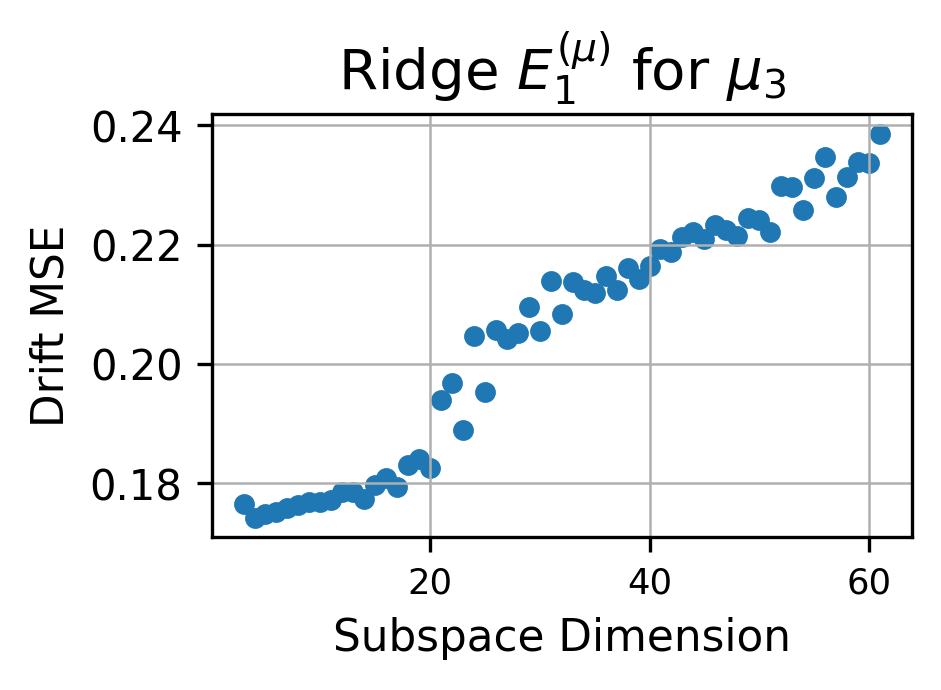}
\par\vspace{0.25em}\par
\small
\captionof{figure}{Ridge drift error \(E^{(\mu)}_{1}\), against number of basis functions. From left to right, top to bottom, the estimation for \(\mu_{1}\), \(\mu_{2}\), \(\mu_{3}\).}
\label{fig:RidgeValidation}
\endgroup
\subsection{IID Nadaraya Estimator}

Figures \ref{fig:NadDDimsBandwidth} and \ref{fig:NadDLnzBandwidth} illustrate how the bandwidth is chosen for the Nadaraya-Watson estimator when estimating \(\mu_{4}\)  and \(\mu_{5}\). For small bandwidths, the Nadaraya estimator discards many training samples, leading to large errors. As the bandwidth increases, the error decreases until it rises again once the estimator enters an over-diffuse regime in which all training points are weighted uniformly. Figure \ref{fig:NadDLnzBandwidth} shows that the optimal bandwidth increases slightly with dimension, as expected. 

\begingroup
\centering
\setlength{\abovecaptionskip}{0pt}
\setlength{\belowcaptionskip}{0pt}
\setlength{\parskip}{0pt}
\includegraphics[width=0.48\columnwidth]{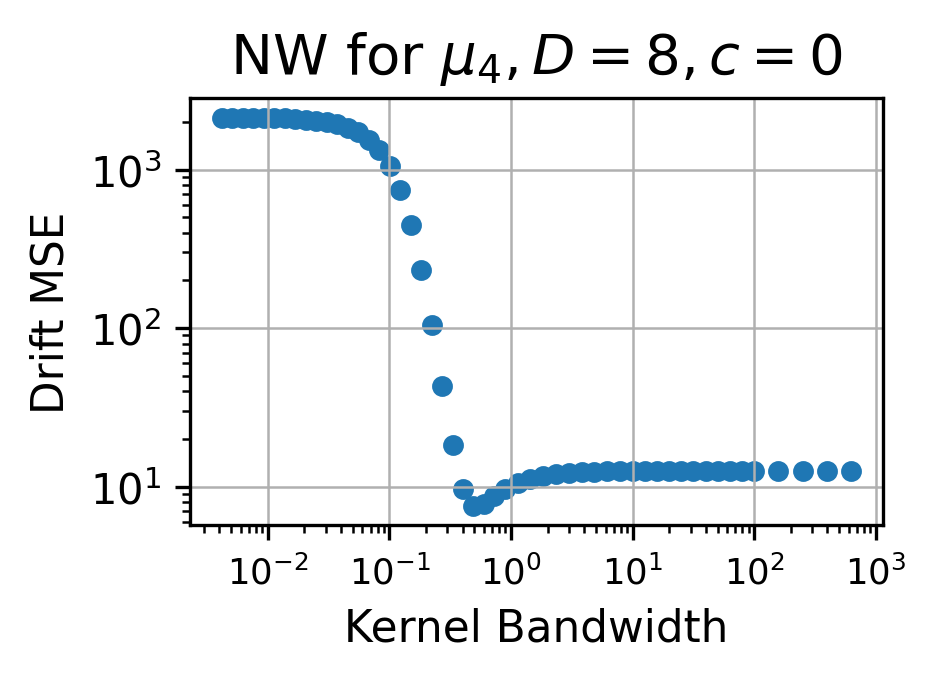}\hspace{0.04\columnwidth}%
\includegraphics[width=0.48\columnwidth]{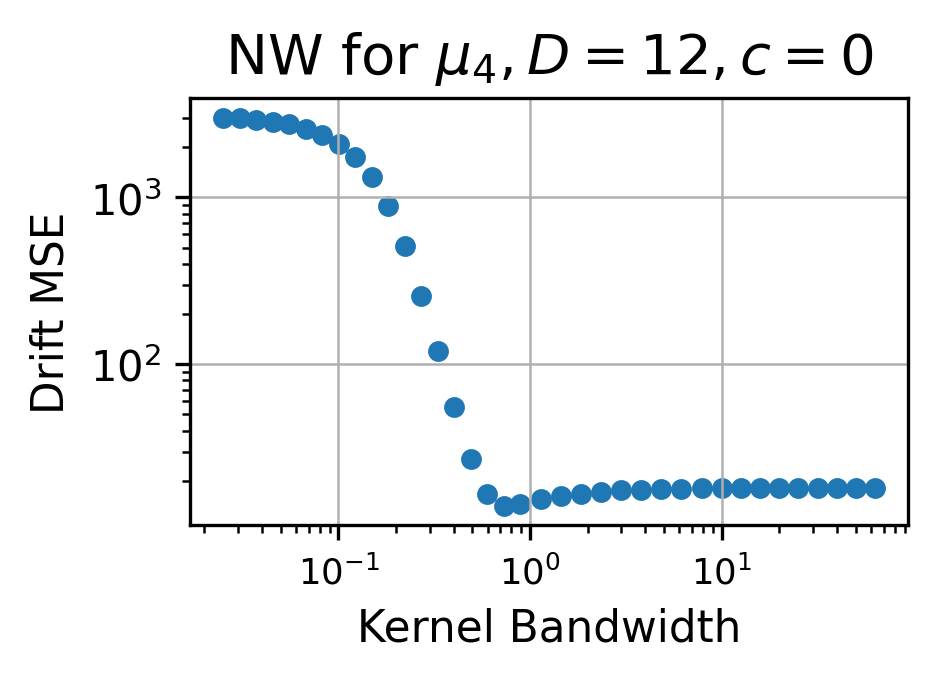}
\vspace{0.25em}
\includegraphics[width=0.48\columnwidth]{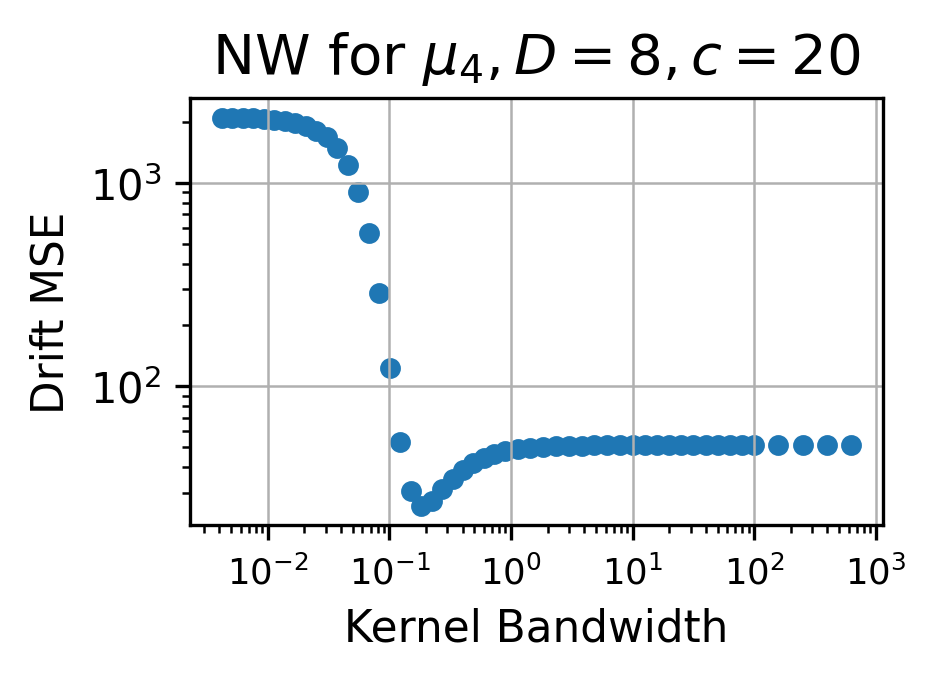}\hspace{0.04\columnwidth}%
\includegraphics[width=0.48\columnwidth]{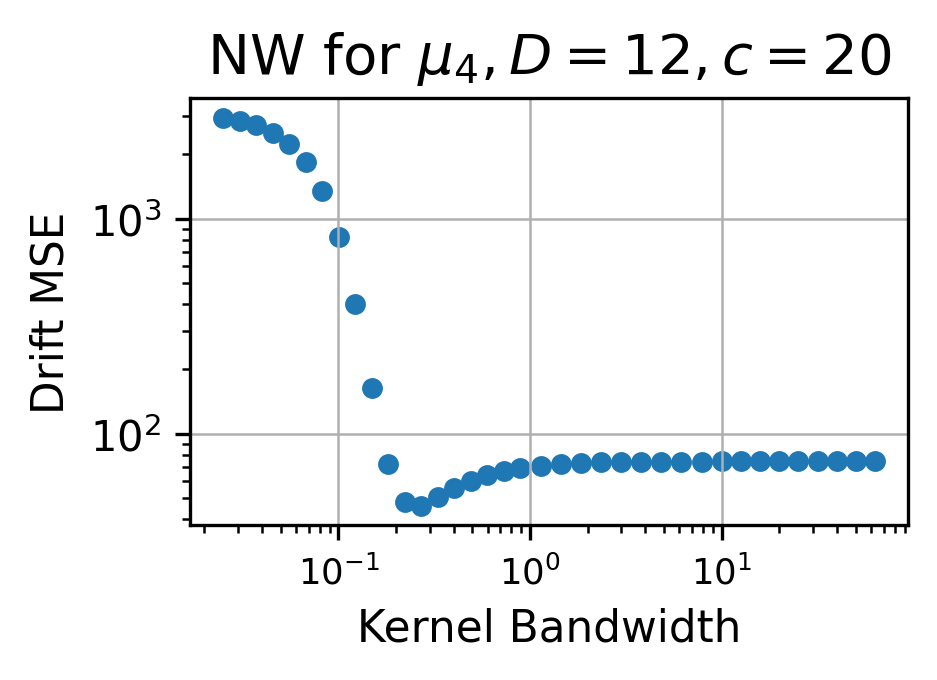}
\captionof{figure}{ Nadaraya drift error \(E^{(\mu)}_{1}\) versus bandwidth \(h\) for estimation of \(\mu_{4}\).}
\label{fig:NadDDimsBandwidth}
\endgroup

\begingroup
\centering
\setlength{\abovecaptionskip}{0pt}
\setlength{\belowcaptionskip}{0pt}
\setlength{\parskip}{0pt}
\includegraphics[width=0.48\columnwidth]{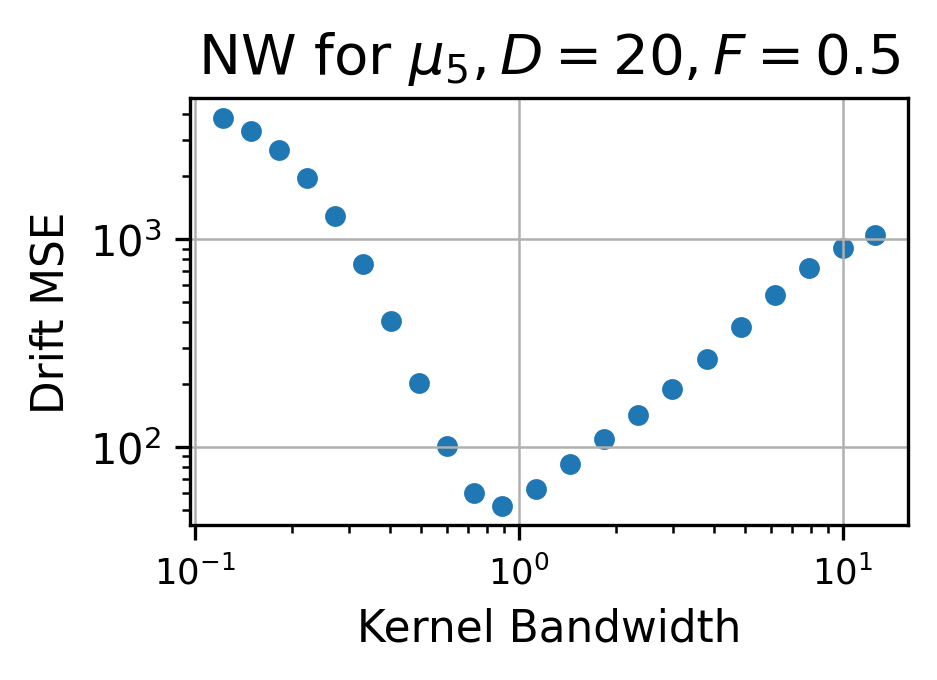}\hspace{0.04\columnwidth}%
\includegraphics[width=0.48\columnwidth]{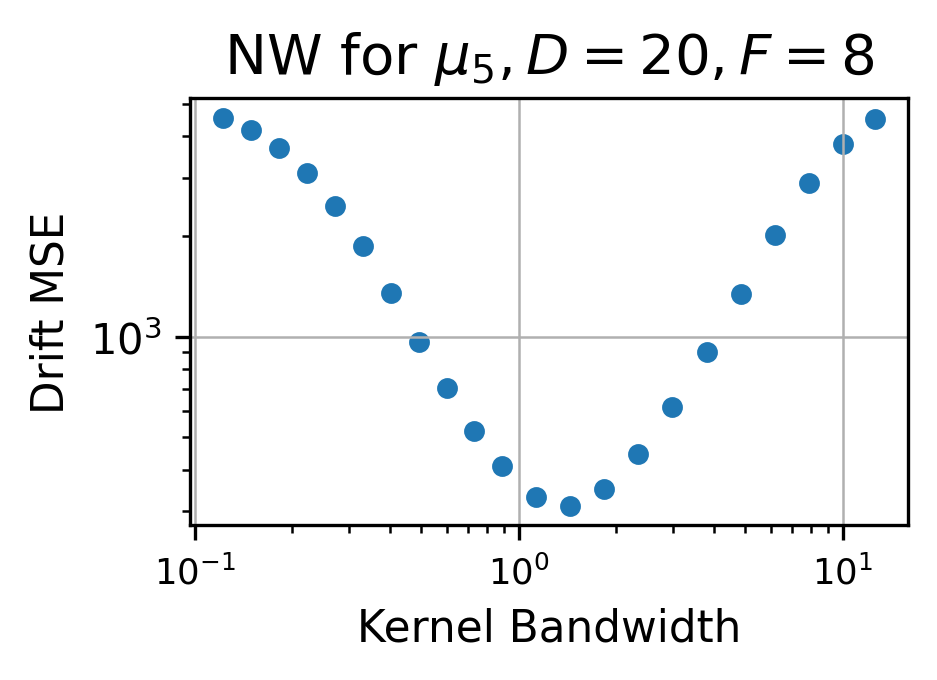}
\vspace{0.25em}
\includegraphics[width=0.48\columnwidth]{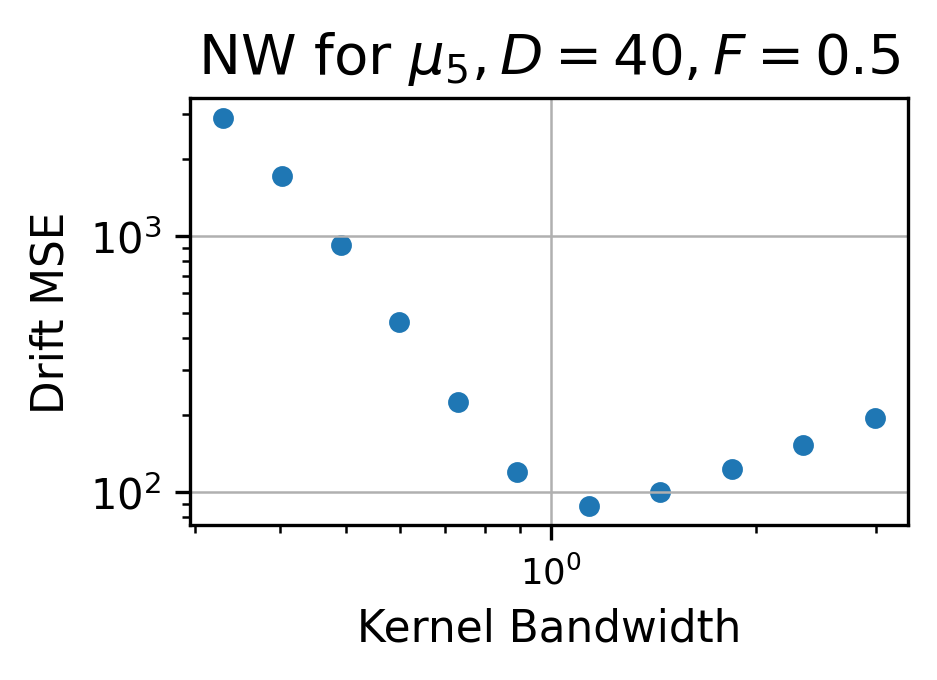}\hspace{0.04\columnwidth}%
\includegraphics[width=0.48\columnwidth]{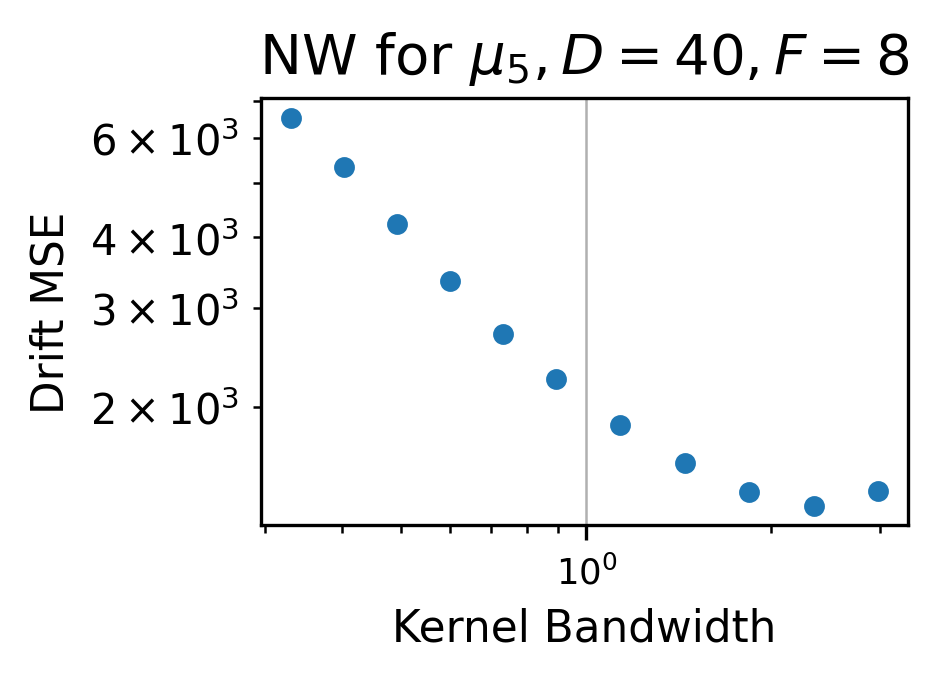}
\captionof{figure}{ Nadaraya drift error \(E^{(\mu)}_{1}\) versus bandwidth \(h\) for estimation of \(\mu_{5}\). Top panels show \(D=20, F \in \{0.5, 8\}\) (left to right); bottom panels show \(D=40, F \in \{0.5, 8\}\).}
\label{fig:NadDLnzBandwidth}
\endgroup

\subsection{Denoising Estimator}
We train the network \(D_{\theta}\) with the \textit{MLPStateMapper} module from Figure \ref{fig:MLPStateMapper} using Adam \cite{KingmaBa2015} with an initial learning rate of \(10^{-2}\). We allow the learning rate to decrease by a constant factor of \(0.9\) whenever the training loss fails to improve for \(60\) consecutive epochs, setting the minimum learning rate at \(10^{-3}\). The output weights in \(g_{\theta}, m_{\theta}\) are initialised to zero. For the conditional diffusion model hyperparameters, we set \(\epsilon = 10^{-3}, S = 10^{4}, \gamma_{0}=0, \gamma_{1}=20\). We train on a single NVIDIA GeForce RTX 3090 GPU (24 GiB of memory), using Pytorch 2.7 and CUDA 12.4 with driver 550.54.15. 

All denoising networks are trained until the validation error \(E_{256}^{(\mu)}\) has stopped improving for \(100\) epochs, as illustrated in Figure \ref{fig:DenoisingTrainingDynamics} for a selection of drift classes. In red, we show the value of \(E^{(\mu)}_{1}\) evaluated at each epoch, suggesting the denoising objective closely matches the learning error in the drift.

\begingroup
\centering
\setlength{\abovecaptionskip}{0pt}
\setlength{\belowcaptionskip}{0pt}
\setlength{\parskip}{0pt}
\includegraphics[width=0.48\columnwidth]{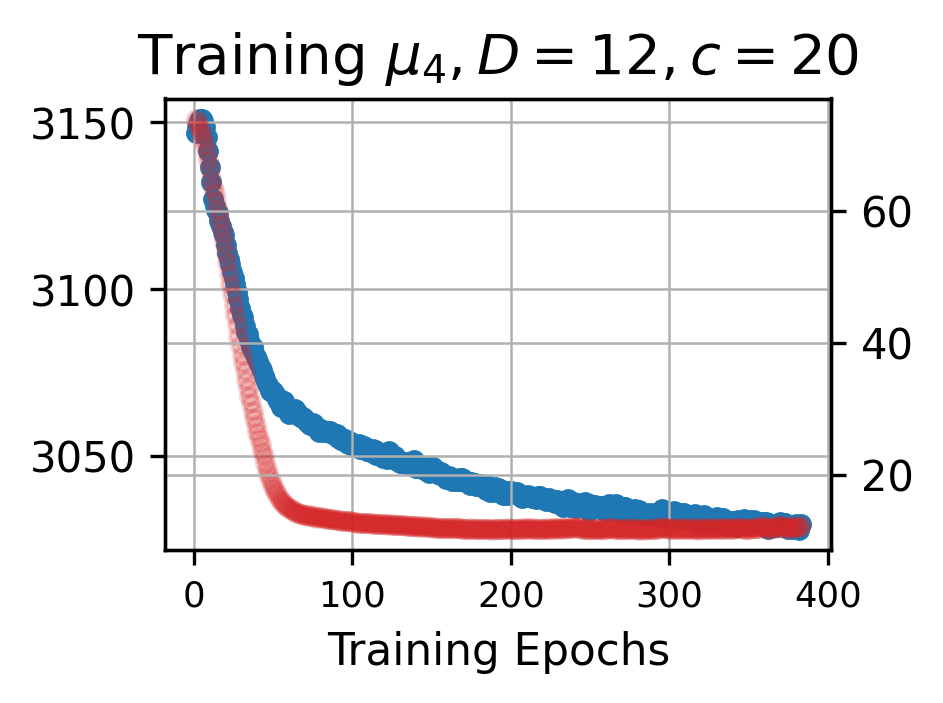} 
\includegraphics[width=0.48\columnwidth]{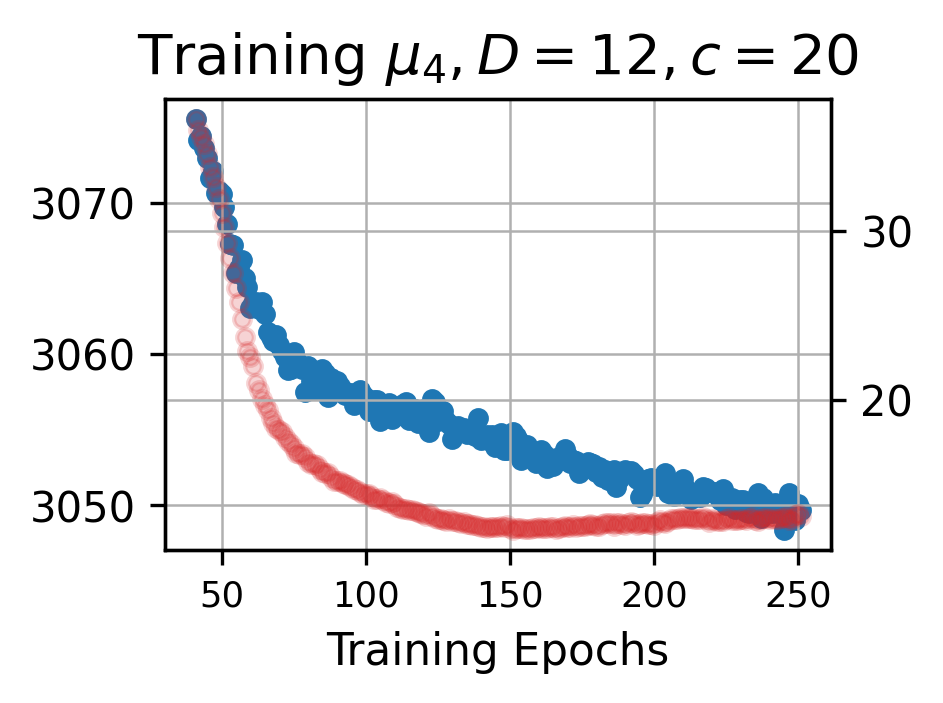} \\
\includegraphics[width=0.48\columnwidth]{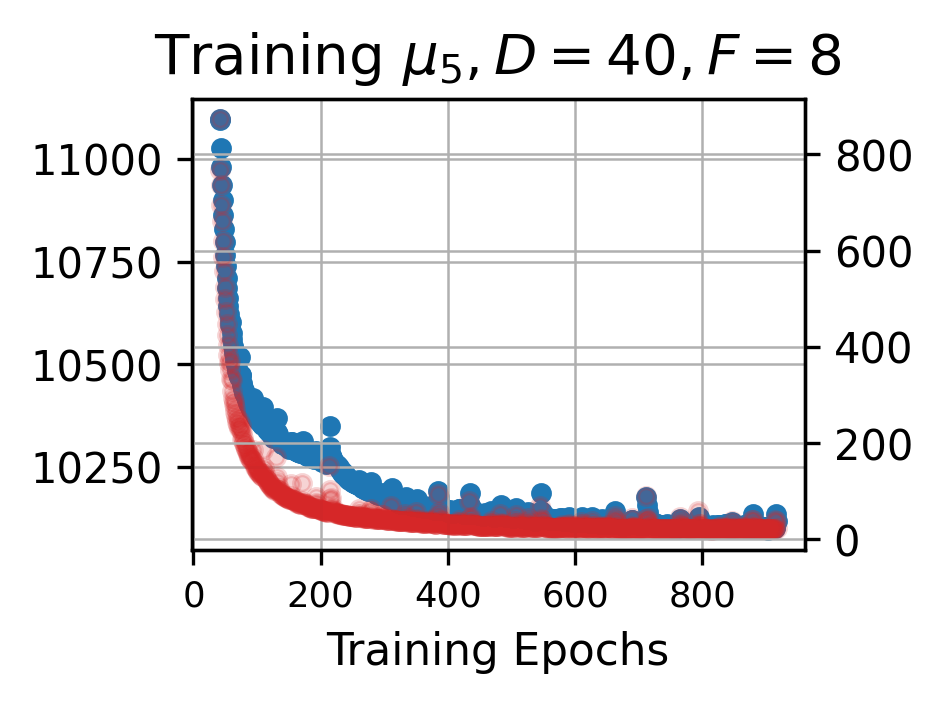} 
\includegraphics[width=0.48\columnwidth]{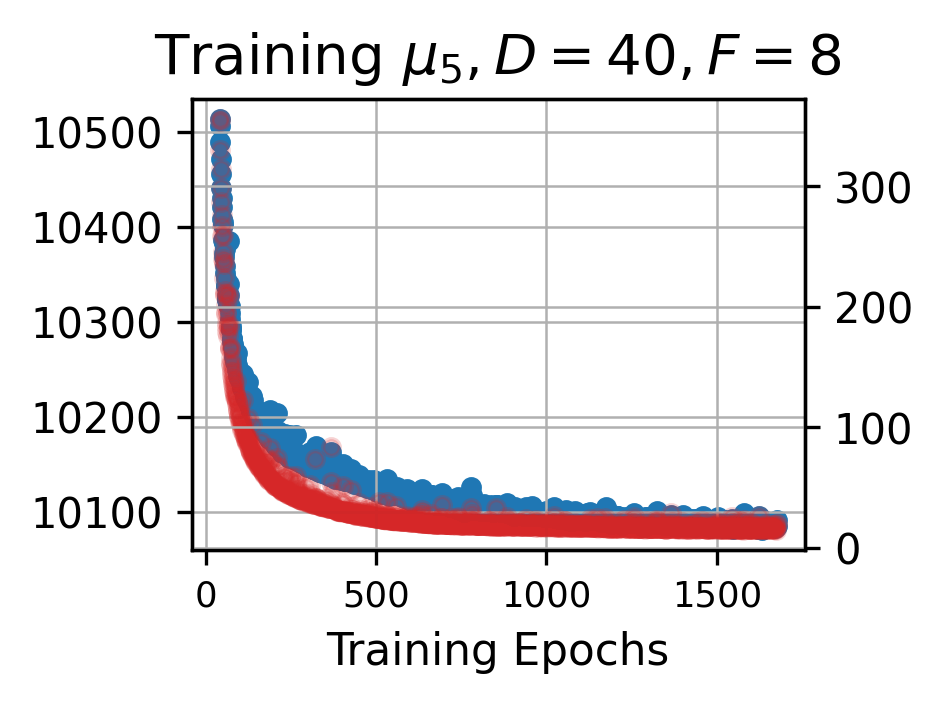} 
\captionof{figure}{\textcolor{mplblue}{Training} loss and \(\mathcolor{red}{E^{(\mu)}_{1}}\) loss for \(\mathrm{DN}\) (left panels), and \(\mathrm{DN}\)-Lin (right panels). Left scale shows training loss, right scale shows validation loss.}
\label{fig:DenoisingTrainingDynamics}
\endgroup

Figure \ref{fig:DenoisingGradientNorms} shows the batch-averaged norm of the gradients of the network output \(D_{\theta}(\tau, X_{\tau}, Y)\) during training of \(\mathrm{DN}\) and \(\mathrm{DN}\)-Lin. Across the selected drift classes, we observe that the gradients with respect to each of the inputs do not vanish across epochs. Figure \ref{fig:DenoisingNormContribution} shows the batch-averaged ratio \(\norm{m_{\theta}}/\norm{D_{\theta}} \) generally increases during training, suggesting the contribution of the diffusion time embedding grows before the model selection occurs. Taken together, Figures \ref{fig:DenoisingGradientNorms} and \ref{fig:DenoisingNormContribution} confirm that the architecture in Appendix \ref{appendix:ModelArchitecture} does not collapse to standard regression using \(Y\).

\begingroup
\centering
\setlength{\abovecaptionskip}{0pt}
\setlength{\belowcaptionskip}{0pt}
\setlength{\parskip}{0pt}
\includegraphics[width=0.48\columnwidth]{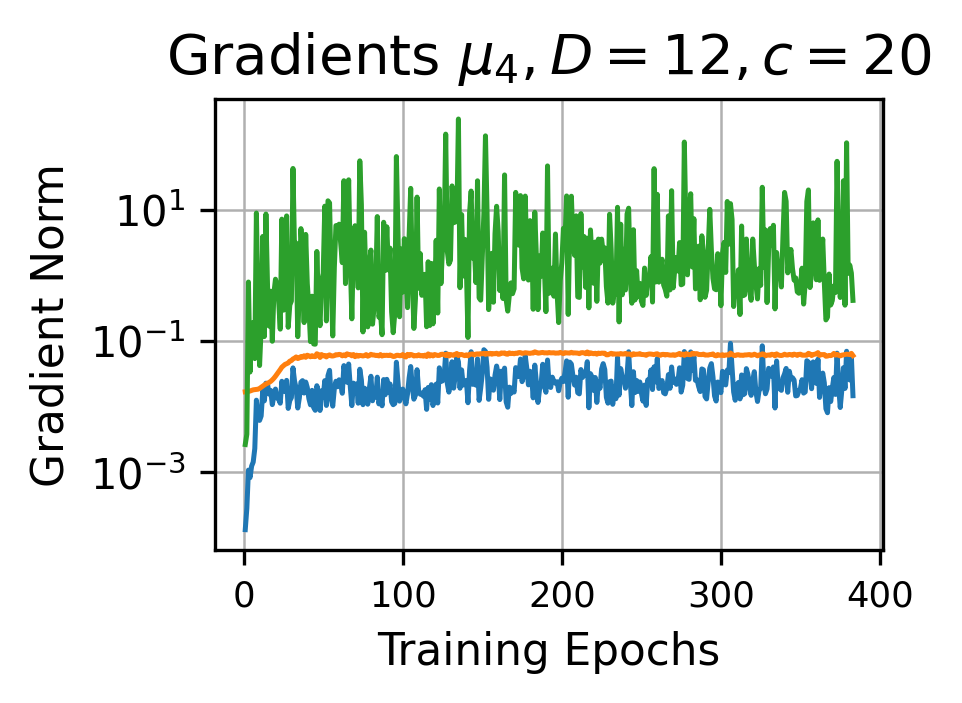} 
\includegraphics[width=0.48\columnwidth]{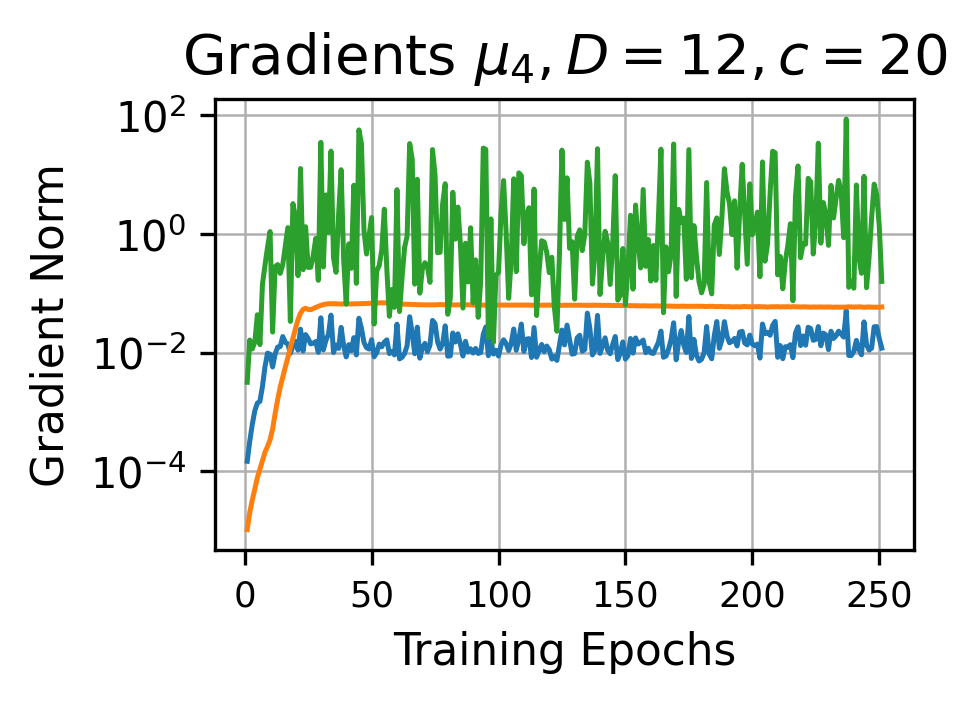} \\
\includegraphics[width=0.48\columnwidth]{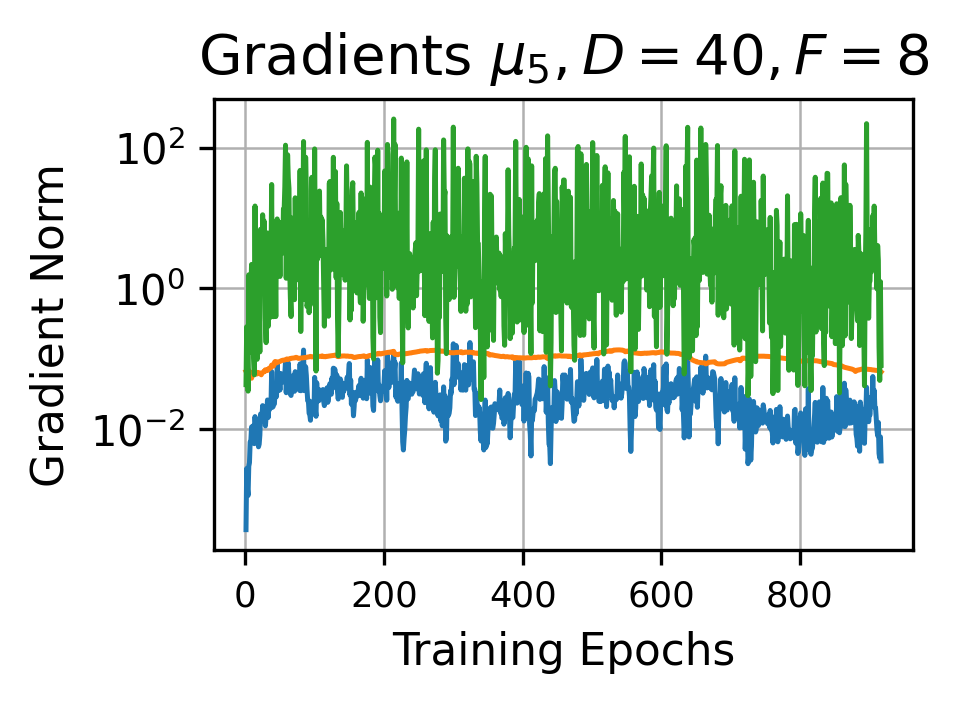} 
\includegraphics[width=0.48\columnwidth]{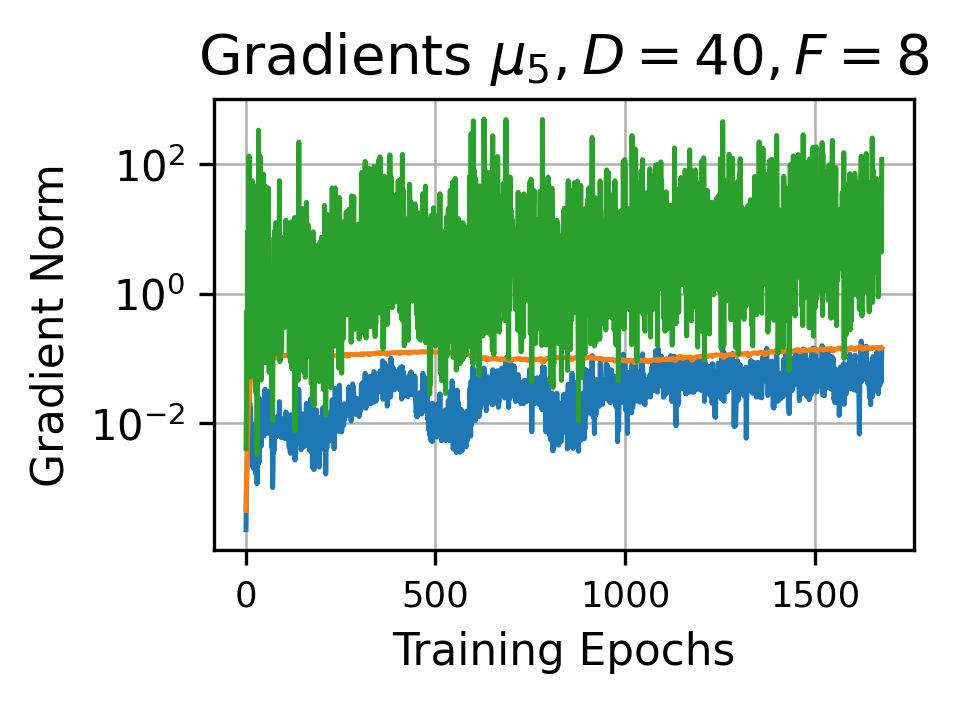} 
\captionof{figure}{Training-time gradient norms with respect to \textcolor{darkgreen}{diffusion time}, \(\mathcolor{mplblue}{X_{\tau}}\), and \(\mathcolor{orange}{Y}\). Left panels show \(\mathrm{DN}\), right panels show \(\mathrm{DN}\)-Lin. }
\label{fig:DenoisingGradientNorms}
\endgroup

\begingroup
\centering
\setlength{\abovecaptionskip}{0pt}
\setlength{\belowcaptionskip}{0pt}
\setlength{\parskip}{0pt}
\includegraphics[width=0.48\columnwidth]{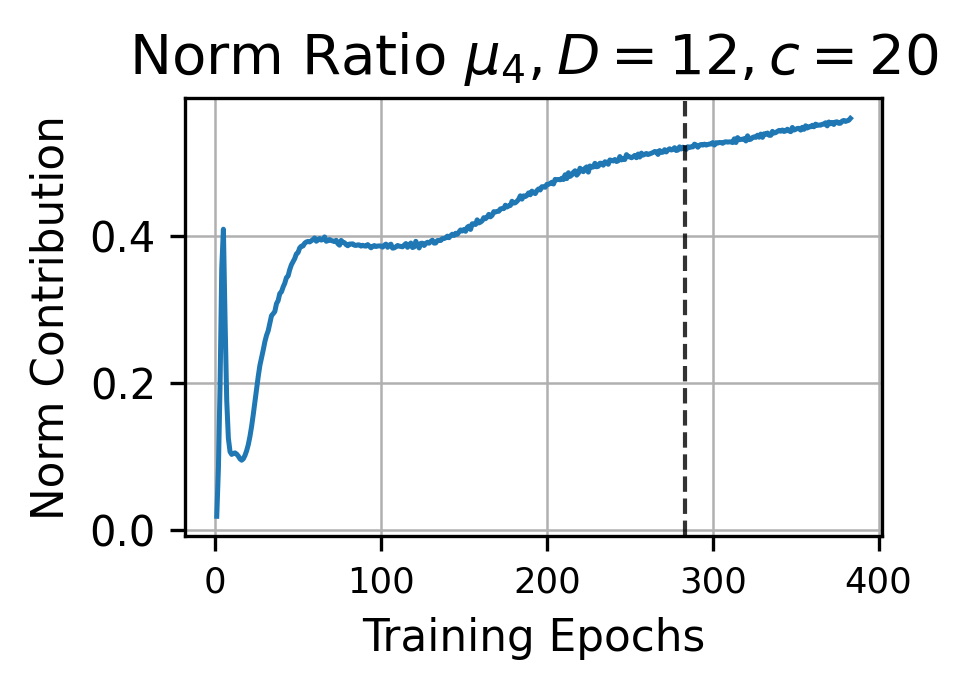} 
\includegraphics[width=0.48\columnwidth]{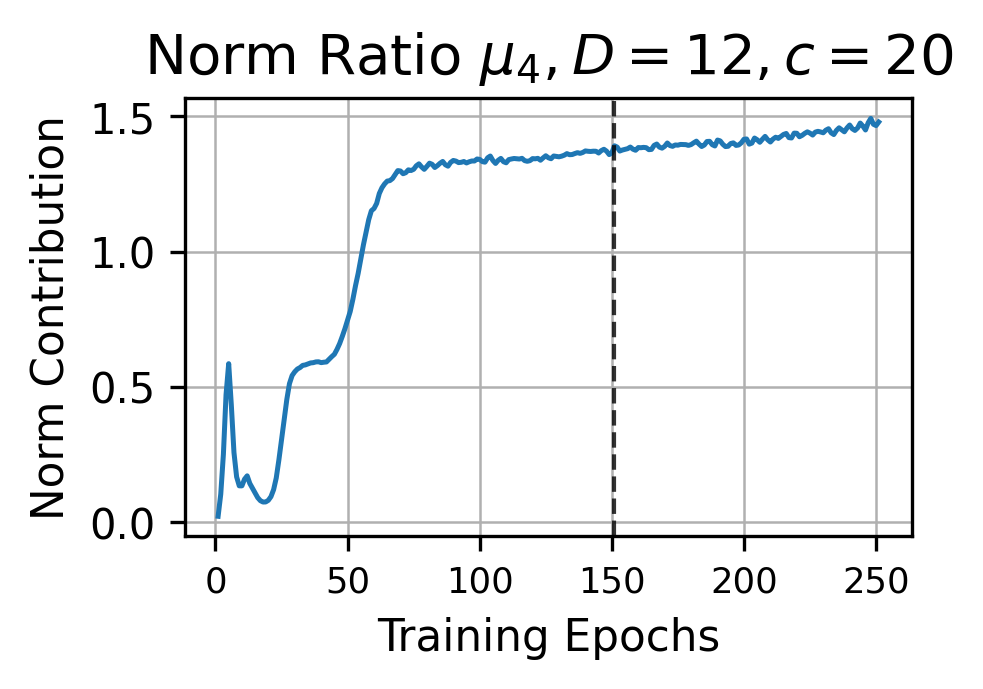} \\
\includegraphics[width=0.48\columnwidth]{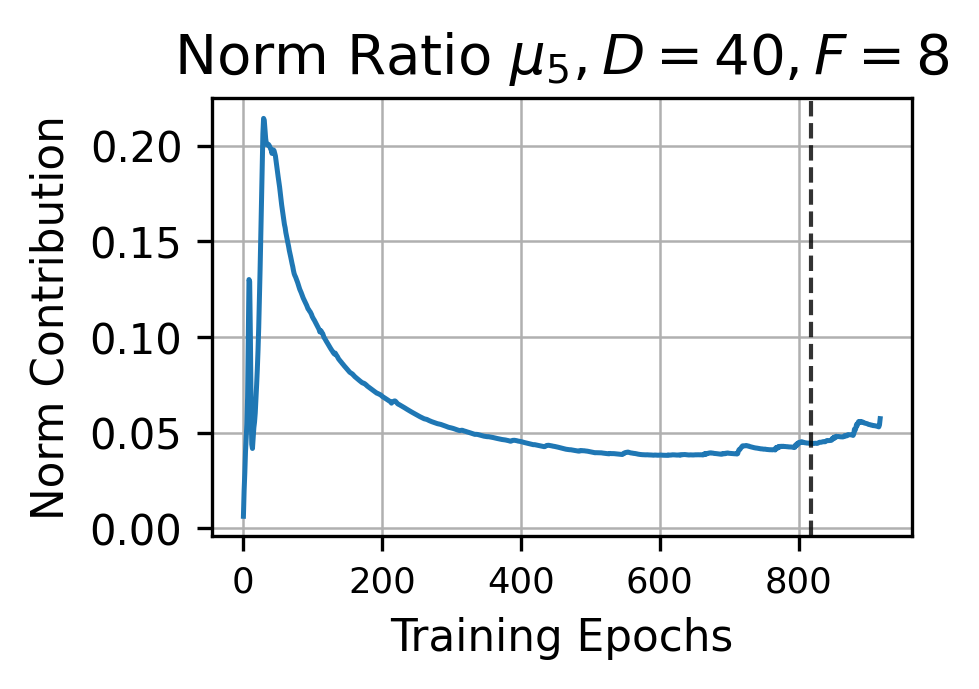} 
\includegraphics[width=0.48\columnwidth]{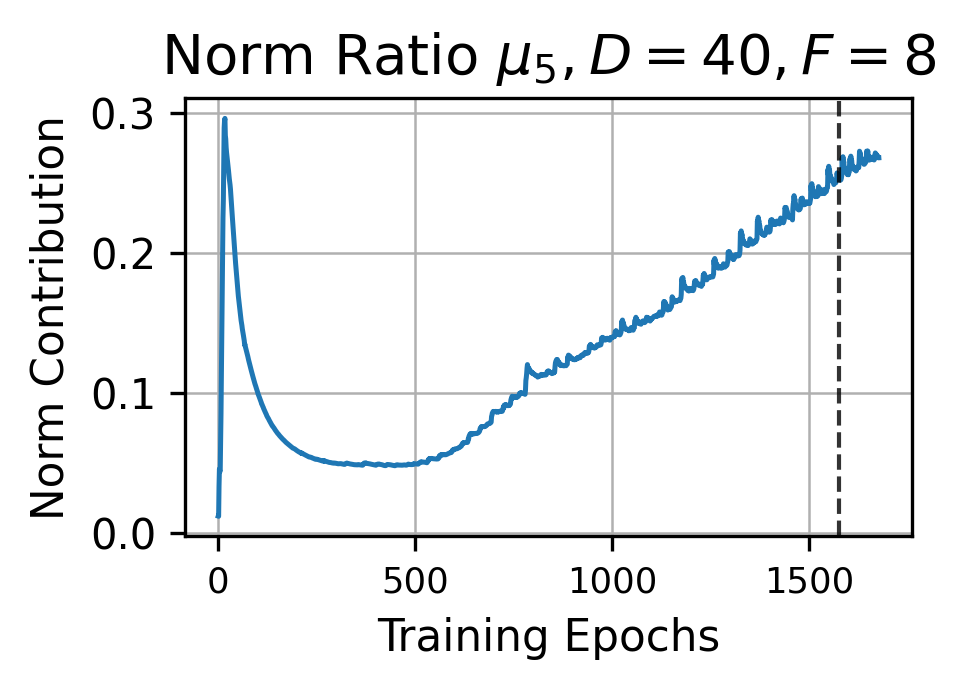} 
\captionof{figure}{Relative norm of \(m_{\theta}\) to network output \(D_{\theta}\) during training. Left panels show \(\mathrm{DN}\), right panels show \(\mathrm{DN}\)-Lin. Vertical line indicates model selection epoch.}
\label{fig:DenoisingNormContribution}
\endgroup

\section{Kernel Drift Estimators in High Dimensions}\label{appendix:KernelEstimators}
For completeness, Table \ref{tab:NWHighDim} shows the out-of-sample performance of the Nadaraya-Watson estimator from \cite{MarieEtAl2021}. As highlighted in Section \ref{section:Implementation} and Appendix \ref{appendix:Baselines}, the kernel bandwidth is chosen through a grid search to minimise \(E_{1}^{(\mu)}\). 

Table~\ref{tab:NWHighDim} shows that the Nadaraya--Watson estimator performs poorly in high dimensions when compared to all baselines reported in Section~\ref{section:OOSFullDataResults}.
\newpage
\begingroup
\refstepcounter{table} {\itshape Table \thetable.}{ Final-time out-of-sample drift error $E^{(\mu)}_{1}$ for kernel (NW) estimators.}
\par\vspace{0.4em}
\noindent
\setlength{\tabcolsep}{3pt}
\renewcommand{\arraystretch}{0.95}
\centerline{
\begin{minipage}[t]{0.48\columnwidth}
\centering
\begin{tabular}{c c c}
\toprule
$c$ & $D$ & $\mathrm{NW}$ \\
\midrule
\multirow{2}{*}{0}
 & 8  & 14.454 \\
 & 12 & 26.544 \\
\midrule
\multirow{2}{*}{20}
 & 8  & 292.256\\
 & 12 & 231.207 \\
\bottomrule
\end{tabular}

\vspace{0.3em}
{\small (a) $\mu_4$}
\end{minipage}
\hfill
\begin{minipage}[t]{0.48\columnwidth}
\centering
\begin{tabular}{c c c}
\toprule
$F$ & $D$ & $\mathrm{NW}$ \\
\midrule
\multirow{2}{*}{$\tfrac{1}{2}$}
 & 20 & 186.914\\
 & 40 & 220.199\\
\midrule
\multirow{2}{*}{8}
 & 20 & 8493\\
 & 40 & 17237\\
\bottomrule
\end{tabular}
\vspace{0.3em}
{\small (b) $\mu_5$}
\end{minipage}
}
\label{tab:NWHighDim}
\endgroup

\section{Sensitivity to number of training trajectories}\label{appendix:LessDataAppendix}

We quantify the sensitivity of the final-time in-sample error \(E_{1}^{(\mu)}\) to a tenfold reduction in the number of training paths, from \(I=10^{3}\) to \(I=10^{2}\). Training procedure is identical to that outlined in Section \ref{section:Implementation}, and we report the in-sample drift error
$E^{(\mu)}_{1}$ for each test drift from Section \ref{section:DriftFunctions}. We define the relative change in the error as \(\left| E^{(\mu)}_{1}(10^{2}) - E^{(\mu)}_{1}(10^{3}) \right|/E^{(\mu)}_{1}(10^{3})\).

Tables~\ref{tab:DNLessData}-\ref{tab:FCConvLessData}
summarize the results for the estimators \((\mathrm{DN})\),  \((\mathrm{DN}\)-Lin), and \(\mathrm{FC}^{+}\)-Conv,
for the bistable potential ($\mu_{4}$) and Lorenz 96 ($\mu_{5}$) drift families. We exclude the \(\mathrm{FC}\) and \(\mathrm{FC}^{+}\) baselines from
this section, as they are not competitive in the corresponding
in-sample regimes in Section \ref{section:FullDataResults}.

\begingroup
\refstepcounter{table}{\itshape Table \thetable.}
In-sample drift error $E^{(\mu)}_1$ of \(\mathrm{DN}\) with $10^2$ training paths,
for $\mu_4$ and $\mu_5$. Relative change is computed with respect to the corresponding $10^3$-path results reported in Section \ref{section:FullDataResults}.
\par\vspace{0.3em}
\centering
\setlength{\tabcolsep}{3pt}
\renewcommand{\arraystretch}{0.95}
\centerline{
\begin{tabular}{c c c c c}
\toprule
Drift & Param & $D$ & $E^{(\mu)}_1(10^2)$ & Rel.\ change \\
\midrule
\multirow{4}{*}{$\mu_4$}
 & \multirow{2}{*}{$c=0$}  & 8  & 5.483  & 4.24 \\
 &        & 12 & 10.415 & 5.20 \\
 & \multirow{2}{*}{\(c=20\)} & 8  & 17.188 & 1.03 \\
 &        & 12 & 19.638 & 0.84 \\
\midrule
\multirow{4}{*}{$\mu_5$}
 & \multirow{2}{*}{\(F=0.5\)} & 20 & 22.629  & 3.65 \\
 &         & 40 & 38.291  & 2.99 \\
 & \multirow{2}{*}{\(F=8\)}   & 20 & 60.648  & 5.36 \\
 &         & 40 & 115.197 & 5.44 \\
\bottomrule
\end{tabular}
}
\label{tab:DNLessData}
\endgroup

\newpage
\begingroup
\refstepcounter{table}{\itshape Table \thetable.}{
In-sample drift error $E^{(\mu)}_1$ of \(\mathrm{DN}\)-Lin with $10^2$ training paths, for $\mu_4$ and $\mu_5$.}
\par\vspace{0.3em}
\centering
\setlength{\tabcolsep}{3pt}
\renewcommand{\arraystretch}{0.95}
\centerline{
\begin{tabular}{c c c c c}
\toprule
Drift & Param & $D$ & $E^{(\mu)}_1(10^2)$ & Rel.\ change \\
\midrule
\multirow{4}{*}{$\mu_4$}
 & \multirow{2}{*}{\(c=0\)}  & 8  & 12.517 & 0.80 \\
 &        & 12 & 19.913 & 0.76 \\
 & \multirow{2}{*}{\(c=20\)} & 8  & 11.935 & 0.16 \\
 &        & 12 & 15.483 & 0.13 \\
\midrule
\multirow{4}{*}{$\mu_5$}
 & \multirow{2}{*}{\(F=0.5\)} & 20 & 19.172 & 5.13 \\
 &         & 40 & 28.209 & 2.63 \\
 & \multirow{2}{*}{\(F=8\)}   & 20 & 53.942 & 8.42 \\
 &         & 40 & 85.047 & 5.97 \\
\bottomrule
\end{tabular}
}
\label{tab:DNLinLessData}
\endgroup

\begingroup
\refstepcounter{table}{\itshape Table \thetable.}{
In-sample drift error $E^{(\mu)}_1$ of \(\mathrm{FC}^{+}\)-Conv with $10^2$ training paths, for $\mu_4$ and $\mu_5$.}
\par\vspace{0.3em}
\centering
\setlength{\tabcolsep}{3pt}
\renewcommand{\arraystretch}{0.95}
\centerline{
\begin{tabular}{c c c c c}
\toprule
Drift & Param & $D$ & $E^{(\mu)}_1(10^2)$ & Rel.\ change \\
\midrule
\multirow{4}{*}{$\mu_4$}
 & \multirow{2}{*}{\(c=0\)}  & 8  & 11.425 & 1.93 \\
 &        & 12 & 19.065 & 2.02 \\
 & \multirow{2}{*}{\(c=20\)} & 8  & 21.307 & 1.36 \\
 &        & 12 & 45.019 & 2.20 \\
\midrule
\multirow{4}{*}{$\mu_5$}
 & \multirow{2}{*}{\(F=0.5\)} & 20 & 18.383 & 3.93 \\
 &         & 40 & 30.650 & 1.57 \\
 & \multirow{2}{*}{\(F=8\)}   & 20 & 47.310 & 3.12 \\
 &         & 40 & 94.261 & 4.78 \\
\bottomrule
\end{tabular}
}
\label{tab:FCConvLessData}
\endgroup

Across all settings, reducing the number of training paths from
$I=10^{3}$ to $I=10^{2}$ leads to a consistent increase in final-time
in-sample error. While the relative ordering of the estimators is not preserved relative to Section \ref{section:FullDataResults}, the observed changes reflect differences in the magnitude of degradation, rather than the collapse of any particular estimator, as no estimator exhibits abrupt divergence behaviour under reduced data.

\section{Extended Out-of-Sample Results}\label{appendix:AdditionalDDims}
We report additional results for the bistable drift \(\mu_{4}\), and we extend the main comparison in Section \ref{section:OOSFullDataResults} to a higher dimension (\(D=20\)) for the case \(c=20\). Tables \ref{tab:AdditionalDDimsResults} and \ref{tab:AdditionalDDimsNSResults} include the denoising estimator (\(\mathrm{DN}\)) and the convolutional baseline (\(\mathrm{FC}^{+}\)-Conv). We additionally introduce (\(\mathrm{FC}^{+}\)-Conv-MLPSM), in which the fully connected layers in \(\mathrm{FC}^{+}\)-Conv are replaced by the \textit{MLPStateMapper} in \(\mathrm{DN}\). Training and evaluation procedures are identical to those reported in Section~\ref{section:Implementation}.

\begingroup
\refstepcounter{table}{\itshape Table \thetable.}{ Final-time OOS drift error \(E^{(\mu)}_{20}\), for \(\mu_4, c=0\).}
\par\vspace{0.3em}
\centering
\setlength{\tabcolsep}{3pt}
\renewcommand{\arraystretch}{0.95}
\centerline{
\begin{tabular}{c c c c c }
\toprule
$c$ & $D$ & \(\mathrm{DN}\) & \(\mathrm{FC}^{+}\)-Conv-MLPSM & \(\mathrm{FC}^{+}\)-Conv  \\
\midrule
\multirow{2}{*}{0}
 & 8  &\underline{1.546}&0.793\(^{*}\)&9.744\\
 & 12 &\underline{2.314}&1.695\(^{*}\)&8.344\\
\bottomrule
\end{tabular}
}
\label{tab:AdditionalDDimsResults}
\endgroup
\newpage
\begingroup
\refstepcounter{table}{\itshape Table \thetable.}{ Final-time OOS drift error \(E^{(\mu)}_{20}\), for \(\mu_4, c=20\).}
\par\vspace{0.3em}
\centering
\setlength{\tabcolsep}{3pt}
\renewcommand{\arraystretch}{0.95}
\begin{tabular}{c c c c c }
\toprule
$c$ & $D$ & \(\mathrm{DN}\) & \(\mathrm{FC}^{+}\)-Conv-MLPSM & \(\mathrm{FC}^{+}\)-Conv  \\
\midrule
\multirow{3}{*}{20}
 & 8  &50.919\(^{*}\)&124.972&\underline{76.061}\\
 & 12 & 81.721\(^{*}\)&\underline{111.634}&121.379\\
  & 20 & 182.566\(^{*}\)&278.475&\underline{245.536}\\
\bottomrule
\end{tabular}
\label{tab:AdditionalDDimsNSResults}
\endgroup

In the separable regime \((c=0)\), the \textit{MLPStateMapper} leads to improved performance of standard regression baselines, as (\(\mathrm{FC}^{+}\)-Conv-MLPSM) lower OOS errors than its denoising-based counterpart. In contrast, in the non-separable regime  \((c=20)\), \(\mathrm{FC}^{+}\)-Conv-MLPSM consistently underperforms \(\mathrm{DN}\) across all dimensions, indicating that architectural bias alone is insufficient to recover the observed OOS gains.

\section{Sensitivity to time series sampling frequency}\label{appendix:SamplingFrequency}
As motivated in the Introduction, in increment-based regression, decreasing the sampling frequency \(\Delta\) can lead to high variance estimators. In this section, we study the stability of the denoising estimator to different sampling frequencies \(\Delta\) and compare it to the estimator \(\mathrm{FC}^{+}\)-Conv-MLPSM (see Appendix \ref{appendix:AdditionalDDims}). We consider the bi-stable drift \(\mu_{4}\) in both separable \((c=0)\) and non-separable \((c=20)\) regimes. We exclude the Lorenz 96 system from this analysis, as its behaviour is dominated by advective dynamics, making a direct comparison of sampling-frequency effects less informative.

\begingroup
\refstepcounter{table}{\itshape Table \thetable.}
{Final time out-of-sample error \(E_{20}^{(\mu)}\) for \(\mu_{4}\) and \(\mathrm{DN}\) and \(\mathrm{FC}^{+}\)-Conv-MLPSM for different sampling frequencies \(\Delta\).}
\par\vspace{0.3em}
\centering
\setlength{\tabcolsep}{3pt}
\renewcommand{\arraystretch}{0.95}
\centerline{
\begin{tabular}{c c c c}
\hline
$c$ & $\Delta$ & $\mathrm{DN}$ & $\mathrm{FC}^{+}$-Conv-MLPSM \\
\hline
\multirow{3}{*}{$0$}
& \(64^{-1}\)   & 0.953 & 0.784 \\
& \(256^{-1}\)  & 1.546 & 0.793 \\
& \(1024^{-1}\) & 1.582 & 0.764 \\
\hline
\multirow{3}{*}{$20$}
& \(64^{-1}\)   & 119.032 & 92.415 \\
& \(256^{-1}\)  & 50.919  & 124.972 \\
& \(1024^{-1}\) & 90.617  & 162.930 \\
\hline
\end{tabular}
}
\label{tab:DeltaTable}
\endgroup

Table~\ref{tab:DeltaTable} shows that out-of-sample performance depends on the sampling frequency \(\Delta\) for both estimators, exhibiting non-monotone behaviour as \(\Delta\) decreases. In the non-separable regime, \(\mathrm{FC}^{+}\)-Conv-MLPSM performs better at coarser sampling ($\Delta = 1/64$) than at finer sampling ($\Delta = 1/1024$), while the denoising estimator $\mathrm{DN}$ remains superior as \(\Delta\) is reduced.

\section{Feasible Model Validation}\label{appendix:FeasibleValidation}
We consider a feasible model validation strategy for denoising-based estimators \(\mathrm{DN}\), \(\mathrm{DN}\)-Lin. The neural network-based baselines in Section \ref{section:FullDataResults} are trained on \eqref{eq:FeasiblePosteriorConditionalObjective} and the best model is selected by early stopping based on the drift error \(E_{1}^{(\mu)}\). Since this requires knowledge of the true drift,  we propose an alternative that does not require oracle access to the drift.

In particular, we define the validation error using the training objective \eqref{eq:FeasiblePosteriorConditionalObjective}, evaluated on \(\mathcal{I}=50\) held out paths. We report the out-of-sample performance of \(\mathrm{DN}, \mathrm{DN}\)-Lin using oracle and feasible validation strategies in Tables \ref{tab:FeasibleDN}-\ref{tab:FeasibleDNLin} below. 

\begingroup
\refstepcounter{table}{\itshape Table \thetable.}{
Final-time out-of-sample drift error \(E_{20}^{(\mu)}\), for $\mathrm{DN}$ under oracle and feasible validation.}
\par\vspace{0.3em}
\centering
\small
\setlength{\tabcolsep}{3pt}
\renewcommand{\arraystretch}{0.95}
\centerline{
\begin{tabular}{c c c c c}
\toprule
Drift & Param & $D$ & Oracle & Feasible \\
\midrule
\multirow{4}{*}{$\mu_4$}
 & \multirow{2}{*}{$c=0$}  
 & 8  & 1.546 & 3.361 \\
 &                       
 & 12 & 2.314 & 3.290 \\
 & \multirow{2}{*}{$c=20$}
 & 8  & 50.919 & 50.218 \\
 &                       
 & 12 & 81.721 & 78.356 \\
\midrule
\multirow{4}{*}{$\mu_5$}
 & \multirow{2}{*}{$F=0.5$}
 & 20 & 8.718 & 12.105 \\
 &                       
 & 40 & 12.198 & 10.680 \\
 & \multirow{2}{*}{$F=8$}
 & 20 & 3973.37 & 1854.08 \\
 &                       
 & 40 & 1243.11 & 1356.77 \\
\bottomrule
\end{tabular}
}
\label{tab:FeasibleDN}
\endgroup

\begingroup
\refstepcounter{table}{\itshape Table \thetable.}{
Final-time out-of-sample drift error \(E_{20}^{(\mu)}\), for $\mathrm{DN}$-Lin under oracle and feasible validation.}
\par\vspace{0.3em}
\centering
\small
\setlength{\tabcolsep}{3pt}
\renewcommand{\arraystretch}{0.95}
\centerline{
\begin{tabular}{c c c c c}
\toprule
Drift & Param & $D$ & Oracle & Feasible \\
\midrule
\multirow{4}{*}{$\mu_4$}
 & \multirow{2}{*}{$c=0$}  
 & 8  & 11.800 & 18.327 \\
 &                       
 & 12 & 16.616 & 25.849 \\
 & \multirow{2}{*}{$c=20$}
 & 8  & 67.297 & 64.894 \\
 &                       
 & 12 & 115.750 & 109.155 \\
\midrule
\multirow{4}{*}{$\mu_5$}
 & \multirow{2}{*}{$F=0.5$}
 & 20 & 12.024 & 10.008 \\
 &                       
 & 40 & 11.134 & 14.835 \\
 & \multirow{2}{*}{$F=8$}
 & 20 & 172.783 & 346.931 \\
 &                       
 & 40 & 332.905 & 371.282 \\
\bottomrule
\end{tabular}
}
\label{tab:FeasibleDNLin}
\endgroup

Across the settings considered, Tables~\ref{tab:FeasibleDN} and~\ref{tab:FeasibleDNLin} show feasible validation selects models of the same order of performance as oracle selection for the same estimator, without inducing catastrophic outcomes. Occasional cases where feasible validation yields an estimator with lower out-of-sample error than oracle selection are expected, since the oracle criterion targets drift recovery under the in-sample distribution, which can be significantly different from the out-of-sample distribution. These results indicate that feasible validation can be sufficient for model selection in practice, without relying on access to the true drift.

\end{document}